\documentclass{article} %
\usepackage{iclr2027_conference,times}

\usepackage{amsmath,amsfonts,bm}

\def\eqref#1{equation~\ref{#1}}

\def\1{\bm{1}}

\DeclareMathAlphabet{\mathsfit}{\encodingdefault}{\sfdefault}{m}{sl}
\SetMathAlphabet{\mathsfit}{bold}{\encodingdefault}{\sfdefault}{bx}{n}

\newcommand{\reg}{\lambda}

\usepackage{xspace}
\usepackage{xcolor}
\usepackage{colortbl}
\usepackage{tabularx}
\definecolor{tablegroup}{HTML}{EDF0F2}
\definecolor{tableours}{HTML}{F0F6F4}
\usepackage{graphicx}
\usepackage{booktabs}
\usepackage{multirow}
\usepackage{makecell}
\usepackage{enumitem}
\usepackage{pifont}
\usepackage{amssymb}
\usepackage{amsthm}
\usepackage{subcaption}
\usepackage{float}
\usepackage{algorithm}
\usepackage{algpseudocode}
\newsavebox{\vsqaForwardBody}
\newsavebox{\vsqaBackwardBody}

\usepackage{placeins}
\usepackage{latexml}
\iflatexml\else
  \usepackage[most]{tcolorbox}
\fi
\usepackage{hyperref}
\definecolor{citebluegray}{HTML}{526D86}
\hypersetup{citecolor=citebluegray}
\makeatletter
\AddToHook{cmd/hyper@natlinkstart/before}{%
    \begingroup
    \hypersetup{pdfborder={0 0 0}}%
    \def\Hy@colorlink{\begingroup\HyColor@UseColor}%
}
\AddToHook{cmd/hyper@natlinkend/after}{\endgroup}
\makeatother
\usepackage{url}
\usepackage{xurl} %

\graphicspath{{figure/}}

\newcommand{\method}{\textsc{VSQA}\xspace}

\newcommand{\xhdr}[1]{{\noindent\colormodeheading{\bfseries #1}.}}

\definecolor{findingaccent}{HTML}{617C96}
\definecolor{findingink}{HTML}{1C2B33}
\iflatexml
\newcounter{takeaway}
\newenvironment{takeawaybox}{%
    \par\refstepcounter{takeaway}\noindent\textsc{Takeaway~\thetakeaway.}\enspace
}{\par}
\else
\newtcolorbox[auto counter]{takeawaybox}{
    enhanced,
    breakable,
    colframe=findingaccent,
    colback=findingaccent!5!white,
    coltext=findingink,
    boxrule=0.65pt,
    arc=3pt,
    outer arc=3pt,
    boxsep=0pt,
    left=7pt,
    right=7pt,
    top=3pt,
    bottom=3pt,
    before skip=3pt,
    after skip=3pt,
    before upper={\colormodeinblock\textsc{Takeaway~\thetcbcounter.}\enspace},
}

\fi

\newcommand{\colormodeheading}{}
\newcommand{\colormodeinblock}{}

\title{JetDiffusion: Pushing the Speed Limits of Video Generation with NVFP4 Sparse Attention}
\iflatexml
\usepackage{authblk}
\author[1]{Yulun Wu}
\author[1]{Zihan Min}
\author[1]{Minshen Zhang}
\author[1]{Kaiqin Kong}
\author[2]{Ji Xie}
\author[1]{Will Lin}
\author[1]{Junda Chen}
\author[3]{Wenxuan Tan}
\author[1]{Aryan Kumar}
\author[1]{Loay Rashid}
\author[4]{Qingyu Yin}
\author[1]{Hao Zhang}
\affil[1]{University of California, San Diego}
\affil[2]{Carnegie Mellon University}
\affil[3]{University of Wisconsin--Madison}
\affil[4]{National University of Singapore}
\else
\author{\bfseries Yulun Wu$^{1*}$, Zihan Min$^{1*}$, Minshen Zhang$^{1*}$, Kaiqin Kong$^{1}$, \\
\bfseries Ji Xie$^{2}$, Will Lin$^{1}$, Junda Chen$^{1}$, Wenxuan Tan$^{3}$, \\
\bfseries Aryan Kumar$^{1}$, Loay Rashid$^{1}$, Qingyu Yin$^{4}$, Hao Zhang$^{1\dagger}$ \\[4pt]
\normalfont $^{1}$University of California, San Diego \quad $^{2}$Carnegie Mellon University \\
\normalfont $^{3}$University of Wisconsin--Madison \quad $^{4}$National University of Singapore \\[2pt]
\normalfont $^{*}$Equal contribution \quad $^{\dagger}$Corresponding author: \texttt{haozhang@ucsd.edu}}
\fi

\iclrfinalcopy

\begin{document}

\maketitle
\iflatexml
\noindent Yulun Wu, Zihan Min, and Minshen Zhang contributed equally.
The corresponding author is Hao Zhang (\href{mailto:haozhang@ucsd.edu}{\texttt{haozhang@ucsd.edu}}).\par
\fi

\begin{figure}[H]
    \centering
    \includegraphics[width=\linewidth]{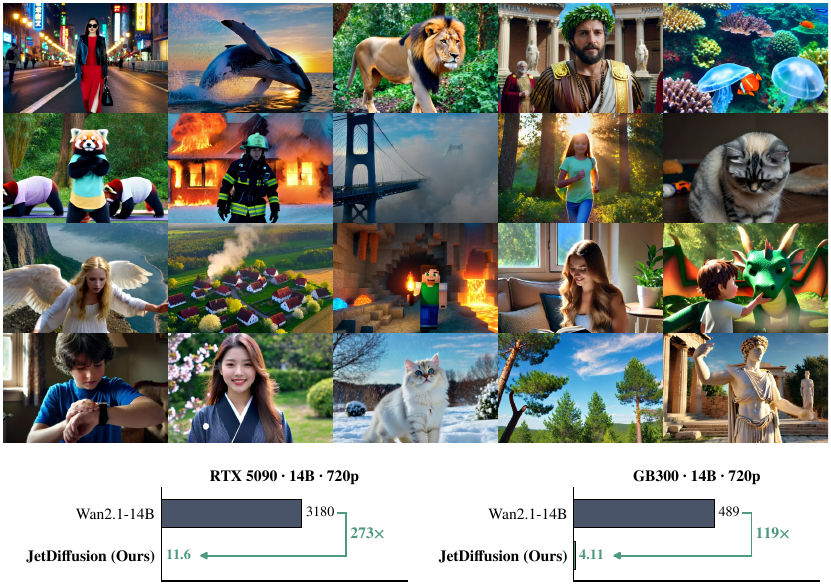}
    \caption{\textbf{Accelerating video generation with JetDiffusion.}
    Top: frames generated by JetDiffusion from Wan2.1-14B.
    Bottom: end-to-end latency (s) of one Wan2.1-14B 720p generation on an RTX 5090 and a GB300, compared with the base model.
    Appendix~\ref{app:qualitative} gives the full prompts and compares frames from each video with the base model.}
    \label{fig:teaser}
\end{figure}
\begin{abstract}
\label{sec:abstract}

Attention over long sequences is a major computational bottleneck in video generation, which can be alleviated through sparsity and low-bit quantization. We propose \textbf{JetDiffusion}, a video generation system built on trainable \textbf{V}ideo \textbf{S}parse \textbf{Q}uantized \textbf{A}ttention (\textbf{VSQA}), with custom inference kernels that combine native NVFP4 computation and kernel fusion. At 90\% sparsity, VSQA accelerates attention by 8--29x compared with FlashAttention on different hardwares. To further accelerate video generation, we combine VSQA with NVFP4 linear quantization and step distillation. Although each technique can individually preserve generation quality, jointly adapting to all three remains challenging, particularly under aggressive NVFP4 quantization. Through empirical studies, we develop an effective staged training recipe that enables stable adaptation to quantization, sparsity, and few-step denoising. On Wan2.1-14B, the resulting model achieves 273x end-to-end video generation speedup on the RTX 5090 while maintaining comparable video quality.
\end{abstract}

\section{Introduction}
\label{sec:intro}

Attention is a major computational bottleneck in video generation: higher resolution and longer duration increase the number of spatiotemporal tokens, and the cost of full attention grows quadratically with it.
Sparse attention~\citep{zhang2025fast,xi2025sparse,zhang2026faster,li2026radial,zhang2026spargeattention2} and low-bit attention~\citep{zhang2025sageattention,zhang2024sageattention2,zhang2025sageattention3,zhang2026attn} reduce this cost in complementary ways: sparsity reduces the number of token interactions, while quantization lowers the cost of the interactions that remain.
However, combining high sparsity with 4-bit quantization significantly perturb the pretrained model, causing compounding loss in generation quality.

Recent works apply post-training quantization to sparse attention without any training~\citep{zhang2025spargeattn,zhao2026paroattention}, or adapt the model to sparsity and quantization jointly through training~\citep{liu2025fpsattention,zhang2026sla2}.
These efforts mainly target 8-bit attention.
Native NVFP4 support on NVIDIA Blackwell GPUs~\citep{alvarez2025nvfp4} makes 4-bit matrix multiplication substantially cheaper, offering an opportunity to further accelerate sparse attention.
However, 4-bit precision leaves far less room for error than 8-bit. With NVFP4, quantization error grows sharply, and post-training calibration alone no longer preserves the quality of the pretrained model~\citep{zhang2026attn}. Moreover, quantizing attention breaks the identities that standard attention kernels rely on in the backward pass, requiring dedicated kernels for trainable NVFP4 sparse attention.

We propose \textbf{V}ideo \textbf{S}parse \textbf{Q}uantized \textbf{A}ttention (\method), to our knowledge the first trainable attention method that combines high sparsity with NVFP4 quantization.
\method keeps the coarse-to-fine structure of VSA~\citep{zhang2026faster}, which pools tokens into cubes to select the cubes each query attends to, and computes both $QK^\top$ and $PV$ of the selected token-level attention in NVFP4.
The inference kernels use native NVFP4 matrix multiplication, traverse only the selected cube pairs, and fuse pooling with quantization so that non-matrix work stays small as sparsity increases. 
We implement both training and inference kernels for the RTX 5090 and the GB300. 

To achieve acceleration in the end-to-end system, we extend \method into \textbf{JetDiffusion}, a complete system where we additionally quantize the linear layers to NVFP4 and distill the model to three denoising steps. 
Realizing these compression techniques without sacrificing generation quality requires adapting the model to the combined effects of low-precision computation, sparse attention, and few-step denoising.
To maximize speedup, our quantization setup operates without smoothing tricks, making quantization error a central challenge for this adaptation.

Through empirical observations, we find that a dedicated reconstruction warm-up substantially stabilizes subsequent training by reducing quantization error. While conventional QAT/QAD objectives can reduce quantization error, our observations motivate a more targeted objective: layer-wise reconstruction, which directly matches quantized features to a frozen, unquantized reference~\citep{li2021brecq}.
Following this warm-up, we enable sparse attention and proceed with ODE initialization followed by DMD refinement, as in prior video distillation work~\citep{yin2024one,yin2025slow,zhu2026causal}, retaining reconstruction regularization during ODE initialization to limit distribution drift.

We apply this recipe to Wan2.1-1.3B at 480p and Wan2.1-14B at 720p~\citep{wan2025wan}.
At $90\%$ sparsity, \method accelerates attention on Wan2.1-14B by $8.0\times$ over dense FlashAttention-4~\citep{zadouri2026flashattention} on the GB300, and $28.5\times$ over FlashAttention-2 on the RTX5090, much faster than other implementations such as SageSLA~\citep{zhang2025turbodiffusion}.
With NVFP4 linear layers and three-step generation, JetDiffusion achieves $273\times$ end-to-end speedup on Wan2.1-14B on the RTX 5090 and $119\times$ on the GB300, with VBench scores that match or exceed those of the dense BF16 base model.

Our main contributions are as follows:
\begin{itemize}[leftmargin=*, topsep=3pt, itemsep=3pt, parsep=0pt]
\item \textbf{Trainable NVFP4 sparse attention.}
\method is, to our knowledge, the first trainable attention that combines high sparsity with NVFP4 quantization, with training kernels and inference kernels for the RTX 5090 and the GB300.
\item \textbf{A stable training recipe for NVFP4 sparse few-step generation.}
Guided by empirical findings on how quantization and sparsity are recovered, we separate quantization recovery from sparse few-step adaptation, using layer reconstruction and regularized ODE initialization before DMD refinement.
\item \textbf{JetDiffusion: ultra-fast video generation.}
On Wan2.1-14B at 720p, \method delivers a $8.0\times$ attention speedup over dense FlashAttention-4 on the GB300, and JetDiffusion delivers $273\times$ end-to-end speedup on the RTX 5090 and $119\times$ on the GB300 while preserving generation quality.
\end{itemize}

\section{Related Work}
\label{sec:related_work}

\xhdr{Sparse and low-bit attention}
Sparse attention reduces token interactions through structured patterns or adaptive token and block selection~\citep{zhang2025fast,zhang2026faster,xi2025sparse,yang2025sparse,xia2025training,shen2025draftattention,wu2026vmoba,zhang2026spargeattention2,li2026radial,chen2026sparse,sun2026vorta,tan2026dynamic,zhan2025bidirectional}.
Other methods compensate for omitted interactions through auxiliary approximations or residual reconstruction~\citep{zhang2026sla,zhou2026svg,taghavi2026partition}.
Low-bit attention accelerates computation by reducing numerical precision~\citep{zhang2025sageattention,chen2024int,zhang2025sageattention2++}, using techniques such as smoothing to mitigate quantization error~\citep{zhang2024sageattention2,zhang2025sageattention3,li2026vc}.
Attn-QAT adapts models to 4-bit attention through quantization-aware training~\citep{zhang2026attn}, while SageBwd studies trainable 8-bit attention~\citep{zhang2026sagebwd}.
Combining sparsity and quantization, SpargeAttn and PAROAttention provide training-free approaches~\citep{zhang2025spargeattn,zhao2026paroattention}, whereas FPSAttention and SLA2 use training to preserve quality~\citep{liu2025fpsattention,zhang2026sla2}.
To the best of our knowledge, \method is the first trainable NVFP4 sparse attention method for video generation.

\xhdr{Few-step distillation}
Few-step generation can be learned through consistency objectives~\citep{song2023consistency,lu2025simplifying,zheng2026large} or distribution matching methods such as Diff-Instruct and Distribution Matching Distillation (DMD), which align student and teacher output distributions~\citep{luo2023diff,yin2024one,yin2024improved}.
For DMD in autoregressive video generation, CausVid initializes the student with teacher-generated ODE trajectories~\citep{yin2025slow}, and Causal Forcing examines their compatibility with the student's causal conditioning~\citep{zhu2026causal}.
Related work reduces train--test conditioning gaps using student-generated histories~\citep{huang2026self} and studies consistency-based initialization and its alignment with DMD refinement~\citep{zhao2026causalplus,li2026distillalign}.
Inspired by these works, we use ODE initialization to adapt the model to both few-step denoising and sparse attention before DMD refinement.

\xhdr{Combining compression and distillation}
TurboDiffusion and LightWan2.2 combine step distillation, sparse attention, and quantization under FP8 ~\citep{zhang2025turbodiffusion,lv2026lightwan}.
FastWan jointly trains sparse attention and few-step generation with    DMD~\citep{fastvideo2025fastwan}.
FastWan-QAD applies quantization-aware fine-tuning and DMD with dense low-bit attention~\citep{fastvideo2026fastwanqad}, while SQuad adapts sub-quadratic attention through flow-matching fine-tuning before DMD2~\citep{karnewar2026squad}.
LongLive-2.0 incorporates NVFP4 into autoregressive training and DMD~\citep{chen2026longlive}.
\method targets ultrafast attention and end-to-end video generation by combining NVFP4 sparse attention, linear quantization, and few-step distillation.

\section{VSQA: Video Sparse Quantized Attention with NVFP4}
\label{sec:method}
\label{sec:method_overview}

\method builds on the coarse-to-fine attention structure of VSA~\citep{zhang2026faster}.
We retain its cube-level routing and gated coarse branch, and apply NVFP4 quantization~\citep{alvarez2025nvfp4} to the token-level sparse attention branch.

\xhdr{Fixed top-$k$ routing}
Given query, key, and value matrices $Q,K,V\in\mathbb R^{N\times d}$ for $N$ video tokens with head dimension $d$, we group spatially and temporally adjacent tokens into $B$ cubes.
Mean pooling within each cube gives the pooled queries, keys, and values $\bar Q,\bar K,\bar V\in\mathbb R^{B\times d}$.
Attention over these pooled representations gives cube-level scores and outputs:
\begin{equation}
    A=\frac{\bar Q\bar K^\top}{\sqrt d},\qquad
    T=\operatorname{softmax}(A),\qquad
    O_c=T\bar V.
    \label{eq:vsqa_coarse_forward}
\end{equation}
For each query cube, we select the $k$ key cubes with the largest scores in the corresponding row of $A$.
At target sparsity $s$, we use $k=\max(1,\lceil(1-s)B\rceil)$.
All query tokens in a cube attend to the same selected key/value cubes, forming a block-sparse pattern without computing full token-level attention scores.
The coarse output $O_c$ also retains information from all cubes for the final output.

\xhdr{Native NVFP4 sparse attention}
We use NVFP4-quantized operands for both $QK^\top$ and $PV$ in the fine branch, while keeping softmax in high precision, following prior FP4 attention methods~\citep{zhang2025sageattention3,zhang2026attn}.
We denote the NVFP4-quantized version of $Q$ by $\widehat Q$, with block scales included, and use the same notation for $K$, $V$, and $P$.
We encode these selections in a token-level mask $M$: $M_{ij}=0$ when key token $j$ belongs to a cube selected for query token $i$, and $M_{ij}=-\infty$ otherwise.
The fine-branch computation is
\begin{equation}
    S=\frac{\widehat Q\widehat K^\top}{\sqrt d}+M,\qquad
    P=\operatorname{softmax}(S),\qquad
    O_f=\widehat P\widehat V.
    \label{eq:vsqa_fine_forward}
\end{equation}
Here $S$ contains the attention scores after masking, $P$ contains the softmax probabilities over the selected tokens, and $\widehat P$ is their NVFP4 approximation used in the value product.
We copy each cube's coarse output to its constituent tokens, and combine it with the fine output:
\begin{equation}
    O=O_f+g\odot O_c,
    \label{eq:vsqa_combined_forward}
\end{equation}
where $g\in\mathbb R^{N\times d}$ is a learned elementwise gate.
Each token therefore receives a quantized sparse-attention output and a gated contribution from the global cube-level attention.

\xhdr{Backward computation}
We keep the top-k sparse selections invariant with forward pass during backward propagation, and train with a straight-through estimator (STE)~\citep{bengio2013estimating}.
Quantizing attention probabilities breaks the identity that allows FlashAttention~\citep{dao2022flashattention} to compute the softmax gradient correction from the forward output.
Following Attn-QAT~\citep{zhang2026attn}, we apply fake quantization, recompute probabilities from the same quantized queries and keys used in forward, and save an auxiliary output $O_f^{\mathrm{aux}}=P\widehat V$ computed with unrounded probabilities for this correction.
Appendix~\ref{app:fake_real_quant} compares videos generated with fake and real quantization, which show no visible difference.

\xhdr{Details of kernel implementation}
\label{sec:method_kernels}
\label{sec:method_inference}
Our inference kernels traverse only the selected cube pairs and use native low-precision matrix multiplication for the fine branch.
On Blackwell, reducing non-matrix operations is important because their throughput scales more slowly than matrix multiplication~\citep{zadouri2026flashattention}.
Pooling and input quantization still process all tokens, so their overhead becomes more prominent as sparsity increases.
We fuse Q/K/V pooling and quantization around a shared input load: the loaded features supply both the unquantized routing inputs and the packed NVFP4 operands, avoiding separate reads for the two operations.

On the GB300, the kernel computes $QK^\top$ in NVFP4 and $PV$ in FP8.
As Attn-QAT reports~\citep{zhang2026attn}, packing FP4 probabilities and transferring their scales add overhead, so an NVFP4 value product is not the faster choice on this GPU.
Appendix~\ref{app:quant_comparisons} compares videos generated with the FP8 and NVFP4 value products and finds no visible difference, as for fake quantization above.
For an NVFP4 value product, the GB300 kernel also reduces the overhead of quantizing the attention probabilities.
A direct implementation exponentiates the scores, finds a scale for each group of probabilities, and rescales the probabilities before packing them.
Because exponentiation preserves ordering, the kernel instead finds each group's maximum in the score domain and folds the rescaling into the exponentiation, and it accumulates the softmax denominator from the same unrounded values.
This removes a separate reciprocal-and-rescaling pass.

\FloatBarrier

\section{ Taming Compression for Stable Distillation}
\label{sec:method_training}
\begin{figure}[t]
    \centering
    \begin{minipage}[c]{0.45\linewidth}
        \centering
        \small
        \setlength{\tabcolsep}{4pt}
        \renewcommand{\arraystretch}{1.15}
        \begin{tabular}{lccc}
        \toprule
        Recipe & Qual. & Sem. & Total \\
        \midrule
        Base model & 83.52 & 64.50 & 79.72 \\
        Direct DMD, FP8 & 85.25 & 70.68 & 82.33 \\
        Direct DMD, NVFP4 & 84.33 & 71.05 & 81.67 \\
        \rowcolor{tableours}
        \textbf{Ours, NVFP4} & \textbf{85.28} & \textbf{72.47} & \textbf{82.72} \\
        \bottomrule
        \end{tabular}
    \end{minipage}\hfill
    \begin{minipage}[c]{0.53\linewidth}
        \centering
        \includegraphics[width=\linewidth]{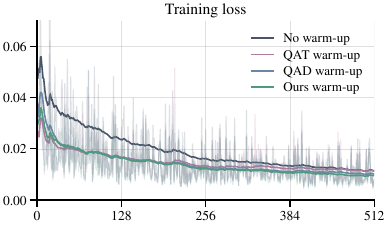}
    \end{minipage}
    \caption{\textbf{The NVFP4 gap and its warmup.}
    Left: VBench scores on Wan2.1-1.3B under different training recipes.
    Right: subsequent training loss at $90\%$ sparsity with different warmup objectives or no warmup.}
    \label{fig:training_warmup_loss}
\end{figure}

To push video generation efficiency to the extreme, building upon \method, we further extend NVFP4 quantization to the linear layers and reduce the number of sampling steps through step distillation.
However, each form of compression perturbs the original model, and jointly adapting to these perturbations can make training less stable.
Prior few-step distillation works have shown~\citep{yin2025slow,zhu2026causal,zhao2026causalplus,li2026distillalign,karnewar2026squad,cheng2025structure} that DMD typically relies on an initialization with basic few-step generation capability.
Our empirical study indicates that a well-prepared initialization supports more stable DMD training in this compressed setting.
In this compressed setting, a suitable initialization should absorb quantization error, establish basic few-step generation capability, and remain close to the original teacher distribution.

\subsection{Overcoming the NVFP4 Quantization Gap}
\label{sec:exp_sparsity_quantization}

Aggressive 4-bit quantization first introduces a pronounced low-bit initialization gap.
Compared with milder 8-bit formats such as FP8, prior work has shown~\citep{huang2026qvgen,li2026dvd,li2024svdquant,wang2025quest,liu2025dilatequant} that full 4-bit weights and activations are more sensitive to limited activation ranges, outliers, and timestep-dependent activation variations in diffusion models, resulting in larger quantization errors.
Figure~\ref{fig:training_warmup_loss} (left) illustrates the challenge: direct DMD with NVFP4 attains lower Quality and Total scores than the FP8 run, motivating additional adaptation before distillation.
To achieve extreme inference acceleration, we don't want to apply any additional compensation or techniques to the quantization process, which results in a larger initial perturbation and makes DMD training less stable.
We therefore need to adapt to the dominant low-bit quantization errors before entering few-step distillation.

\begin{takeawaybox}
NVFP4 introduces a large quantization perturbation that calls for a dedicated warmup before few-step distillation.
\end{takeawaybox}

\subsection{Why QAT and QAD Are Not the Best Choice}
\label{sec:exp_initialization}
\label{sec:method_reconstruction}

A natural choice for quantization error recovery is quantization-aware training (QAT)~\citep{fastvideo2026fastwanqad, sui2024bitsfusion} or quantization-aware distillation (QAD)~\citep{polino2018model,mishra2018apprentice,he2024efficientdm}, which can reduce quantization errors and recover standalone generation quality through finetuning or velocity distillation.
For QAT, we consider fine-tuning the quantized student $v_\theta$ on real videos with the flow-matching objective of the base model~\citep{wan2025wan}:
\begin{equation}
    \mathcal L_{\mathrm{FT}}(\theta)
    = \mathbb E_{x_0,\epsilon,t,c}\left[
        \left\|v_\theta(x_t,t,c)-(\epsilon-x_0)\right\|_2^2
    \right],
    \qquad x_t=(1-t)x_0+t\epsilon,
    \label{eq:vsqa_ft}
\end{equation}
where $x_0$ is a clean video latent, $\epsilon\sim\mathcal N(0,I)$, $t\in[0,1]$ is the noise level, and $c$ is the condition.
QAD instead matches the velocity prediction of the frozen unquantized teacher $v_{\mathrm T}$ on the same input:
\begin{equation}
    \mathcal L_{\mathrm{VD}}(\theta)
    = \mathbb E_{x_t,t,c}\left[
        \left\|v_\theta(x_t,t,c)-v_{\mathrm T}(x_t,t,c)\right\|_2^2
    \right].
    \label{eq:vsqa_vd}
\end{equation}
Both objectives supervise only the final velocity prediction.

\begin{figure}[tb]
    \centering
    \includegraphics[width=\linewidth]{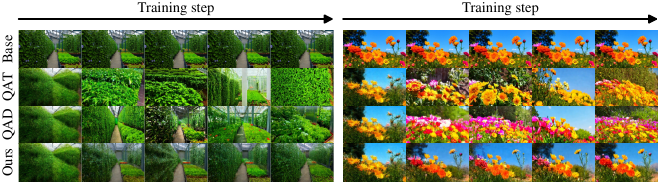}
    \caption{\textbf{Generations during quantization warmup.}
    Matched frames for two prompts, comparing the BF16 base model with QAT, QAD, and layer reconstruction. Training progresses from left to right.}
    \label{fig:training_warmup_frames}
\end{figure}

However, we find that the final-output objectives used by QAT or QAD are not necessarily well suited for DMD initialization.
As shown in Figure~\ref{fig:training_warmup_frames}, QAT and QAD recover the standalone generation quality of the quantized model, but the generated videos gradually deviate from the teacher in global layout and fine-grained content during warmup.
Such deviations introduce additional teacher-student mismatch that may complicate subsequent DMD optimization.

\begin{takeawaybox}
Recovering standalone quality with QAT or QAD does not ensure a suitable DMD initialization, because teacher alignment can deteriorate.
\end{takeawaybox}

Inspired by reconstruction-based quantization methods~\citep{nagel2020up,li2021brecq,hubara2021accurate}, which constrain quantization-induced deviations by reconstructing intermediate outputs at the layer or block level, we instead adopt fine-grained reconstruction supervision during quantization warmup.
As illustrated in Figure~\ref{fig:training_warmup_frames}, layer reconstruction substantially better preserves the layout and content of the original teacher generation throughout training.

This stronger teacher alignment comes at the cost of somewhat lower standalone generation quality than QAD. Since DMD serves as a refinement of an initialized few-step generator~\citep{zhao2026causalplus}, we prioritize preserving teacher alignment while reducing disruptive quantization errors to provide a more stable starting point for subsequent quality refinement.

Following this principle, we introduce deeply supervised layer-wise reconstruction into our quantization warmup to reduce the additional deviation introduced during low-bit adaptation.
With attention kept dense, the quantized and unquantized versions of each supervised layer receive the same teacher input $h_{\mathrm T}^{(\ell)}$, and the warmup minimizes the reconstruction error over the supervised layers $\mathcal R$:
\begin{equation}
    \mathcal L_{\mathrm{rec}}(\theta)
    =\mathbb E_{x_t,t,c}\left[
        \sum_{\ell\in\mathcal R}
        \left\|f_\theta^{(\ell)}\bigl(\operatorname{sg}(h_{\mathrm T}^{(\ell)});t,c\bigr)
        -\operatorname{sg}\bigl(f_{\mathrm T}^{(\ell)}(h_{\mathrm T}^{(\ell)};t,c)\bigr)\right\|_F^2
    \right],
    \label{eq:vsqa_layer_rec}
\end{equation}
where $f_\theta^{(\ell)}$ and $f_{\mathrm T}^{(\ell)}$ denote the student and teacher layers, and $\operatorname{sg}$ stops gradients through the frozen teacher.
Shared reference inputs isolate local quantization error from deviations accumulated in preceding layers.

\begin{figure}[tb]
    \centering
    \includegraphics[width=\linewidth]{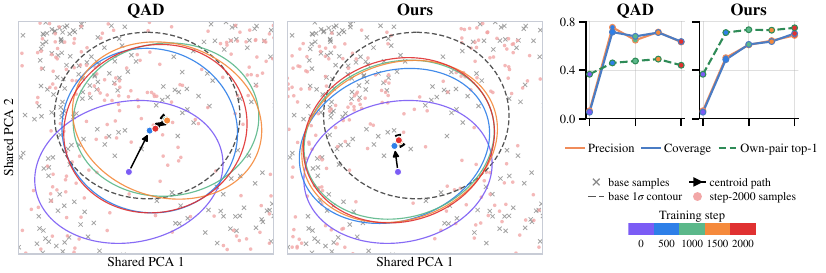}
    \caption{\textbf{V-JEPA feature alignment during quantization warmup.}
    Left: PCA projections relative to the base model.
    Right: set-level alignment (precision and coverage) and sample-wise correspondence (own-pair top-1).}
    \label{fig:training_warmup_drift}
\end{figure}

We analyze teacher alignment during warmup using V-JEPA 2.1~\citep{mur2026v} features.
Figure~\ref{fig:training_warmup_drift} shows that layer reconstruction settles into a stable offset from the teacher distribution while retaining stronger sample-wise correspondence.
We interpret this residual shift as reflecting the standalone quality trade-off observed above, leaving room for subsequent DMD quality refinement.
In contrast, QAD with velocity supervision exhibits fluctuating alignment across checkpoints, which may make subsequent DMD refinement less stable.
This distinction favors a stable, teacher-aligned initialization with a remaining quality gap over a fluctuating warmup trajectory.

\begin{figure}[tb]
    \centering
    \includegraphics[width=\linewidth]{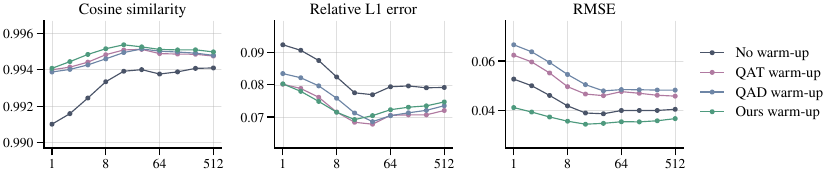}
    \caption{\textbf{Quantization error during subsequent sparse adaptation.}
    Cosine similarity, relative L1 error, and RMSE with different warmup objectives or no warmup, evaluated at the training sparsity of $90\%$.}
    \label{fig:training_warmup_adaptation}
\end{figure}

We further evaluate whether these warmup choices improve subsequent training.
Figure~\ref{fig:training_warmup_adaptation} shows that layer reconstruction warmup yields the lowest RMSE throughout subsequent sparse adaptation and improves cosine similarity over training without warmup.
QAT and QAD also improve cosine similarity and relative L1 error, but retain higher RMSE, indicating that these gains do not necessarily reduce absolute output deviations.
Figure~\ref{fig:training_warmup_loss} (right) further shows lower training loss with reconstruction warmup than without warmup, particularly early in adaptation.
Together, these results support reducing quantization error before introducing sparsity to ease subsequent optimization.

\begin{takeawaybox}
Layer reconstruction preserves teacher alignment while reducing quantization error, supporting more stable subsequent distillation.
\end{takeawaybox}

Meanwhile, following the ODE initialization scheme of~\citep{yin2025slow}, we construct trajectory pairs corresponding to the target few-step schedule from the teacher PF-ODE trajectory.
The frozen teacher integrates
\begin{equation}
    \frac{d x_t^{\mathrm T}}{dt}=v_{\mathrm T}(x_t^{\mathrm T},t,c),
    \qquad x_1^{\mathrm T}\sim\mathcal N(0,I),
    \label{eq:vsqa_teacher_ode}
\end{equation}
from Gaussian noise at $t=1$ to a clean latent $x_0^{\mathrm T}$ at $t=0$.
At the three student timesteps $t_1>t_2>t_3>0$, we retain the teacher states and their shared clean endpoint:
\begin{equation}
    \mathcal D_{\mathrm{ODE}}
    =\left\{(c,x_{t_k}^{\mathrm T},t_k,x_0^{\mathrm T})\right\}_{k=1}^{3},
    \label{eq:vsqa_ode_pairs}
\end{equation}
collected over teacher trajectories for different conditions and initial noise samples.
Specifically, we extract the corresponding target states from the teacher PF-ODE trajectory and use these states for quantization warmup, instead of randomly sampling timesteps from the entire diffusion trajectory.
This design better aligns quantization warmup with subsequent few-step training.
In particular, quantization error recovery is concentrated on the trajectory states that the final generator actually visits during sampling, rather than being distributed across a large number of timesteps that are never directly encountered by the deployed few-step generator.

\FloatBarrier

\subsection{Anchoring Sparse Initialization to the Dense Teacher}
\label{sec:init_ode}

After quantization warmup, we continue along the same teacher PF-ODE trajectory and perform standard few-step ODE initialization.
Specifically, we train the student at the target sampling timesteps while enabling the sparse attention used for final deployment, allowing the model to acquire basic few-step generation capability while simultaneously adapting to sparse computation.
Let $F_\theta(x_t,t,c)$ denote the student's clean-latent prediction.
The teacher endpoints provide the supervision
\begin{equation}
    \mathcal L_{\mathrm{ODE}}(\theta)
    =\mathbb E_{(c,x_t^{\mathrm T},t,x_0^{\mathrm T})\sim\mathcal D_{\mathrm{ODE}}}
    \left[
        \frac{1}{2}\left\|F_\theta(x_t^{\mathrm T},t,c)-x_0^{\mathrm T}\right\|_2^2
    \right].
    \label{eq:vsqa_init}
\end{equation}

However, we further observe that directly applying standard ODE initialization to the sparse student introduces an additional sparsity-induced structural gap.
Although the ODE objective enables the student to learn the target few-step mapping, sparse attention can cause its intermediate representations to deviate further from those of the dense teacher, resulting in additional teacher-student mismatch during the early stage of subsequent DMD training.

\begin{takeawaybox}
Sparse adaptation can leave a structural gap that endpoint supervision alone does not close, motivating layer reconstruction regularization.
\end{takeawaybox}

To mitigate this gap, we extend the fine-grained teacher-alignment constraint introduced during quantization warmup and add a lightweight reconstruction regularizer on top of the standard ODE initialization objective.
Specifically, we constrain the sparse attention outputs to match their corresponding dense-teacher outputs, allowing the student to acquire sparse few-step generation capability while maintaining better teacher alignment.
We optimize
\begin{equation}
    \mathcal L_{\mathrm{init}}(\theta)
    =\frac{1}{1+\lambda}\mathcal L_{\mathrm{ODE}}(\theta)
    +\frac{\lambda}{1+\lambda}\mathcal L_{\mathrm{rec}}(\theta),
    \qquad\lambda\geq 0,
    \label{eq:vsqa_sparse_init}
\end{equation}
where $\mathcal L_{\mathrm{rec}}$ now supervises the quantized sparse attention layers on the shared reference inputs defined in Equation~\ref{eq:vsqa_layer_rec}, with $\mathcal R$ restricted to these layers.
The weight $\lambda$ controls the contribution of reconstruction relative to endpoint supervision.
This regularization provides a better-aligned starting point for subsequent DMD training.

\subsection{Distribution Matching Done Right}
\label{sec:recipe_dmd}

We use DMD as a refinement~\citep{zhao2026causalplus} of our initialized few-step student.
Let $p_\theta(\cdot\mid c)$ and $p_{\mathrm T}(\cdot\mid c)$ denote the student's and teacher's conditional generation distributions.
DMD~\citep{yin2024one,yin2024improved} minimizes the reverse-KL objective
\begin{equation}
    \mathcal L_{\mathrm{DMD}}(\theta)
    =\mathbb E_c\left[
        D_{\mathrm{KL}}\!\left(
            p_\theta(\cdot\mid c)\,\|\,p_{\mathrm T}(\cdot\mid c)
        \right)
    \right],
    \label{eq:vsqa_dmd}
\end{equation}
with gradients approximated using teacher and student-distribution score estimates on noised samples.
Our initialization supports more stable DMD refinement to improve generation quality, with quantization and sparsity enabled throughout training.
Figure~\ref{fig:training_warmup_loss} (left) shows that the complete recipe achieves higher Quality, Semantic, and Total VBench scores than both the base model and direct NVFP4 DMD, demonstrating strong final generation quality despite aggressive compression.
Table~\ref{tab:main_results} further shows that, reconstruction warmup achieves higher scores on all three VBench metrics after DMD than QAD warmup, supporting its use as an initialization for subsequent refinement.
\FloatBarrier

\FloatBarrier
\section{Experiments}
\label{sec:experiments}

\begin{table}[!tb]
    \centering
    \caption{\textbf{VBench results}~\citep{huang2024vbench} over 946 prompts and 16 dimensions, pooled over five seeds with the original Wan VAE and no prompt extension. JetDiffusion and its training ablations use three steps, $90\%$ sparsity, and NVFP4 attention and linear layers. Training-ablation rows share JetDiffusion-1.3B's video size and step count. Bold/underline mark the best/second-best students per size. Evaluation settings and multi-seed results are detailed in Appendices~\ref{app:evaluation_protocol} and~\ref{app:seed_statistics}.}
    \label{tab:main_results}
    \small
    \setlength{\tabcolsep}{3pt}
    \renewcommand{\arraystretch}{1.08}
    \begin{tabularx}{\linewidth}{@{}Xccccc@{}}
    \toprule
    Method & \makecell{Frames $\times$ H $\times$ W} & Steps & Quality $\uparrow$ & Semantic $\uparrow$ & Total $\uparrow$ \\
    \midrule
    \textit{Wan2.1-1.3B base} & $61\times448\times896$ & 50 & 83.52 & 64.50 & 79.72 \\
    TurboDiffusion-1.3B & $81\times480\times832$ & 4 & 83.94 & 68.57 & 80.86 \\
    FastWan-1.3B & $81\times480\times848$ & 3 & \underline{85.14} & 66.14 & 81.34 \\
    FastWan-QAD-1.3B & $81\times480\times832$ & 3 & 81.62 & 62.89 & 77.87 \\
    Direct DMD & $61\times448\times896$ & 3 & 84.33 & 71.05 & 81.67 \\
    \rowcolor{tableours}
    \textbf{Ours (JetDiffusion)} & $61\times448\times896$ & 3 & \textbf{85.28} & \underline{72.47} & \textbf{82.72} \\
    \quad QAD warmup & & & 85.11 & 70.54 & 82.19 \\
    \quad w/o ODE init & & & 84.73 & \textbf{73.54} & \underline{82.49} \\
    \quad w/o ODE init, QAD warmup & & & 84.51 & 69.71 & 81.55 \\
    \quad w/o reconstruction regularization & & & 84.83 & 70.33 & 81.93 \\
    \midrule
    \textit{Wan2.1-14B base} & $77\times768\times1280$ & 50 & 83.37 & 75.63 & 81.82 \\
    TurboDiffusion-14B & $81\times720\times1280$ & 4 & 83.51 & 73.82 & 81.57 \\
    LightWan2.2-A14B & $81\times720\times1280$ & 4 & \underline{83.95} & \underline{78.78} & \underline{82.92} \\
    \rowcolor{tableours}
    \textbf{Ours (JetDiffusion)} & $77\times768\times1280$ & 3 & \textbf{84.99} & \textbf{80.32} & \textbf{84.06} \\
    \bottomrule
    \end{tabularx}
\end{table}

\subsection{Setup}
\label{sec:exp_setup}

\xhdr{Training}
We train Wan2.1-1.3B at 480p ($61\times448\times896$) and Wan2.1-14B at 720p ($77\times768\times1280$)~\citep{wan2025wan}, with dimensions given as frames $\times$ height $\times$ width.
JetDiffusion and its training ablations use three denoising steps, $90\%$ sparsity, and NVFP4 attention and linear layers.
For native NVFP4 computation, we use $4\times4\times8$ and $4\times8\times8$ cubes at 480p and 720p, respectively.
Training uses only text prompts from Wan-Syn~\citep{zhang2026faster} and VidProM~\citep{wang2024vidprom}.

\textbf{Evaluation.}
We use the original VBench prompts without prompt extension and the original Wan VAE for quality comparisons, including TurboDiffusion, LightWan2.2, and FastWan/FastWan-QAD.
Results are pooled over five generation seeds for both model sizes.
The scoring protocol and sampling settings are detailed in Appendix~\ref{app:evaluation_protocol}.

\xhdr{Efficiency}
We measure attention and total DiT latency on RTX 5090 at both resolutions and GB300 at 720p, using FP8 $PV$ on GB300 for higher throughput.
Dense baselines are FlashAttention-2~\citep{dao2024flashattention} on RTX 5090 and FlashAttention-4~\citep{zadouri2026flashattention} on GB300.
RTX 5090 comparisons also include SageAttention2++~\citep{zhang2025sageattention2++}, SageAttention3~\citep{zhang2025sageattention3}, and SageSLA~\citep{zhang2025turbodiffusion}.
All sparse methods use $90\%$ sparsity.
For timing comparisons, we use TAEHV~\citep{BoerBohan2025TAEHV} to match the decoder used by other accelerated systems. Decoder evaluations with each model and its sampling settings fixed are detailed in Appendix~\ref{app:decoder_quality}.

\subsection{Generation Quality}
\label{sec:exp_main}

Table~\ref{tab:main_results} lists the VBench scores.
JetDiffusion runs three denoising steps with $90\%$ attention sparsity and NVFP4 attention and linear layers, and on both model sizes its scores stay at the level of the base model.
JetDiffusion achieves the highest VBench Total among the compared methods at both 1.3B and 14B.
Our full staged recipe achieves the highest Total among the training variants, improving over direct DMD by 1.05 points.
Removing ODE initialization or reconstruction regularization lowers Total, as does replacing the layerwise reconstruction warmup with QAD.
Multi-seed results are detailed in Appendix~\ref{app:seed_statistics}.

\FloatBarrier
\subsection{Inference Efficiency}
\label{sec:exp_inference}

\suppressfloats[t]
\begin{figure}[!htb]
    \centering
    \includegraphics[width=\linewidth]{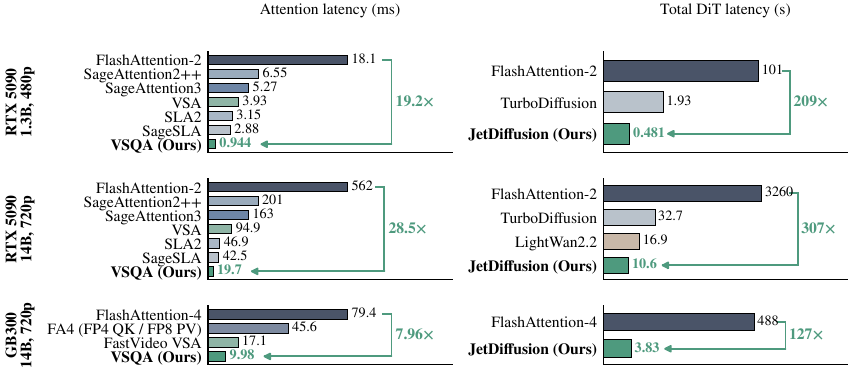}
    \caption{\textbf{Attention and total DiT latency on Blackwell GPUs.}
    Left: self-attention. Right: total DiT time per video at each method's default sampling steps (base: 50 steps with CFG).
    Annotations show speedups over the dense baseline on the same GPU.
}
    \label{fig:inference_speedup}
\end{figure}

Figure~\ref{fig:inference_speedup} shows that \method achieves $8.0$--$28.5\times$ attention speedups over the respective dense FlashAttention baselines across the evaluated Blackwell configurations.
On the RTX 5090, it also outperforms SageSLA by $2.15$--$3.06\times$ at the same $90\%$ sparsity.

Combined with linear quantization and few-step distillation, JetDiffusion reduces total DiT time per generation by $127$--$307\times$ relative to the base model.
On the RTX 5090, it is $3.08$--$4.01\times$ faster than TurboDiffusion and $1.59\times$ faster than LightWan2.2 at 720p.
End-to-end speedups with TAEHV are $273\times$ on RTX 5090 and $119\times$ on GB300. Timing protocols and component analyses are detailed in Appendix~\ref{app:inference_latencies} and Appendices~\ref{app:controlled_attention}--\ref{app:gb300_aligned}.

\textbf{Contributions of sparsity and quantization.}
Table~\ref{tab:controlled_attention} compares four attention variants on identical inputs within each hardware/model group. Combining sparsity and quantization gives larger speedups than either alone across all three configurations. Experimental settings and the DiT-step ablation are detailed in Appendix~\ref{app:controlled_attention}.

\textbf{Pooling--quantization fusion.}
On Wan2.1-14B at $90\%$ sparsity on GB300, fusing pooling and quantization reduces preparation latency from 1.20 to 0.73 ms ($1.64\times$) and full-attention latency from 10.46 to 9.98 ms ($1.05\times$), with identical inputs, selected cubes, and fine-attention kernel. The cumulative implementation ablation and stage profiling are detailed in Appendix~\ref{app:kernel_profiling}.

\subsection{Component Quality Ablation}
\label{sec:component_quality}

We train a separate 1.3B model for each attention and linear-layer precision, using one training run and five generation seeds per configuration (Table~\ref{tab:retrained_components}). All sparse variants use $90\%$ sparsity; evaluation uses three-step generation and Wan VAE. With staged training, the fully compressed model achieves 82.72 Total, comparable to the dense BF16 control (82.69) and 1.05 points above direct NVFP4 DMD.

\begin{table}[!htb]
    \centering
    \begin{minipage}[t]{0.42\linewidth}
        \vspace{0pt}
        \caption{Attention latency (ms) and speedup over dense BF16 at $90\%$ sparsity.}
        \label{tab:controlled_attention}
        \centering
        \small
        \setlength{\tabcolsep}{2pt}
        \renewcommand{\arraystretch}{1.10}
        \begin{tabularx}{\linewidth}{@{}l*{3}{>{\centering\arraybackslash}X}@{}}
            \toprule
            & \multicolumn{2}{c}{RTX 5090} & GB300 \\
            \cmidrule(lr){2-3}\cmidrule(l){4-4}
            Attention & 1.3B & 14B & 14B \\
            \midrule
            Dense BF16 & \makecell{17.59\\(1.0$\times$)} & \makecell{546.95\\(1.0$\times$)} & \makecell{79.44\\(1.0$\times$)} \\
            Dense low-bit & \makecell{4.86\\(3.6$\times$)} & \makecell{149.83\\(3.7$\times$)} & \makecell{45.58\\(1.7$\times$)} \\
            Sparse BF16 & \makecell{3.53\\(5.0$\times$)} & \makecell{87.72\\(6.2$\times$)} & \makecell{13.88\\(5.7$\times$)} \\
            \rowcolor{tableours}
            \textbf{VSQA} & \makecell{\textbf{0.855}\\\textbf{(20.6$\times$)}} & \makecell{\textbf{18.29}\\\textbf{(29.9$\times$)}} & \makecell{\textbf{9.98}\\\textbf{(8.0$\times$)}} \\
            \bottomrule
        \end{tabularx}
    \end{minipage}%
    \iflatexml
        \end{table}
        \begin{table}[!htb]\centering
    \else\hfill\fi
    \begin{minipage}[t]{0.56\linewidth}
        \vspace{0pt}
        \caption{Component quality after separate training on Wan2.1-1.3B. All sparse variants use $90\%$ sparsity.}
        \label{tab:retrained_components}
        \centering
        \small
        \setlength{\tabcolsep}{2pt}
        \renewcommand{\arraystretch}{1.12}
        \begin{tabular}{@{}llcccc@{}}
            \toprule
            Training & Attention & Linears & Qual. & Sem. & Total \\
            \midrule
            \multirow{6}{*}{\makecell[l]{Direct\\DMD}}
                & Dense BF16 & BF16 & 85.98 & 69.56 & 82.69 \\
                & Sparse BF16 & BF16 & 85.61 & 72.10 & 82.91 \\
                & Sparse NVFP4 & BF16 & 84.71 & 72.75 & 82.32 \\
                & Dense BF16 & NVFP4 & 84.80 & 70.59 & 81.96 \\
                & Sparse FP8 & FP8 & 85.25 & 70.68 & 82.33 \\
                & Sparse NVFP4 & NVFP4 & 84.33 & 71.05 & 81.67 \\
            \midrule
            \rowcolor{tableours}
            \makecell[l]{Staged\\recipe} & Sparse NVFP4 & NVFP4 & 85.28 & 72.47 & 82.72 \\
            \bottomrule
        \end{tabular}
    \end{minipage}
\end{table}
\FloatBarrier

\section{Conclusion}
\label{sec:conclusion}

We presented \method, a trainable sparse attention method that combines high sparsity with native NVFP4 computation for video diffusion.
\method achieves a $8.0\times$ attention speedup over dense FlashAttention-4 at $90\%$ sparsity on NVIDIA GB300 while preserving generation quality.
Building on \method, we further presented \textbf{JetDiffusion}, which combines sparse NVFP4 attention with NVFP4 linear layers and few-step distillation to accelerate the full video generation pipeline.
Together, these techniques enable $273\times$ end-to-end acceleration on Wan2.1-14B on the RTX 5090, demonstrating the potential of jointly exploiting sparsity, low precision, and generation-step reduction for efficient video generation.

\section*{AI Use Statement}

We used generative AI tools to assist with polishing the manuscript. The authors take responsibility for the final content of this work, including its text, claims, and artifacts produced with AI assistance.

\section*{Ethics Statement}

This work improves video generation efficiency by combining sparse attention, low-bit quantization, and few-step distillation.
Reducing inference cost may make video generation more accessible to users with limited computational resources.
However, faster generation may also facilitate the creation of misleading or harmful synthetic content, and compression may alter the behavior of the base model, including its safety mechanisms.
Deployment should therefore include safety evaluations of the compressed models, appropriate content safeguards, and clear disclosure of AI-generated videos.

\section*{Reproducibility Statement}

We describe the sparse attention formulation and inference kernels in Section~\ref{sec:method}, and the training objectives and initialization procedure in Section~\ref{sec:method_training}.
Section~\ref{sec:experiments} specifies the base models, prompt sources, generation settings, hardware, and comparison methods used in our experiments.
Appendix~\ref{app:experimental_details} reports per-dimension quality scores and detailed inference timings.

\begingroup
\interlinepenalty=10000

\endgroup

\appendix
\clearpage
\section{Additional Experimental Results}
\label{app:experimental_details}
\subsection{Training Setup}
\label{app:training_setup}
Each student is trained in three stages from the base model.
The quantization warmup trains the dense NVFP4 student with the layer reconstruction loss on cached teacher states.
The sparse initialization then trains the sparse student at the three student timesteps on cached teacher PF-ODE trajectories (48 teacher steps, flow shift 8), with the ODE endpoint loss and the reconstruction regularizer weighted equally ($\lambda=1$).
DMD refines the result.
Every warmup stage uses an effective batch size of 16 and a learning rate of $10^{-5}$, and we take its checkpoint after 500 updates.
Table~\ref{tab:dmd_hparams} lists the DMD hyperparameters.
All DMD runs in Table~\ref{tab:main_results} had converged when training ended, and only then were their checkpoints evaluated.

\begin{table}[!htb]
    \centering
    \caption{\textbf{DMD hyperparameters.}}
    \label{tab:dmd_hparams}
    \small
    \renewcommand{\arraystretch}{1.15}
    \begin{tabular}{ccc}
    \toprule
    Hyperparameters & Wan2.1-1.3B & Wan2.1-14B \\
    \midrule
    Real score network & Wan2.1-T2V-1.3B & Wan2.1-T2V-14B \\
    Real score CFG weight & 3.5 & 6.0 \\
    Critic network initialization & Wan2.1-T2V-1.3B & Wan2.1-T2V-14B \\
    Student timesteps & 1000, 941, 800 & 1000, 941, 800 \\
    Batch size & 16 & 16 \\
    Learning rate ($G_\theta$) & 2e-6 & 2e-6 \\
    Learning rate ($f_\psi$) & 4e-7 & 4e-7 \\
    Generator/critic update ratio & 1:5 & 1:1 \\
    EMA decay & none & none \\
    Training iterations & 1000 & 200 \\
    Prompt set & Wan-Syn & VidProM \\
    \bottomrule
    \end{tabular}
\end{table}

\subsection{Per-Dimension VBench Results}
\label{app:vbench_dimensions}
Tables~\ref{tab:vbench_dims_13b} and~\ref{tab:vbench_dims_14b} report per-dimension scores for the 1.3B training variants and the five-seed 14B student.

\begin{table}[!htb]
    \centering
    \caption{\textbf{Per-dimension VBench scores on Wan2.1-1.3B at 480p.} Ours is JetDiffusion. Bold and underline mark the best and second-best students.}
    \label{tab:vbench_dims_13b}
    \footnotesize
    \setlength{\tabcolsep}{4pt}
    \renewcommand{\arraystretch}{1.1}
    \begin{tabular}{lcc>{\columncolor{tableours}}ccccc}
    \toprule
    VBench dimension & \makecell{Base\\model} & \makecell{Direct\\DMD} & \textbf{Ours} & \makecell{QAD\\warmup} & \makecell{w/o ODE\\init.} & \makecell{w/o ODE init.,\\QAD warmup} & \makecell{w/o recon.\\reg.} \\
    \midrule
    Subject consistency & 91.93 & 95.81 & 95.74 & \underline{95.86} & 95.77 & \textbf{96.75} & 94.74 \\
    Background consistency & 94.13 & 93.40 & 93.80 & \underline{94.37} & 93.85 & \textbf{95.73} & 93.12 \\
    Temporal flickering & 99.25 & 98.91 & \underline{99.25} & 99.10 & 99.23 & \textbf{99.32} & \textbf{99.32} \\
    Motion smoothness & 96.83 & 98.46 & 98.21 & 98.32 & \underline{98.65} & \textbf{98.80} & 97.75 \\
    Aesthetic quality & 60.45 & 60.30 & 61.23 & 61.55 & \underline{62.09} & \textbf{62.15} & 60.79 \\
    Imaging quality & 64.94 & 65.98 & \underline{66.74} & 65.88 & 65.55 & 65.62 & \textbf{67.43} \\
    Dynamic degree & 93.89 & 86.11 & \underline{94.17} & 91.11 & 84.44 & 72.50 & \textbf{94.72} \\
    \textit{Quality} & 83.52 & 84.33 & \textbf{85.28} & \underline{85.11} & 84.73 & 84.51 & 84.83 \\
    \midrule
    Object class & 66.65 & 78.23 & \textbf{81.41} & 76.93 & \underline{80.89} & 73.13 & 79.59 \\
    Multiple objects & 41.48 & 58.87 & \underline{63.14} & 59.25 & \textbf{67.94} & 61.69 & 58.70 \\
    Human action & 71.40 & 70.60 & \underline{74.00} & \textbf{75.80} & 71.80 & 72.80 & 70.80 \\
    Color & 72.02 & \underline{82.95} & 79.58 & \textbf{85.40} & 81.02 & 79.89 & 80.78 \\
    Spatial relationship & 65.95 & 75.81 & \underline{77.87} & 73.64 & \textbf{81.21} & 74.72 & 73.85 \\
    Scene & 29.53 & \underline{38.60} & 37.21 & 31.86 & \textbf{39.77} & 34.19 & 33.26 \\
    Appearance style & 28.73 & 28.37 & \textbf{28.69} & 28.05 & \underline{28.52} & 28.09 & 28.18 \\
    Temporal style & 22.78 & 23.07 & \textbf{23.97} & 22.83 & 23.88 & 22.64 & \underline{23.94} \\
    Overall consistency & 23.26 & 23.04 & \underline{23.52} & 23.32 & \textbf{23.71} & 22.94 & 23.43 \\
    \textit{Semantic} & 64.50 & 71.05 & \underline{72.47} & 70.54 & \textbf{73.54} & 69.71 & 70.33 \\
    \midrule
    \textbf{Total} & 79.72 & 81.67 & \textbf{82.72} & 82.19 & \underline{82.49} & 81.55 & 81.93 \\
    \bottomrule
    \end{tabular}
\end{table}

\begin{table}[!htb]
\centering
\caption{Five-seed pooled per-dimension VBench scores of the 14B models. Each model uses its setting in Table~\ref{tab:main_results}: the Wan2.1-14B base and JetDiffusion-14B are generated at $77\times768\times1280$ with 50 and 3 steps, TurboDiffusion-14B and LightWan2.2-A14B at $81\times720\times1280$ with 4 steps.}
\label{tab:vbench_dims_14b}
\small
\begin{tabular}{lcccc}
\toprule
VBench dimension & \makecell{Wan2.1-14B\\base} & \makecell{TurboDiffusion\\-14B} & \makecell{LightWan2.2\\-A14B} & \makecell{JetDiffusion\\-14B} \\
\midrule
Subject consistency & 95.89 & 96.12 & 95.68 & 95.48 \\
Background consistency & 97.80 & 93.97 & 96.60 & 95.51 \\
Temporal flickering & 99.52 & 98.50 & 98.93 & 99.04 \\
Motion smoothness & 98.69 & 99.01 & 98.10 & 97.63 \\
Aesthetic quality & 65.07 & 61.40 & 66.28 & 66.41 \\
Imaging quality & 68.43 & 72.07 & 71.24 & 69.33 \\
Dynamic degree & 42.22 & 57.22 & 52.78 & 75.83 \\
Object class & 83.29 & 83.70 & 80.51 & 89.46 \\
Multiple objects & 65.61 & 62.35 & 79.45 & 75.47 \\
Human action & 79.20 & 79.00 & 84.80 & 83.40 \\
Color & 74.87 & 78.86 & 70.79 & 84.72 \\
Spatial relationship & 76.06 & 77.99 & 84.22 & 79.96 \\
Scene & 48.14 & 43.49 & 54.88 & 53.95 \\
Appearance style & 31.55 & 29.38 & 30.35 & 30.47 \\
Temporal style & 23.38 & 22.18 & 24.24 & 24.34 \\
Overall consistency & 24.87 & 23.93 & 25.33 & 25.72 \\
\midrule
Quality & 83.37 & 83.51 & 83.95 & 84.99 \\
Semantic & 75.63 & 73.82 & 78.78 & 80.32 \\
Total & 81.82 & 81.57 & 82.92 & 84.06 \\
\bottomrule
\end{tabular}\end{table}
\FloatBarrier

\FloatBarrier
\subsection{Inference Latencies}
\label{app:inference_latencies}
Table~\ref{tab:kernel_timings} reports the multi-method benchmark shown in Figure~\ref{fig:inference_speedup}. The RTX 5090 controlled four-variant experiment in Table~\ref{tab:controlled_attention} is a separate benchmark with settings detailed in Appendix~\ref{app:controlled_attention}; its timings should not be combined with the multi-method baseline rows. On GB300, the VSA attention entry is the FastVideo VSA configuration with the FA4 CuTe BF16 fine kernel in Table~\ref{tab:fusion_ablation}. Dashes indicate unmeasured DiT timings.
Table~\ref{tab:kernel_timings} gives one self-attention call, one DiT forward pass including the linear layers, and the total DiT time of one generation, which sums the forward passes at each method's default number of sampling steps, with 50 steps and CFG for the base model.
Attention is compared across attention methods, and the DiT times across the base model with these attention methods applied directly and the few-step distilled systems TurboDiffusion~\citep{zhang2025turbodiffusion} and LightWan2.2~\citep{lv2026lightwan}.
Table~\ref{tab:generation_timings} compares one whole generation between the base model and JetDiffusion. The 14B base uses an optimized Wan VAE; the RTX 5090 BF16 reference is estimated, and the GB300 pipeline is measured at 77 frames.
The non-DiT part covers everything else in producing one video, such as text encoding and VAE decoding.
To push end-to-end latency further, JetDiffusion decodes with TAEHV~\citep{BoerBohan2025TAEHV}, a tiny video autoencoder, while the base model keeps its default VAE.
The other systems are left out of this table because their VAE choices differ, which would make end-to-end times incomparable.
480p is 61 frames at $448\times896$ and 720p is 77 frames at $768\times1280$.
JetDiffusion uses \method with 128-token cubes at 480p and 256-token cubes at 720p, NVFP4 linear layers with dynamic scaling, and three denoising steps.
The RTX 5090 VSA benchmark uses its default 64-token cubes. The GB300 comparison uses the matched input and cube layout described in Appendix~\ref{app:controlled_attention}.
SLA2 has no public implementation, so we use our own reproduction; its DiT times are composed from its attention latency.

\begin{table}[!htb]
    \centering
    \caption{\textbf{Attention and DiT timings.}
    One self-attention call (ms), one DiT forward pass (s), and the total DiT time of one generation (s), each followed in parentheses by its speedup over the dense base model of the same group. Sparse methods run at $90\%$ sparsity.}
    \label{tab:kernel_timings}
    \footnotesize
    \setlength{\tabcolsep}{4pt}
    \renewcommand{\arraystretch}{1.1}
    \begin{tabularx}{\linewidth}{>{\raggedright\arraybackslash}Xcccc}
    \toprule
    Method & Steps & Attention (ms) & DiT / step (s) & DiT total (s) \\
    \midrule
    \multicolumn{5}{l}{\textit{RTX 5090, 1.3B, 480p}} \\
    FlashAttention-2 & 50 & 18.1 (1.00$\times$) & 1.00 (1.00$\times$) & 101 (1.00$\times$) \\
    SageAttention2++ & 50 & 6.55 (2.77$\times$) & 0.640 (1.57$\times$) & 64.0 (1.57$\times$) \\
    SageAttention3 & 50 & 5.27 (3.44$\times$) & 0.613 (1.64$\times$) & 61.3 (1.64$\times$) \\
    VSA & 50 & 3.93 (4.61$\times$) & 0.647 (1.55$\times$) & 64.7 (1.56$\times$) \\
    SLA2 & 50 & 3.15 (5.76$\times$) & 0.558 (1.80$\times$) & 55.8 (1.80$\times$) \\
    TurboDiffusion (SageSLA) & 4 & 2.88 (6.28$\times$) & 0.482 (2.08$\times$) & 1.93 (52.2$\times$) \\
    \rowcolor{tableours}
    \textbf{JetDiffusion (\method, Ours)} & 3 & 0.944 (19.2$\times$) & 0.160 (6.27$\times$) & 0.481 (209$\times$) \\
    \midrule
    \multicolumn{5}{l}{\textit{RTX 5090, 14B, 720p}} \\
    FlashAttention-2 & 50 & 562 (1.00$\times$) & 32.6 (1.00$\times$) & 3260 (1.00$\times$) \\
    SageAttention2++ & 50 & 201 (2.79$\times$) & 17.7 (1.85$\times$) & 1770 (1.85$\times$) \\
    SageAttention3 & 50 & 163 (3.45$\times$) & 16.1 (2.03$\times$) & 1610 (2.03$\times$) \\
    VSA & 50 & 94.9 (5.92$\times$) & 15.4 (2.12$\times$) & 1540 (2.12$\times$) \\
    SLA2 & 50 & 46.9 (12.0$\times$) & 12.0 (2.72$\times$) & 1200 (2.72$\times$) \\
    TurboDiffusion (SageSLA) & 4 & 42.5 (13.2$\times$) & 8.17 (3.99$\times$) & 32.7 (99.8$\times$) \\
    LightWan2.2 & 4 & 32.5 (17.3$\times$) & 4.20 (7.77$\times$) & 16.9 (193$\times$) \\
    \rowcolor{tableours}
    \textbf{JetDiffusion (\method, Ours)} & 3 & 19.7 (28.5$\times$) & 3.57 (9.14$\times$) & 10.6 (307$\times$) \\
    \midrule
    \multicolumn{5}{l}{\textit{GB300, 14B, 720p}} \\
    FlashAttention-4 & 50 & 79.44 (1.00$\times$) & 4.88 (1.00$\times$) & 487.60 (1.00$\times$) \\
    FlashAttention-4 (NVFP4 QK / FP8 PV) & 50 & 45.58 (1.74$\times$) & --- & --- \\
    FastVideo VSA (FA4 CuTe BF16) & 50 & 17.06 (4.66$\times$) & --- & --- \\
    \rowcolor{tableours}
    \textbf{JetDiffusion (\method, Ours)} & 3 & 9.98 (8.0$\times$) & 1.28 (3.8$\times$) & 3.83 (127$\times$) \\
    \bottomrule
    \end{tabularx}
\end{table}

\begin{table}[!htb]
    \centering
    \caption{\textbf{Generation timings (s).}
    Total DiT time of one generation, the non-DiT part of the pipeline, and end-to-end time, each followed in parentheses by its speedup over the dense base model of the same group. The 14B base uses optimized Wan VAE and JetDiffusion uses TAEHV. RTX 5090 14B baseline timings are estimated; GB300 timings use 77-frame videos.}
    \label{tab:generation_timings}
    \footnotesize
    \setlength{\tabcolsep}{4pt}
    \renewcommand{\arraystretch}{1.1}
    \begin{tabularx}{\linewidth}{>{\raggedright\arraybackslash}Xcccc}
    \toprule
    Method & Steps & DiT total (s) & Non-DiT (s) & End-to-end (s) \\
    \midrule
    \multicolumn{5}{l}{\textit{RTX 5090, 1.3B, 480p}} \\
    FlashAttention-2 & 50 & 101 (1.00$\times$) & 2.75 (1.00$\times$) & 103 (1.00$\times$) \\
    \rowcolor{tableours}
    \textbf{JetDiffusion (\method, Ours)} & 3 & 0.481 (209$\times$) & 0.39 (7.05$\times$) & 0.866 (119$\times$) \\
    \midrule
    \multicolumn{5}{l}{\textit{RTX 5090, 14B, 720p}} \\
    FlashAttention-2 & 50 & 3172 (1.00$\times$) & 4.904 (1.00$\times$) & 3177 (1.00$\times$) \\
    \rowcolor{tableours}
    \textbf{JetDiffusion (\method, Ours)} & 3 & 10.76 (295$\times$) & 0.879 (5.58$\times$) & 11.64 (273$\times$) \\
    \midrule
    \multicolumn{5}{l}{\textit{GB300, 14B, 720p}} \\
    FlashAttention-4 & 50 & 487.60 (1.00$\times$) & 1.376 (1.00$\times$) & 488.97 (1.00$\times$) \\
    \rowcolor{tableours}
    \textbf{JetDiffusion (\method, Ours)} & 3 & 3.83 (127$\times$) & 0.280 (4.9$\times$) & 4.11 (119$\times$) \\
    \bottomrule
    \end{tabularx}
\end{table}

\FloatBarrier

\FloatBarrier
\subsection{Evaluation Protocol and Baselines}
\label{app:evaluation_protocol}

Our main-table evaluation uses the original VBench prompts without prompt extension and the original Wan VAE. Evaluation uses all 16 VBench dimensions over 946 prompts, commit \texttt{45e79ec}, with AVoCaDO replacing Tag2Text for scene scoring. Videos use 16 fps. Bases use Wan's negative prompt translated into English; JetDiffusion students use none. The 14B base uses shift 5 for both Quality and Semantic. Table~\ref{tab:eval_settings} lists the sampling settings. Main-table scores are pooled over seeds 42--46; Appendix~\ref{app:seed_statistics} separately reports means and sample SDs of the per-seed scores.

\begin{table}[!htb]
\centering
\caption{Sampling settings for the base models and JetDiffusion.}
\label{tab:eval_settings}
\footnotesize
\setlength{\tabcolsep}{3pt}
\renewcommand{\arraystretch}{1.12}
\begin{tabularx}{\linewidth}{@{}Xcccc@{}}
\toprule
Model & Frames $\times$ H $\times$ W & Steps / sampler & CFG & Shift \\
\midrule
Wan2.1-1.3B base & 61 $\times$ 448 $\times$ 896 & 50 / UniPC & 5 & 3 \\
JetDiffusion-1.3B & 61 $\times$ 448 $\times$ 896 & 3 / DMD & 1 & 8 \\
Wan2.1-14B base & 77 $\times$ 768 $\times$ 1280 & 50 / UniPC & 5 & 5 \\
JetDiffusion-14B & 77 $\times$ 768 $\times$ 1280 & 3 / DMD & 1 & 8 \\
\bottomrule
\end{tabularx}
\end{table}

We evaluate TurboDiffusion~\citep{zhang2025turbodiffusion}, LightWan2.2, and FastWan/FastWan-QAD~\citep{fastvideo2025fastwan,fastvideo2026fastwanqad} with the same VBench prompts and original Wan VAE as Table~\ref{tab:main_results}. Results are pooled over five seeds (42--46) for both model sizes. The compared systems retain the frame counts, resolutions, and step budgets listed in the main table, so the comparison does not isolate model changes at a fixed geometry and sampling budget. LightWan2.2-A14B~\citep{lv2026lightwan} is based on Wan2.2-A14B rather than Wan2.1-14B.
FastWan-QAD is evaluated through the upstream GB200/SM100 path with NVFP4 $QK^\top$, BF16 $PV$, and the official inference linear forward. This differs from its RTX 5090 SageAttention3 deployment path; its training and inference activation-scaling implementations also differ. The row therefore characterizes the evaluated implementation, and we do not attribute its quality gap to QAD itself.
For TurboDiffusion, we use our own reproduction with numerical precision settings, including low-bit formats, matched to the original implementation.

\FloatBarrier
\subsection{Training Ablation and Variation Across Seeds}
\label{app:seed_statistics}

To assess how the two initialization stages affect DMD training, we compare different warm-up and ODE initialization settings on Wan2.1-1.3B at fixed sparsity and precision. All six training variants use three-step generation, $90\%$ sparse NVFP4 attention, NVFP4 linears, and Wan VAE. Direct DMD uses neither initialization stage. ``QAD warmup'' in Table~\ref{tab:main_results} replaces layerwise reconstruction with QAD while retaining ODE initialization and reconstruction regularization. ``w/o ODE initialization'' retains only the warm-up stage; its QAD variant replaces that warm-up with QAD. ``w/o reconstruction regularization'' retains layerwise reconstruction warm-up and ODE initialization but removes the regularizer from the latter stage.
Our full staged recipe achieves the highest pooled VBench Total among these variants, improving over direct DMD by 1.05 points (82.72 versus 81.67).

\begin{table}[!htb]
\centering
\caption{Mean and sample SD across five generation seeds (42--46). Recon. denotes reconstruction.}
\label{tab:seed_statistics}
\footnotesize
\setlength{\tabcolsep}{3pt}
\renewcommand{\arraystretch}{1.12}
\begin{tabularx}{\linewidth}{@{}p{2cm}p{2cm}Xccc@{}}
\toprule
Method & Warm-up & ODE initialization & Quality & Semantic & Total \\
\midrule
1.3B base & --- & --- & 83.50 $\pm$ 1.19 & 64.50 $\pm$ 3.38 & 79.70 $\pm$ 1.59 \\
1.3B DMD & None & None & 84.32 $\pm$ 0.21 & 71.05 $\pm$ 1.21 & 81.67 $\pm$ 0.15 \\
 & QAD & None & 84.51 $\pm$ 0.46 & 69.71 $\pm$ 2.51 & 81.55 $\pm$ 0.43 \\
 & Layerwise recon. & None & 84.71 $\pm$ 0.40 & \textbf{73.54 $\pm$ 1.54} & 82.47 $\pm$ 0.22 \\
 & QAD & w/ recon. reg. & 85.08 $\pm$ 0.29 & 70.54 $\pm$ 1.61 & 82.17 $\pm$ 0.48 \\
 & Layerwise recon. & w/o recon. reg. & 84.82 $\pm$ 0.41 & 70.33 $\pm$ 0.94 & 81.92 $\pm$ 0.40 \\
1.3B Ours & Layerwise recon. & w/ recon. reg. & \textbf{85.25 $\pm$ 0.28} & 72.47 $\pm$ 1.74 & \textbf{82.69 $\pm$ 0.42} \\
14B base & --- & --- & 83.37 $\pm$ 1.32 & 75.63 $\pm$ 2.07 & 81.82 $\pm$ 1.21 \\
14B Ours & Layerwise recon. & w/ recon. reg. & \textbf{84.98 $\pm$ 0.40} & \textbf{80.32 $\pm$ 0.91} & \textbf{84.05 $\pm$ 0.21} \\
\bottomrule
\end{tabularx}
\end{table}

SD measures variation across generation seeds for a fixed trained checkpoint and fixed evaluation prompts, not variation across independent training runs. Pooled scores and means of per-seed scores are distinct statistics and can differ slightly; for example, JetDiffusion-1.3B has pooled Total 82.72 and mean Total 82.69. The 14B student's mean Total is 2.23 points higher than the base, with a 95\% confidence interval for the student's mean Total of [83.79, 84.31]. The small 1.3B QAD warm-up gap is not established as statistically significant.

\FloatBarrier
\subsection{Component Quality Ablations}
\label{app:component_quality}

First, we swap components of the final three-step model at inference, without retraining (Table~\ref{tab:inference_swaps}). The reference uses VSQA, NVFP4 linears, and Wan VAE. Each row changes only the named components; the checkpoint, prompts, sampling settings, and remaining components stay fixed. We use five evaluation seeds for 1.3B and seed 42 for 14B.

\begin{table}[!htb]
\centering
\caption{Inference-time component swaps without retraining. Scores are VBench Total.}
\label{tab:inference_swaps}
\footnotesize
\setlength{\tabcolsep}{3pt}
\renewcommand{\arraystretch}{1.12}
\begin{tabularx}{\linewidth}{@{}Xcc@{}}
\toprule
Configuration & 1.3B Total, 5 seeds & 14B Total, seed 42 \\
\midrule
VSQA + NVFP4 linears + Wan VAE & 82.72 & 84.02 \\
Dense BF16 attention & 81.93 & 83.40 \\
BF16 linears & 82.75 & \textbf{84.52} \\
Dense BF16 attention + BF16 linears & 82.06 & 83.63 \\
TAEHV decoder & \textbf{82.78} & 83.78 \\
\bottomrule
\end{tabularx}
\end{table}

Second, we train a separate 1.3B model for each attention and linear-layer precision, with one training run and five evaluation seeds per configuration (Table~\ref{tab:retrained_components}). All sparse variants use $90\%$ sparsity. These rows compare trained configurations, unlike the component swaps above.

Restoring dense attention or BF16 precision does not consistently improve Total without retraining. With staged training, the fully compressed model achieves 82.72, comparable to the dense BF16 control (82.69) and 1.05 points above direct NVFP4 DMD. The quality reference already uses Wan VAE; the TAEHV swap changes Total by only $+0.06$ / $-0.24$ points for 1.3B / 14B.

\FloatBarrier
\subsection{Quality and Speed at Different Sparsity Levels}
\label{app:sparsity_sweep}

We sweep inference sparsity from $70\%$ to $90\%$ on the final trained models and measure DiT speedup on RTX 5090. Only inference sparsity changes; models are not retrained for each row. Quality uses Wan VAE, five generation seeds for 1.3B, and seed 42 for 14B. Speedups are relative to dense BF16 attention with NVFP4 linears fixed. The 14B timings are extrapolated from models with fewer layers because the BF16 model exceeds RTX 5090's 32\,GB memory capacity.

\begin{table}[!htb]
\centering
\caption{Inference sparsity sweep. The 14B DiT timings are extrapolated.}
\label{tab:sparsity_sweep}
\footnotesize
\setlength{\tabcolsep}{3pt}
\renewcommand{\arraystretch}{1.12}
\begin{tabularx}{\linewidth}{@{}Xcccc@{}}
\toprule
Sparsity & 1.3B Total & 1.3B DiT speedup & 14B Total & 14B DiT speedup \\
\midrule
70\% & \textbf{82.82} & 4.13$\times$ & 84.20 & 5.44$\times$ \\
80\% & 82.79 & 4.41$\times$ & \textbf{84.52} & 6.24$\times$ \\
90\% & 82.72 & \textbf{4.87$\times$} & 84.02 & \textbf{7.26$\times$} \\
\bottomrule
\end{tabularx}
\end{table}

Across $70$--$90\%$ sparsity, Total varies by only 0.10 / 0.50 points for 1.3B / 14B while DiT speedup increases, showing no sharp quality drop in the tested range.

\FloatBarrier
\subsection{Evaluation Settings and Prompt Extension}
\label{app:prompt_extension}

For 1.3B, Wan recommends CFG 6, shift 8, and $81\times480\times832$ videos, with a Chinese negative prompt. VBench's configuration uses CFG 6 and shift 3 at the same resolution and frame count. Our base evaluation uses CFG 5, shift 3, and $61\times448\times896$ videos, without prompt extension. Since these configurations differ and the Wan report does not disclose its full evaluation protocol, we ablate the original checkpoint under different settings, using seed 42 throughout. Prompt extension uses VBench's published Wan2.1 extended prompts. Table~\ref{tab:protocol_ablation} reports the settings and results.

\begin{table}[!htb]
\centering
\caption{Evaluation-setting ablation on Wan2.1-1.3B, seed 42. Ext. denotes prompt extension.}
\label{tab:protocol_ablation}
\footnotesize
\setlength{\tabcolsep}{3pt}
\renewcommand{\arraystretch}{1.12}
\begin{tabularx}{\linewidth}{@{}ccXccccc@{}}
\toprule
CFG & Shift & Frames $\times$ H $\times$ W & Ext. & Decoder & Quality & Semantic & Total \\
\midrule
5 & 3 & 61 $\times$ 448 $\times$ 896 & Off & Wan VAE & 84.42 & 66.10 & 80.75 \\
5 & 3 & 61 $\times$ 448 $\times$ 896 & Off & TAEHV & 84.22 & 65.53 & 80.48 \\
5 & 3 & 61 $\times$ 448 $\times$ 896 & On & Wan VAE & \textbf{85.29} & 81.01 & 84.43 \\
6 & 8 & 81 $\times$ 480 $\times$ 832 & Off & Wan VAE & 84.15 & 68.34 & 80.99 \\
6 & 8 & 81 $\times$ 480 $\times$ 832 & On & Wan VAE & 85.17 & \textbf{83.88} & \textbf{84.91} \\
5 & 3 & 81 $\times$ 480 $\times$ 832 & Off & Wan VAE & 84.17 & 67.79 & 80.89 \\
5 & 5 & 81 $\times$ 480 $\times$ 832 & Off & Wan VAE & 83.77 & 68.70 & 80.75 \\
3.5 & 8 & 61 $\times$ 448 $\times$ 896 & Off & Wan VAE & 84.63 & 66.07 & 80.92 \\
\bottomrule
\end{tabularx}
\end{table}

Prompt extension has the largest observed effect in this ablation, raising Semantic by 14.91 points under our settings and 15.54 points under Wan's upstream settings. Changes to CFG, shift, and resolution have only modest effects in this ablation. These results suggest that whether prompt extension is enabled is the main contributor to the gap with Wan's reported VBench scores.

We also test prompt extension for both the base and the original JetDiffusion checkpoint, using five seeds (42--46) for 1.3B and one seed (42) for 14B. This inference-time ablation changes only the prompts; each model retains its sampling settings in Table~\ref{tab:eval_settings}. Thus, these extended-prompt rows do not constitute a switch to all of Wan's upstream sampling settings. We additionally evaluate a separate 1.3B model trained with extended prompts, as indicated in Table~\ref{tab:prompt_extension}.

\begin{table}[!htb]
\centering
\caption{Base and student with and without prompt extension at evaluation. The separately trained extended-prompt variant is marked explicitly. Results are pooled over five seeds for 1.3B; the 14B rows use one seed.}
\label{tab:prompt_extension}
\footnotesize
\setlength{\tabcolsep}{3pt}
\renewcommand{\arraystretch}{1.12}
\begin{tabularx}{\linewidth}{@{}Xccccc@{}}
\toprule
Model & Extension & Seeds & Quality & Semantic & Total \\
\midrule
Wan2.1-1.3B & Off & 42--46 & 83.52 & 64.50 & 79.72 \\
 & On & 42--46 & 85.21 & 80.95 & 84.36 \\
JetDiffusion-1.3B & Off & 42--46 & 85.28 & 72.47 & 82.72 \\
 & On & 42--46 & \textbf{86.20} & 81.25 & \textbf{85.21} \\
JetDiffusion-1.3B (trained with extended prompts) & On & 42--46 & 86.00 & \textbf{82.03} & 85.20 \\
Wan2.1-14B & Off & 42 & 84.38 & 75.41 & 82.59 \\
 & On & 42 & \textbf{85.60} & 81.53 & \textbf{84.78} \\
JetDiffusion-14B & Off & 42 & 85.16 & 79.48 & 84.02 \\
 & On & 42 & 84.84 & \textbf{83.85} & 84.64 \\
\bottomrule
\end{tabularx}
\end{table}

We also train a separate JetDiffusion-1.3B model with extended prompts and evaluate it on extended prompts, pooling results over five seeds (42--46). It achieves 86.00 Quality, 82.03 Semantic, and 85.20 Total. Compared with the original checkpoint evaluated on extended prompts, Semantic increases by 0.78 points, while Quality decreases by 0.20 points and Total remains nearly unchanged (85.20 versus 85.21).

With prompt extension enabled for both models, JetDiffusion scores higher on all three VBench metrics at 1.3B and remains comparable at 14B. Our experiments use Wan2.1 and evaluate generation quality at 61 and 77 frames; evaluation on other architectures and beyond the base models' supported duration range remains future work.

\FloatBarrier
\subsection{Decoder Evaluation}
\label{app:decoder_quality}

For efficiency comparisons with other accelerated systems, we use the lightweight TAEHV decoder; main-table quality comparisons use the original Wan VAE. To isolate the decoder's effect, we compare both decoders with each model and its sampling settings fixed. Table~\ref{tab:decoder_quality} reports VBench Quality. The 1.3B comparisons use five seeds, whereas JetDiffusion-14B uses seed 42.

\begin{table}[!htb]
\centering
\caption{VBench Quality with the original Wan VAE and TAEHV.}
\label{tab:decoder_quality}
\footnotesize
\setlength{\tabcolsep}{3pt}
\renewcommand{\arraystretch}{1.12}
\begin{tabularx}{\linewidth}{@{}Xcccc@{}}
\toprule
Model & Seeds & Wan VAE Quality & TAEHV Quality & Change \\
\midrule
Wan2.1-1.3B base & 5 & \textbf{83.53} & 83.31 & $-$0.22 \\
JetDiffusion-1.3B & 5 & 85.28 & \textbf{85.37} & +0.09 \\
JetDiffusion-14B & 1 (42) & \textbf{85.16} & 85.00 & $-$0.16 \\
\bottomrule
\end{tabularx}
\end{table}

Changing the decoder affects VBench Quality by at most 0.22 points across these comparisons. The changes in Total for the JetDiffusion models are reported in Table~\ref{tab:inference_swaps}.

\FloatBarrier
\subsection{Controlled Sparsity and Quantization Ablations}
\label{app:controlled_attention}

We benchmark four attention variants of Wan2.1-1.3B and Wan2.1-14B on identical inputs on RTX 5090 and GB300. Table~\ref{tab:controlled_attention} gives the latency of the full attention operator in ms, with speedup over dense BF16 in parentheses. Wan2.1-1.3B uses $61\times448\times896$ videos (25,088 tokens) and $4\times4\times8$ cubes; Wan2.1-14B uses $77\times768\times1280$ videos (76,800 tokens) and $4\times8\times8$ cubes. Sparse variants select identical cubes at $90\%$ sparsity.
On RTX 5090, dense BF16 uses PyTorch SDPA (FlashAttention-2); sparse BF16 uses FastVideo's Triton block-sparse kernel with PyTorch pooling and combining. Both dense and sparse low-bit variants use NVFP4 for $QK^\top$ and $PV$. On GB300, dense BF16 and low-bit use FA4 CuTe, and both low-bit variants use NVFP4 for $QK^\top$ and FP8 for $PV$. These operator comparisons include implementation and preprocessing costs, rather than isolating arithmetic precision alone.

On RTX 5090, sparsity gives $5.0\times$ / $6.2\times$, quantization $3.6\times$ / $3.7\times$, and their combination $20.6\times$ / $29.9\times$ attention speedup for 1.3B / 14B, respectively. On GB300-14B, the corresponding gains are $5.7\times$, $1.7\times$, and $8.0\times$. The RTX 5090 controlled measurements are separate from the multi-method benchmark in Table~\ref{tab:kernel_timings}; the GB300 dense BF16, dense low-bit, and VSQA attention entries are shared between the two benchmarks.

We also time one full DiT forward step with each attention variant in a separate benchmark (Table~\ref{tab:controlled_dit}). Linear precision is fixed within each hardware group: NVFP4 on RTX 5090 and BF16 on GB300. RTX 5090 14B DiT times are extrapolated from models with fewer layers because the BF16 model exceeds its 32\,GB GPU memory capacity. The RTX 5090 14B multi-method baseline in Table~\ref{tab:kernel_timings} uses CPU offloading; the controlled single-step comparison and the cumulative estimates below use the stated extrapolation protocol.

\begin{table}[!htb]
\centering
\caption{DiT forward-step latency with fixed linear precision. RTX 5090 14B times are extrapolated.}
\label{tab:controlled_dit}
\footnotesize
\setlength{\tabcolsep}{3pt}
\renewcommand{\arraystretch}{1.12}
\begin{tabularx}{\linewidth}{@{}Xccccc@{}}
\toprule
Hardware / model & Linears & Dense BF16 & Dense low-bit & Sparse BF16 & VSQA \\
\midrule
RTX 5090 / 1.3B & NVFP4 & 748.0 ms & 258.1 ms (2.9$\times$) & 400.2 ms (1.9$\times$) & \textbf{153.1 ms (4.9$\times$)} \\
RTX 5090 / 14B & NVFP4 & 25,258 ms & 7,899 ms (3.2$\times$) & 8,686 ms (2.9$\times$) & \textbf{3,470 ms (7.3$\times$)} \\
GB300 / 14B & BF16 & 6.132 s & 4.795 s (1.3$\times$) & 3.331 s (1.8$\times$) & \textbf{3.221 s (1.9$\times$)} \\
\bottomrule
\end{tabularx}
\end{table}

These attention speedups translate into $4.9\times$ / $7.3\times$ faster DiT steps on RTX 5090-1.3B / 14B with NVFP4 linears, and $1.9\times$ on GB300-14B with BF16 linears.

\FloatBarrier
\subsection{Performance at Short Sequence Lengths}
\label{app:short_sequences}

We sweep the token count on RTX 5090, keeping each model's default cube, and report full-attention latency in ms. The 1.3B setting uses 12 heads, a $14\times28$ spatial token grid per latent frame, and $4\times4\times8$ cubes; the 14B setting uses 40 heads, a $24\times40$ grid, and $4\times8\times8$ cubes. The sweep varies the number of latent frames per chunk while holding each spatial grid fixed. VSQA uses NVFP4 for both $QK^\top$ and $PV$.

\begin{table}[!htb]
\centering
\caption{Short-sequence attention latency (ms) on RTX 5090 with default cubes.}
\label{tab:short_sequences}
\footnotesize
\setlength{\tabcolsep}{3pt}
\renewcommand{\arraystretch}{1.12}
\begin{tabularx}{\linewidth}{@{}Xccccc@{}}
\toprule
Model & Tokens & Dense BF16 & Dense FP4 / FP8 & Dense FP4 / FP4 & VSQA \\
\midrule
1.3B & 392 & \textbf{0.024} & 0.025 & 0.026 & 0.049 \\
1.3B & 784 & 0.045 & \textbf{0.030} & 0.031 & 0.050 \\
1.3B & 1,568 & 0.090 & \textbf{0.042} & 0.044 & 0.050 \\
1.3B & 3,136 & 0.327 & 0.105 & 0.107 & \textbf{0.084} \\
1.3B & 6,272 & 1.266 & 0.369 & 0.363 & \textbf{0.204} \\
14B & 960 & 0.110 & \textbf{0.054} & 0.055 & 0.257 \\
14B & 1,920 & 0.403 & 0.150 & \textbf{0.144} & 0.257 \\
14B & 3,840 & 1.564 & 0.525 & 0.514 & \textbf{0.278} \\
14B & 7,680 & 5.758 & 1.815 & 1.684 & \textbf{0.563} \\
\bottomrule
\end{tabularx}
\end{table}

In Table~\ref{tab:short_sequences}, the two dense low-bit columns use NVFP4 $QK^\top$ with FP8 or NVFP4 $PV$, respectively. With default cubes, VSQA is fastest from 3,136 tokens for 1.3B and 3,840 for 14B among the tested sizes. Dense attention is preferable for shorter chunks; NVFP4 and FP8 $PV$ differ by only a few microseconds there. These crossover points apply to the tested shapes and default cubes, not to arbitrary sequence layouts or retuned cube sizes.

\FloatBarrier
\subsection{Kernel Profiling and Pooling--Quantization Fusion}
\label{app:kernel_profiling}

We profile four VSQA stages at $90\%$ sparsity on RTX 5090 and GB300 using the video sizes and precision settings in Appendix~\ref{app:controlled_attention}. Table~\ref{tab:kernel_profile} reports stage latency and each stage's fraction of the summed stage times. Stage profiling is measured separately from the full-operator benchmark, so its sum need not equal the operator latency in Table~\ref{tab:controlled_attention}.

\begin{table}[!htb]
\centering
\caption{Stage-level VSQA latency (ms) and share of the stage total.}
\label{tab:kernel_profile}
\footnotesize
\setlength{\tabcolsep}{3pt}
\renewcommand{\arraystretch}{1.12}
\begin{tabularx}{\linewidth}{@{}Xccc@{}}
\toprule
Stage & RTX 5090, 1.3B & RTX 5090, 14B & GB300, 14B \\
\midrule
Quantization/pooling and packing/tile & 0.195 (22.5\%) & 1.93 (10.2\%) & 4.518 (45.1\%) \\
Routing & 0.033 (3.8\%) & 0.18 (1.0\%) & 0.257 (2.6\%) \\
Fine attention & 0.485 (55.9\%) & 15.66 (82.8\%) & 4.538 (45.3\%) \\
Combining branches / untile & 0.155 (17.9\%) & 1.15 (6.1\%) & 0.710 (7.1\%) \\
\bottomrule
\end{tabularx}
\end{table}

Fine attention accounts for $55.9\%$ / $82.8\%$ / $45.3\%$ of the stage totals; routing adds only $1.0$--$3.8\%$.

To isolate pooling--quantization fusion, we conduct a four-step ablation from FastVideo VSA to VSQA on GB300 at $77\times768\times1280$, with identical inputs and selected cubes at $90\%$ sparsity (Table~\ref{tab:fusion_ablation}).

\begin{table}[!htb]
\centering
\caption{GB300 full-operator ablation from FastVideo VSA to VSQA.}
\label{tab:fusion_ablation}
\footnotesize
\setlength{\tabcolsep}{3pt}
\renewcommand{\arraystretch}{1.12}
\begin{tabularx}{\linewidth}{@{}Xccc@{}}
\toprule
Configuration & Latency (ms) & Vs. previous & Vs. dense BF16 \\
\midrule
FastVideo VSA, FA4 CuTe BF16 fine kernel & 17.06 & --- & 4.7$\times$ \\
+ Optimized pooling, mask packing, and branch combining; BF16 & 13.88 & 1.23$\times$ & 5.7$\times$ \\
+ NVFP4 QK / FP8 PV, separate pooling and quantization & 10.46 & 1.33$\times$ & 7.6$\times$ \\
+ Fused pooling and quantization (VSQA) & \textbf{9.98} & 1.05$\times$ & \textbf{8.0$\times$} \\
\bottomrule
\end{tabularx}
\end{table}

The final two rows use the same fine-attention kernel and differ only in pooling--quantization fusion. Fusion reduces preparation time from 1.20 to 0.73 ms ($1.64\times$), yielding a $1.05\times$ full-operator speedup. Together, the changes make VSQA $1.71\times$ faster than FastVideo VSA. The 13.88 ms row is the sparse BF16 variant in the four-variant ablation.

\FloatBarrier
\subsection{End-to-End Component Ablation and Decoder Control}
\label{app:e2e_ablation}

We ablate the RTX 5090 14B pipeline cumulatively, including the decoder as the final change. The video size is $77\times768\times1280$. The base uses 50 denoising steps with CFG; removing CFG leaves 50 forwards, followed by reducing this to three forwards. Wan VAE stays fixed until the last row. BF16 14B exceeds RTX 5090's 32\,GB memory, so estimates combine DiT times extrapolated from fewer layers with measured non-DiT times. The final three-step model is also timed directly with each decoder.

\begin{table}[!htb]
\centering
\caption{Cumulative RTX 5090 14B pipeline ablation. Intermediate BF16 configurations are estimated; the final pipeline is measured.}
\label{tab:e2e_ablation}
\footnotesize
\setlength{\tabcolsep}{3pt}
\renewcommand{\arraystretch}{1.12}
\begin{tabularx}{\linewidth}{@{}Xc@{}}
\toprule
Cumulative configuration & End-to-end latency \\
\midrule
Base: 50 steps with CFG, dense BF16, optimized Wan VAE & 3,177 s \\
Remove CFG: 50 forwards & 1,591 s \\
Reduce to three forwards & 100.0 s \\
+ VSQA & 34.68 s \\
+ NVFP4 linears & 15.29 s estimated / 15.64 s measured \\
+ TAEHV decoder & \textbf{11.64 s measured} \\
\bottomrule
\end{tabularx}
\end{table}

On top of three-step distillation, VSQA provides a further $2.88\times$ acceleration with the decoder fixed. NVFP4 linears add another $2.27\times$ using the estimated 15.29 s configuration (or $2.22\times$ using its 15.64 s measured latency), and switching to TAEHV adds $1.34\times$ using the measured final pipelines. With the same Wan decoder, the combined pipeline achieves $203\times$ acceleration. Comparing the estimated 3,177 s BF16 base with the measured 11.64 s TAEHV pipeline gives $273\times$ end-to-end acceleration.

\FloatBarrier
\subsection{GB300 Generation Timings}
\label{app:gb300_aligned}

We measure GB300 generation latency at 77 frames using the same optimized Wan VAE implementation as RTX 5090 for the base, and TAEHV for JetDiffusion. Both use $768\times1280$ resolution. The base uses dense FA4, 50 steps, and CFG; JetDiffusion uses VSQA, NVFP4 linears, and three steps. This full-pipeline measurement therefore differs from the BF16-linear controlled DiT benchmark in Table~\ref{tab:controlled_dit}.

\begin{table}[!htb]
\centering
\caption{GB300 generation timings for 77-frame videos, using the same baseline decoder implementation as RTX 5090.}
\label{tab:gb300_aligned}
\footnotesize
\setlength{\tabcolsep}{3pt}
\renewcommand{\arraystretch}{1.12}
\begin{tabularx}{\linewidth}{@{}Xccc@{}}
\toprule
Timing & Base, optimized Wan VAE & JetDiffusion, TAEHV & Speedup \\
\midrule
Average DiT forward & 4.88 s & \textbf{1.28 s} & 3.8$\times$ \\
Total DiT & 487.60 s & \textbf{3.83 s} & 127$\times$ \\
Non-DiT & 1.376 s & \textbf{0.280 s} & 4.9$\times$ \\
End-to-end & 488.97 s & \textbf{4.11 s} & 119$\times$ \\
\bottomrule
\end{tabularx}
\end{table}

With the decoder implementation and frame count aligned, GB300 achieves a $4.9\times$ non-DiT speedup, comparable to the approximately $5\times$ on RTX 5090. End-to-end latency decreases from 488.97 s to 4.11 s ($119\times$).

\FloatBarrier
\subsection{Matched-Setting Quality and Latency Comparison}
\label{app:matched_quality_latency}
We compare JetDiffusion with TurboDiffusion and LightWan2.2 on the 14B models at one setting. All three generate 77 frames at $768\times1280$ with three denoising steps, and TAEHV decodes the latents for both the latency and the quality measurements. Each baseline runs in its released pipeline with its released checkpoints, at $90\%$ attention sparsity. The three-step schedules are chosen as follows.

\begin{itemize}
    \item \textbf{TurboDiffusion.} We use the three-step setting recommended by its authors.
    \item \textbf{LightX2V.} Since no recommended three-step setting is released, we start from its four-step schedule (high, high, low, low) and conduct a preliminary comparison of (high, high, low) and (high, low, low). The former produces noticeably better samples in this smoke test, so we use (high, high, low) for evaluation.
\end{itemize}

Table~\ref{tab:matched_latency} reports latency on one RTX 5090 as the median of three generations after one warm-up generation. At equal steps, TurboDiffusion takes $2.48\times$ the DiT time of JetDiffusion and LightWan2.2 takes $1.09\times$. End to end with TAEHV the ratios are $2.36\times$ and $1.06\times$. With the Wan VAE path of each system they are $2.16\times$ and $1.20\times$.

\begin{table}[!htb]
\centering
\caption{Latency (s) on RTX 5090 at the matched 14B setting: 77 frames at $768\times1280$, three steps. Numbers in parentheses are the time relative to JetDiffusion.}
\label{tab:matched_latency}
\footnotesize
\setlength{\tabcolsep}{4pt}
\renewcommand{\arraystretch}{1.12}
\begin{tabular}{@{}lcccc@{}}
\toprule
Method & DiT per step & DiT, 3 steps & End to end, TAEHV & End to end, own Wan VAE \\
\midrule
TurboDiffusion-14B & 8.83 & 26.47 (2.48$\times$) & 27.33 (2.36$\times$) & 33.65 (2.16$\times$) \\
LightWan2.2-A14B & 3.87 & 11.60 (1.09$\times$) & 12.28 (1.06$\times$) & 18.64 (1.20$\times$) \\
\rowcolor{tableours}
\textbf{Ours (JetDiffusion)} & \textbf{3.56} & \textbf{10.67} & \textbf{11.60} & \textbf{15.55} \\
\bottomrule
\end{tabular}
\end{table}

Table~\ref{tab:matched_quality} reports VBench at the same setting for generation seed 42, with the 946 prompts and 16 dimensions of Table~\ref{tab:main_results}. As for Table~\ref{tab:main_results}, the videos are generated on GB200. JetDiffusion reaches 83.78 Total, against 83.63 for LightWan2.2 and 80.16 for TurboDiffusion.

\begin{table}[!htb]
\centering
\caption{VBench at the matched 14B setting with TAEHV decoding, seed 42.}
\label{tab:matched_quality}
\footnotesize
\setlength{\tabcolsep}{6pt}
\renewcommand{\arraystretch}{1.12}
\begin{tabular}{@{}lccc@{}}
\toprule
Method & Quality $\uparrow$ & Semantic $\uparrow$ & Total $\uparrow$ \\
\midrule
TurboDiffusion-14B & 82.62 & 70.29 & 80.16 \\
LightWan2.2-A14B & 84.32 & \textbf{80.87} & 83.63 \\
\rowcolor{tableours}
\textbf{Ours (JetDiffusion)} & \textbf{85.00} & 78.88 & \textbf{83.78} \\
\bottomrule
\end{tabular}
\end{table}
\FloatBarrier

\FloatBarrier
\subsection{Training Curves of Staged and Direct DMD}
\label{app:dmd_curves}
\begin{figure}[!htb]
    \centering
    \includegraphics[width=\linewidth]{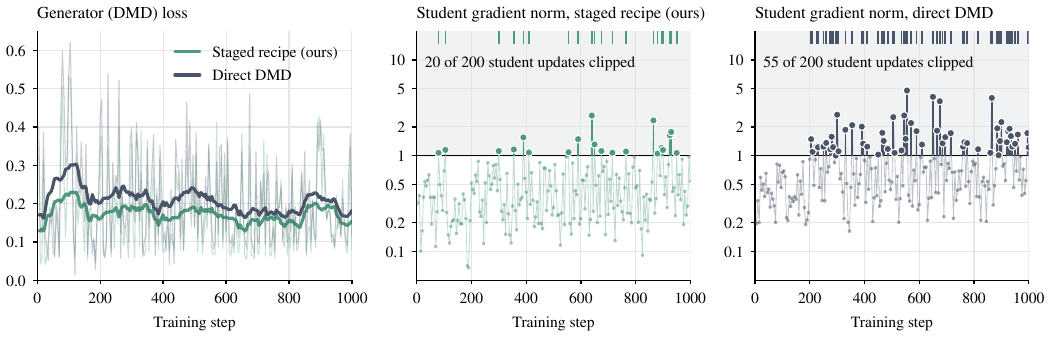}
    \caption{\textbf{DMD training curves on Wan2.1-1.3B.} Left: generator loss at every student update (thin) and its mean over 20 updates (thick). Middle and right: student gradient norm before clipping. The line marks the clip threshold of 1. Updates above it are drawn as large markers and as ticks on the top edge.}
    \label{fig:dmd_curves}
\end{figure}

Figure~\ref{fig:dmd_curves} compares the DMD stage of our staged recipe with direct DMD on Wan2.1-1.3B. The two runs use one training recipe and differ only in the student initialization. Both take 1,000 steps with one student update every five steps, and gradients are clipped at norm 1. Neither run produces a non-finite loss or gradient norm, and both generator losses decrease over training. The generator loss of the staged recipe is lower throughout, with a mean of 0.173 against 0.210. Direct DMD exceeds the clip threshold on 55 of 200 student updates and the staged recipe on 20. The largest gradient norms are 4.80 and 2.63.

\FloatBarrier

\FloatBarrier
\subsection{Human Preference Evaluation}
\label{app:human_preference}
We conduct a pairwise human preference study comparing JetDiffusion-14B with Wan2.1-14B, TurboDiffusion-14B, and LightWan2.2-A14B. Each method generates videos using its recommended inference configuration. For each prompt, we align the two videos for presentation by cropping to the smaller frame count, height, and width within the pair. We trim excess frames from the beginning and end of the longer video and apply a spatial center crop to the larger video. Participants select the video they prefer overall or indicate a tie. We report the preference share for JetDiffusion, counting ties as half a vote.

\begin{figure}[!htb]
    \centering
    \includegraphics[width=\linewidth]{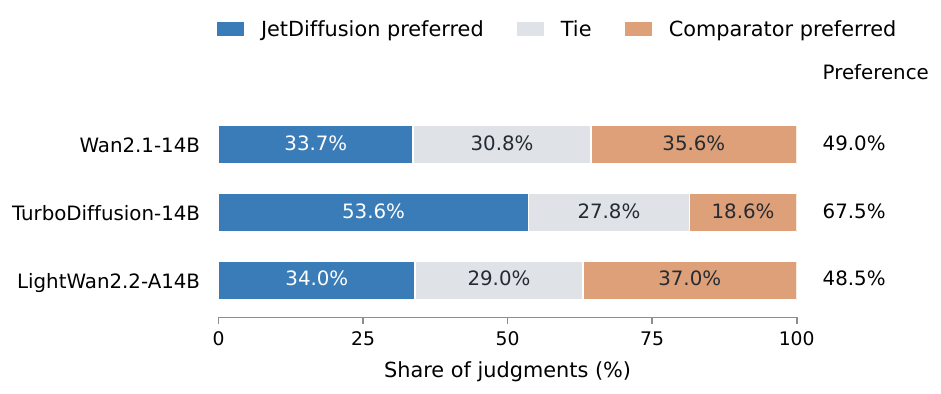}
    \caption{\textbf{Human preference evaluation on 14B models.} Bars show the proportions of judgments favoring JetDiffusion, ties, and judgments favoring the comparator. Ties count as half a vote.}
    \label{fig:human_preference}
\end{figure}

Figure~\ref{fig:human_preference} shows that JetDiffusion receives 67.5\% preference against TurboDiffusion, while preferences against Wan2.1 and LightWan2.2 are approximately balanced at 49.0\% and 48.5\%, respectively.

\FloatBarrier

\section{Quantization in Training and Inference}
\label{app:quant_comparisons}
\label{app:fake_real_quant}
\begin{figure}[!htb]
    \centering
    \newlength{\frqgap}\setlength{\frqgap}{1pt}%
    \newlength{\frqframe}\setlength{\frqframe}{\dimexpr(0.9\linewidth-2\frqgap)/3\relax}%
    \makebox[\linewidth][c]{%
        \parbox[b]{\frqframe}{\centering\small\textbf{Training forward}\\(fake quantization)}\hspace{\frqgap}%
        \parbox[b]{\frqframe}{\centering\small\textbf{Inference, RTX 5090}\\(NVFP4 $QK^\top$ and $PV$)}\hspace{\frqgap}%
        \parbox[b]{\frqframe}{\centering\small\textbf{Inference, GB300}\\(NVFP4 $QK^\top$, FP8 $PV$)}}\\[2pt]
    \makebox[\linewidth][c]{\includegraphics[width=\frqframe]{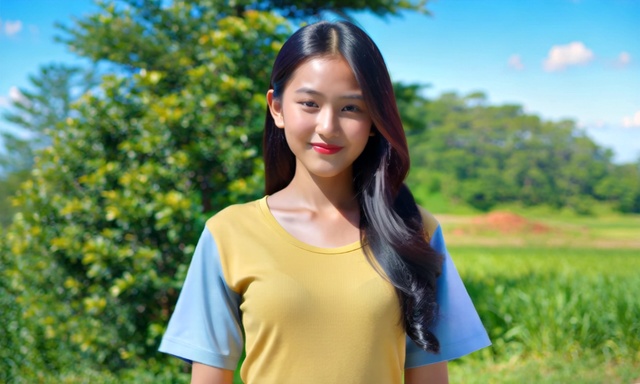}\hspace{\frqgap}\includegraphics[width=\frqframe]{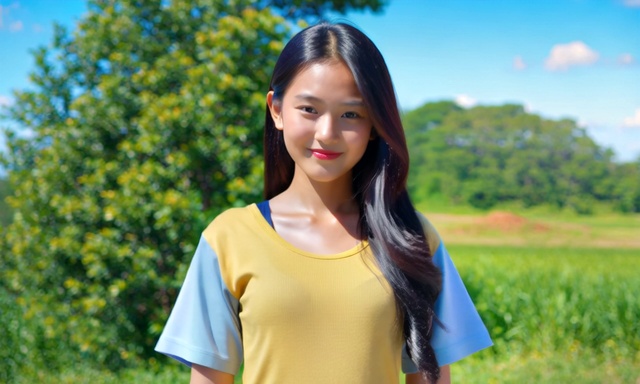}\hspace{\frqgap}\includegraphics[width=\frqframe]{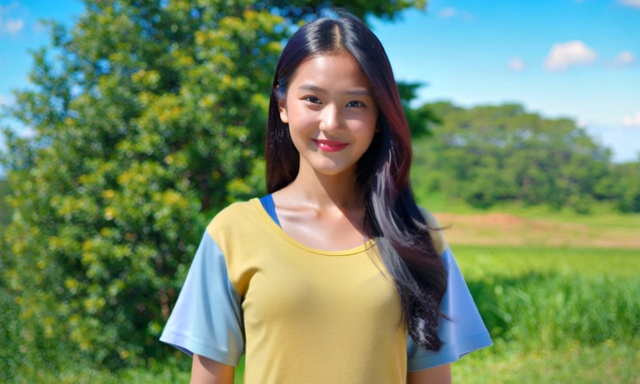}}\\[-0.6pt]
    \makebox[\linewidth][c]{\includegraphics[width=\frqframe]{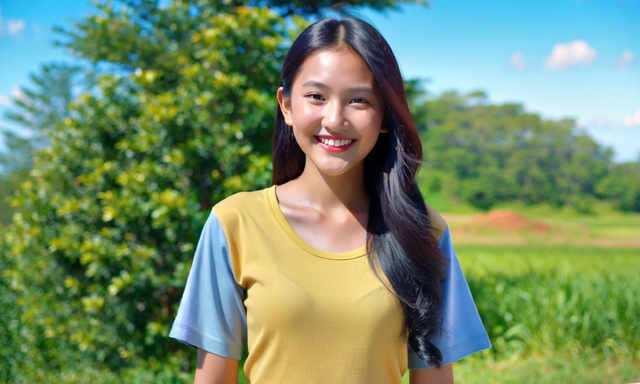}\hspace{\frqgap}\includegraphics[width=\frqframe]{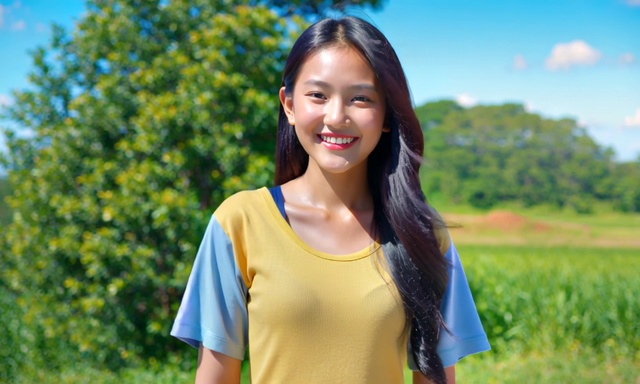}\hspace{\frqgap}\includegraphics[width=\frqframe]{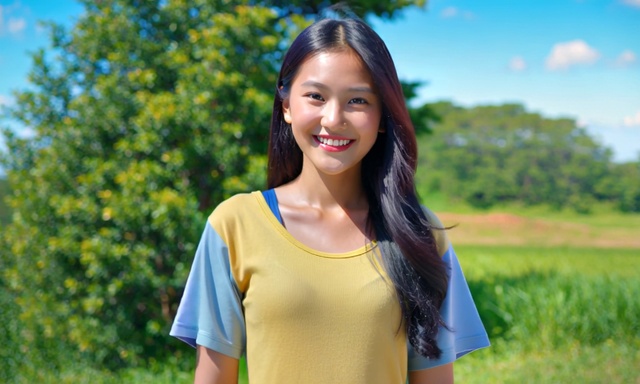}}\\[-0.6pt]
    \makebox[\linewidth][c]{\includegraphics[width=\frqframe]{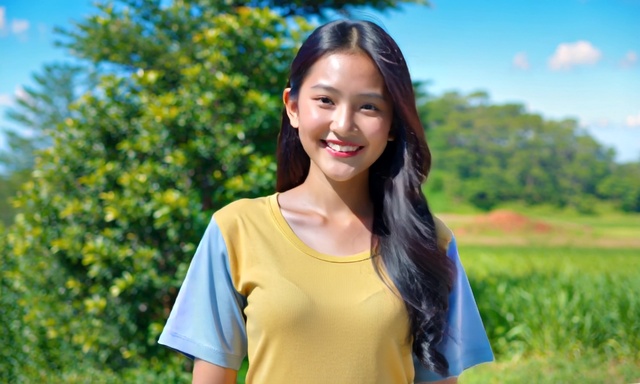}\hspace{\frqgap}\includegraphics[width=\frqframe]{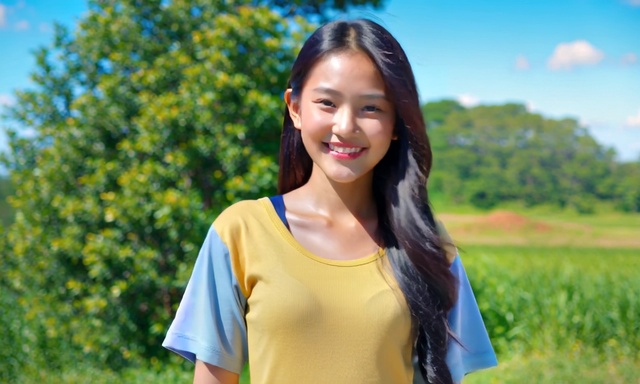}\hspace{\frqgap}\includegraphics[width=\frqframe]{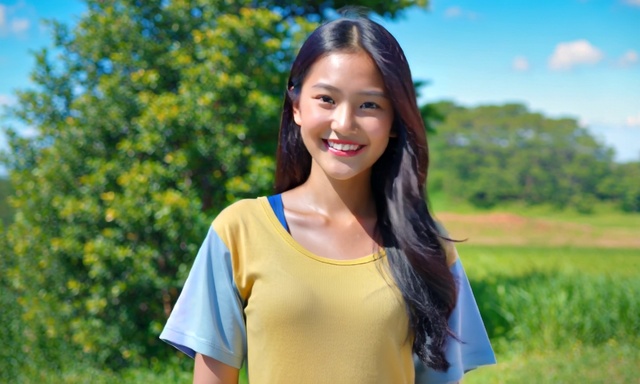}}\\[-0.6pt]
    \makebox[\linewidth][c]{\includegraphics[width=\frqframe]{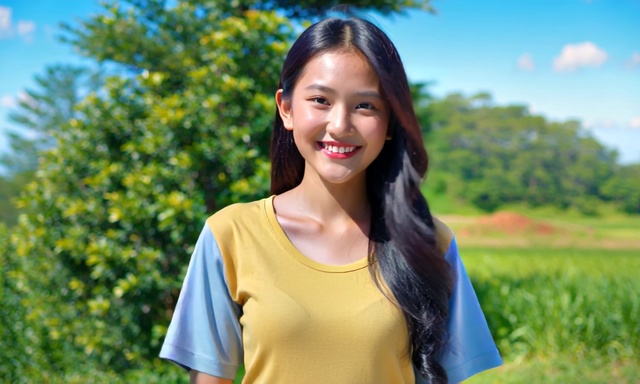}\hspace{\frqgap}\includegraphics[width=\frqframe]{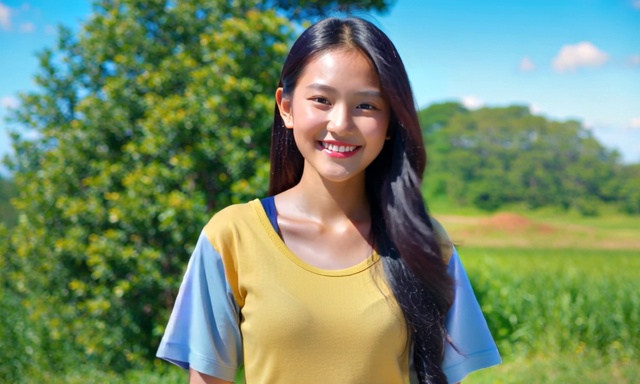}\hspace{\frqgap}\includegraphics[width=\frqframe]{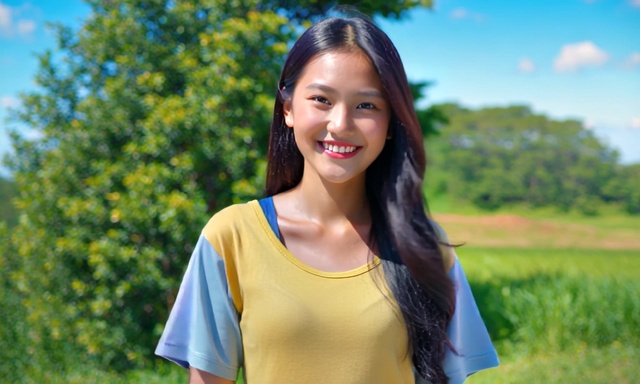}}\\[-0.6pt]
    \makebox[\linewidth][c]{\includegraphics[width=\frqframe]{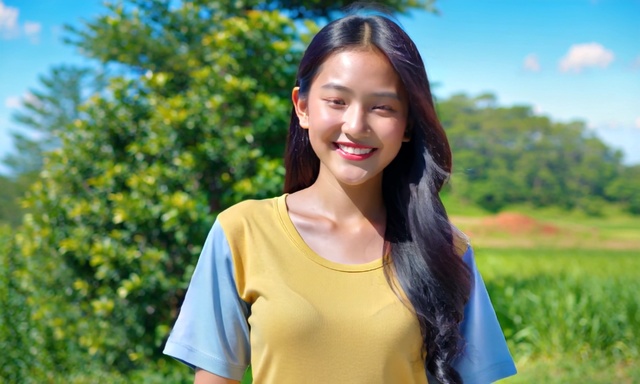}\hspace{\frqgap}\includegraphics[width=\frqframe]{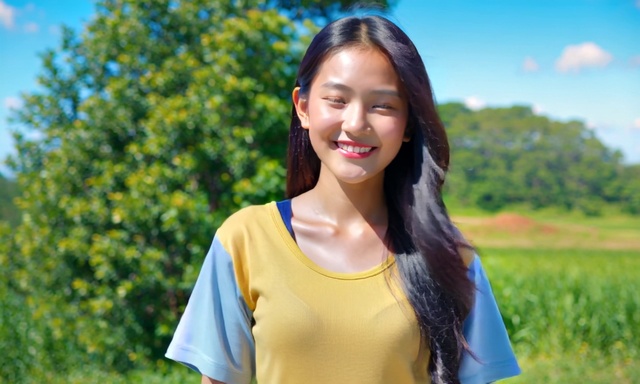}\hspace{\frqgap}\includegraphics[width=\frqframe]{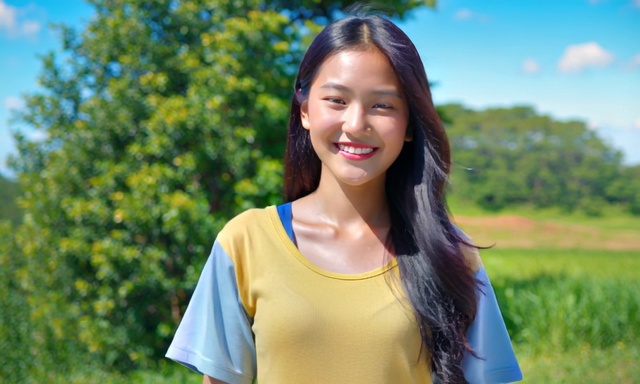}}\\[-0.6pt]
    \makebox[\linewidth][c]{\includegraphics[width=\frqframe]{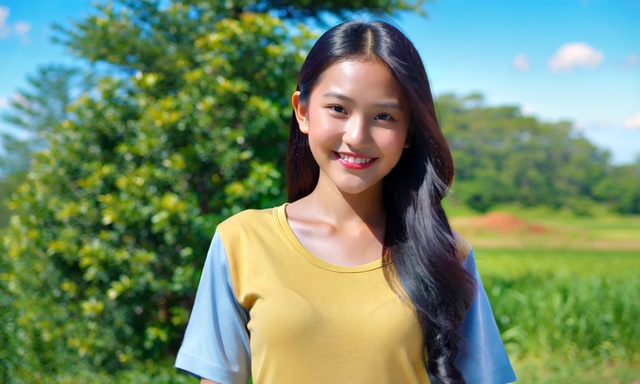}\hspace{\frqgap}\includegraphics[width=\frqframe]{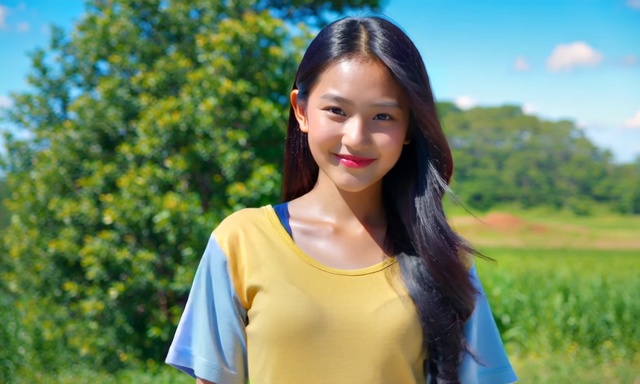}\hspace{\frqgap}\includegraphics[width=\frqframe]{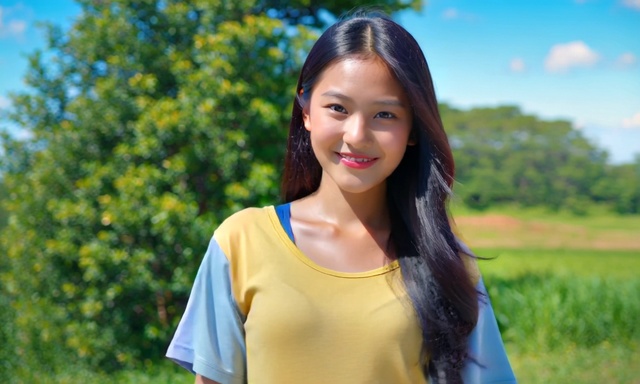}}\\[-0.6pt]
    \caption{\textbf{Fake and real quantization, and NVFP4 versus FP8 $PV$.}
    Six evenly spaced frames of one Wan2.1-14B 720p video, from the first frame (top) to the last (bottom).
    Text prompt: \textit{``A Chinese girl with a radiant smile, exuding beauty and confidence, stands in front of the camera under natural sunlight. She is dressed in a casual outfit with a harmonious blend of yellow and blue tones, complementing the vibrant hues of her surroundings. Her posture is relaxed and natural, capturing a positive and joyful perspective. The background showcases a picturesque natural landscape, featuring lush greenery and serene blue skies. The scene is captured in a medium close-up half-body shot, highlighting her engaging facial expressions and the cheerful ambiance.''}}
    \label{fig:fake_real_quant}
\end{figure}

Figure~\ref{fig:fake_real_quant} compares one Wan2.1-14B video at 720p generated three ways from the same prompt: the training forward with fake quantization, inference with NVFP4 $QK^\top$ and $PV$ on the RTX 5090, and inference with NVFP4 $QK^\top$ and FP8 $PV$ on the GB300, where the kernel computes $PV$ in FP8 for higher throughput.
The first column against the other two compares fake and real quantization, and the last two columns compare NVFP4 and FP8 $PV$.
The three videos show no visible difference.
Training with fake quantization therefore carries over to both real inference kernels.

\FloatBarrier

\section{Qualitative Results}
\label{app:qualitative}
Figures in this section compare videos from the base model and from JetDiffusion for the same text prompt.
Each row shows six frames evenly spaced from the first frame to the last.

\newlength{\vsgap}\setlength{\vsgap}{1pt}
\newlength{\vsframe}\setlength{\vsframe}{\dimexpr(\textwidth-5\vsgap)/6\relax}

\subsection{Wan2.1-T2V-1.3B}

\begin{figure}[!htb]
    \centering\lineskip=1pt
    {\small\textbf{Wan2.1-T2V-1.3B}}\\[1pt]
    \makebox[\linewidth][c]{\includegraphics[width=\vsframe]{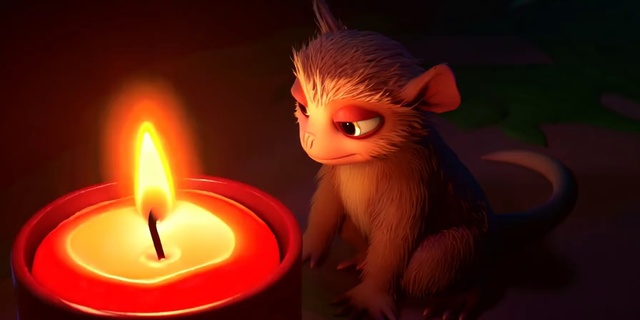}\hspace{\vsgap}\includegraphics[width=\vsframe]{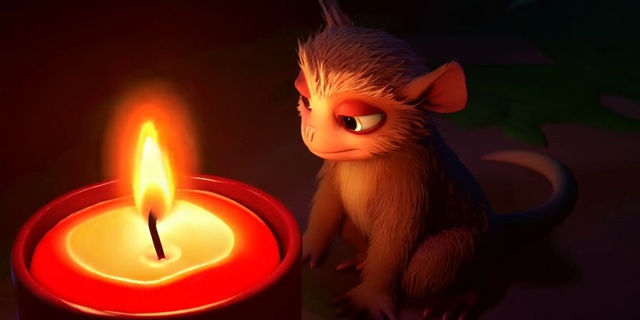}\hspace{\vsgap}\includegraphics[width=\vsframe]{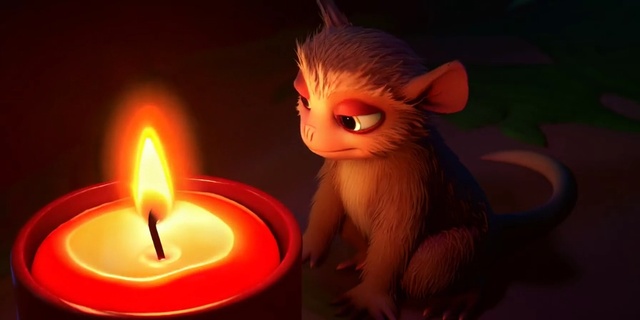}\hspace{\vsgap}\includegraphics[width=\vsframe]{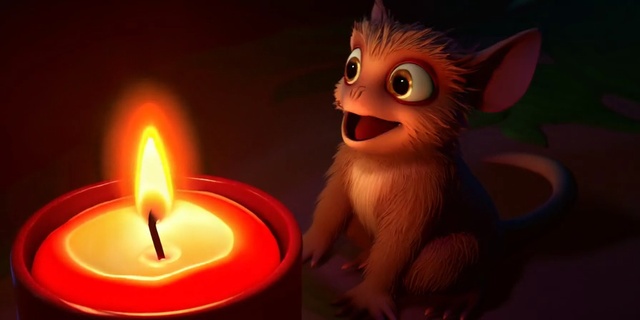}\hspace{\vsgap}\includegraphics[width=\vsframe]{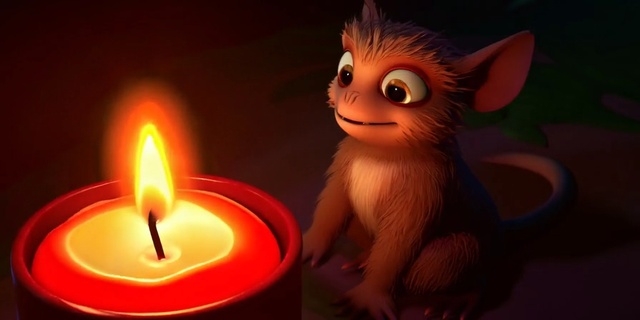}\hspace{\vsgap}\includegraphics[width=\vsframe]{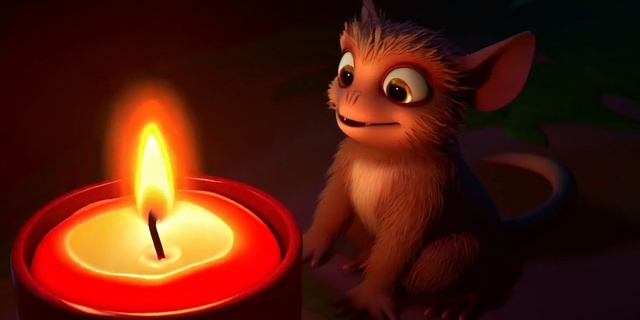}}\\
    [3pt]{\small\textbf{JetDiffusion (Ours)}}\\[1pt]
    \makebox[\linewidth][c]{\includegraphics[width=\vsframe]{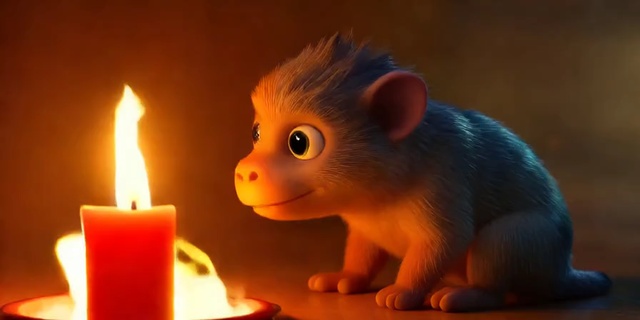}\hspace{\vsgap}\includegraphics[width=\vsframe]{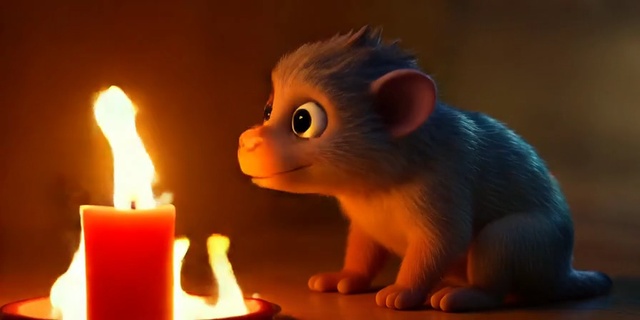}\hspace{\vsgap}\includegraphics[width=\vsframe]{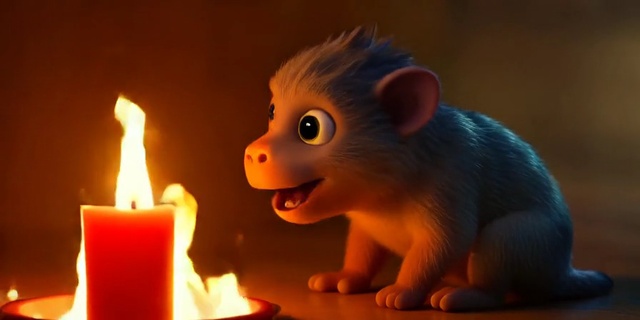}\hspace{\vsgap}\includegraphics[width=\vsframe]{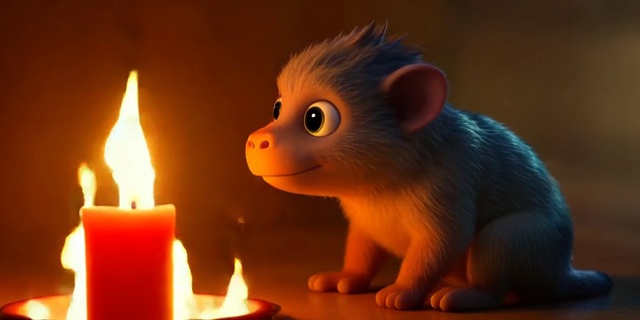}\hspace{\vsgap}\includegraphics[width=\vsframe]{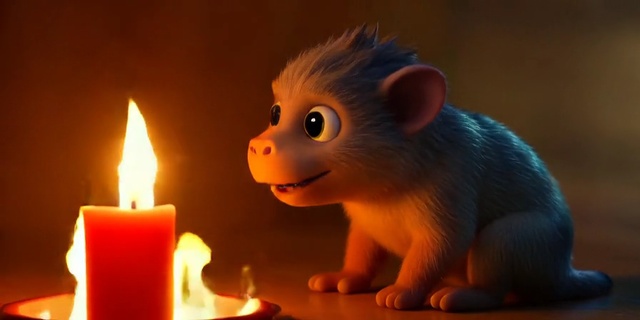}\hspace{\vsgap}\includegraphics[width=\vsframe]{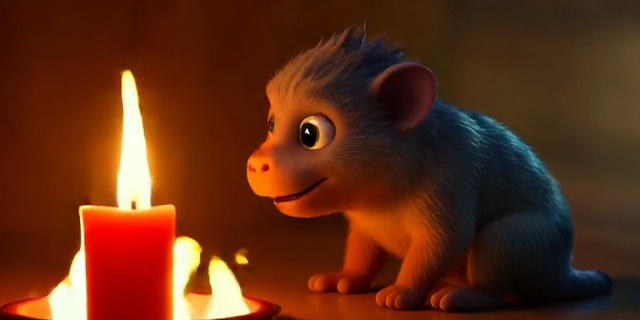}}\\
    \caption{Video generation with Wan2.1-T2V-1.3B. Top: the base model. Bottom: JetDiffusion.
    Text prompt: \textit{``Animated scene features a close-up of a short fluffy monster kneeling beside a melting red candle. The art style is 3D and realistic, with a focus on lighting and texture. The mood of the painting is one of wonder and curiosity, as the monster gazes at the flame with wide eyes and open mouth. Its pose and expression convey a sense of innocence and playfulness, as if it is exploring the world around it for the first time. The use of warm colors and dramatic lighting further enhances the cozy atmosphere of the image.''}}
    \label{fig:video_wan13b_s01}
\end{figure}

\begin{figure}[!htb]
    \centering\lineskip=1pt
    {\small\textbf{Wan2.1-T2V-1.3B}}\\[1pt]
    \makebox[\linewidth][c]{\includegraphics[width=\vsframe]{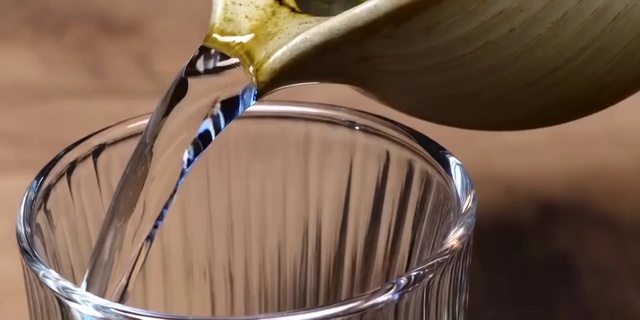}\hspace{\vsgap}\includegraphics[width=\vsframe]{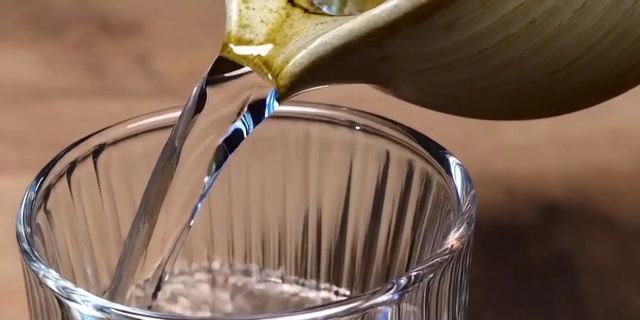}\hspace{\vsgap}\includegraphics[width=\vsframe]{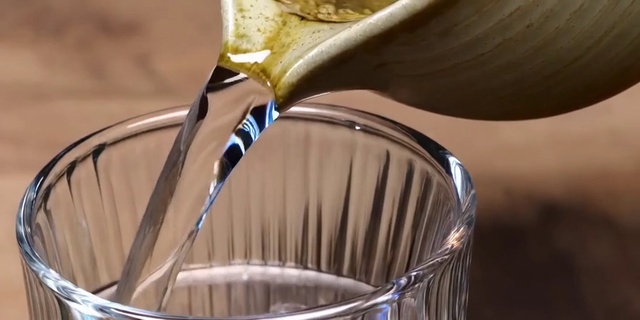}\hspace{\vsgap}\includegraphics[width=\vsframe]{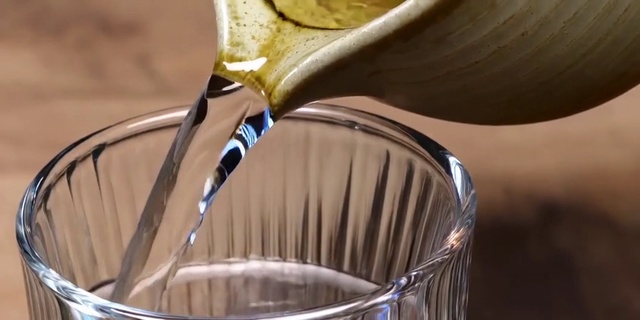}\hspace{\vsgap}\includegraphics[width=\vsframe]{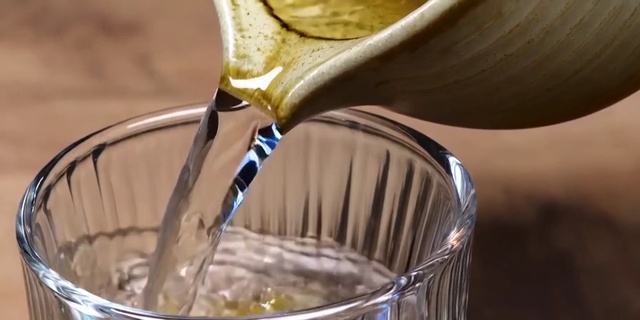}\hspace{\vsgap}\includegraphics[width=\vsframe]{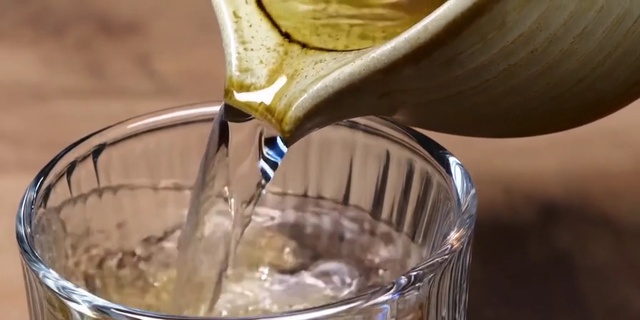}}\\
    [3pt]{\small\textbf{JetDiffusion (Ours)}}\\[1pt]
    \makebox[\linewidth][c]{\includegraphics[width=\vsframe]{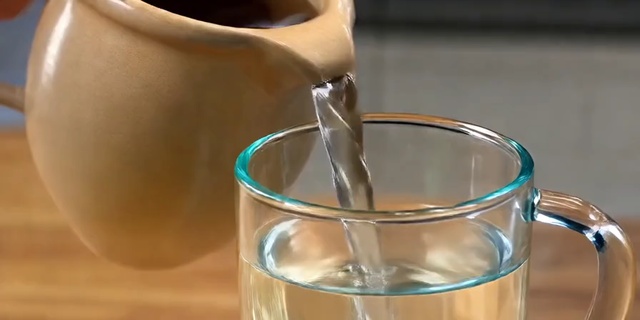}\hspace{\vsgap}\includegraphics[width=\vsframe]{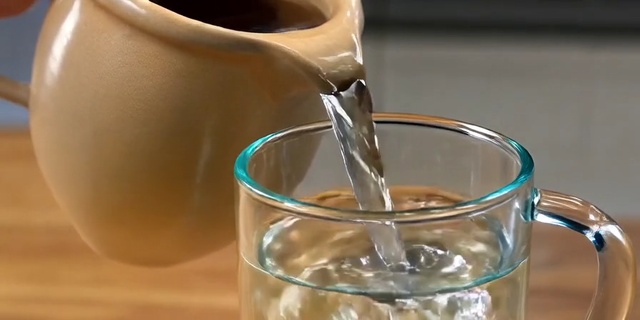}\hspace{\vsgap}\includegraphics[width=\vsframe]{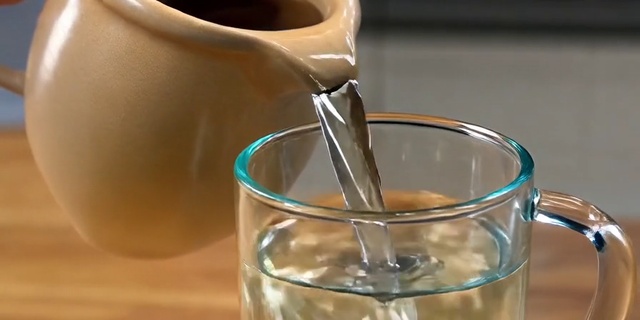}\hspace{\vsgap}\includegraphics[width=\vsframe]{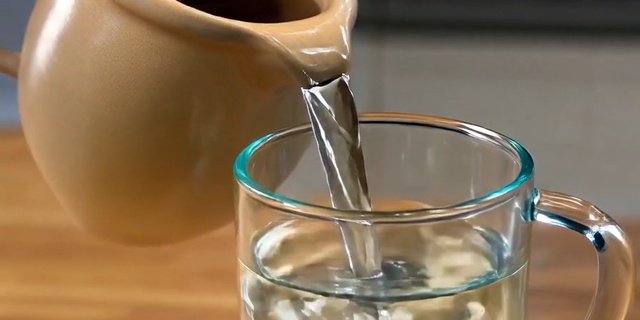}\hspace{\vsgap}\includegraphics[width=\vsframe]{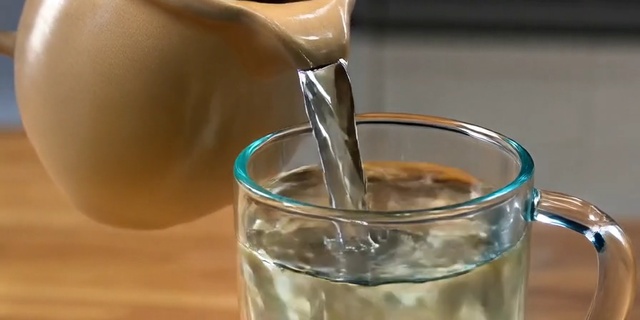}\hspace{\vsgap}\includegraphics[width=\vsframe]{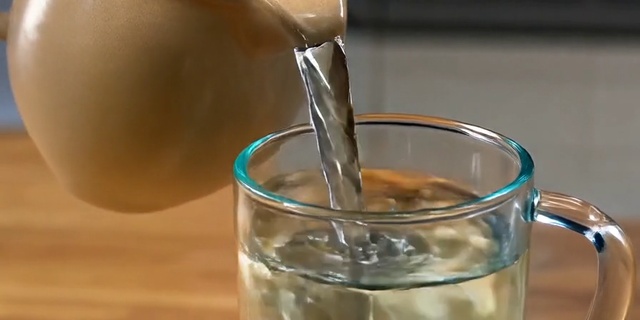}}\\
    \caption{Video generation with Wan2.1-T2V-1.3B. Top: the base model. Bottom: JetDiffusion.
    Text prompt: \textit{``A close-up shot of a ceramic teacup slowly pouring water into a glass mug. The water flows smoothly from the spout of the teacup into the mug, creating gentle ripples as it fills up. Both cups have detailed textures, with the teacup having a matte finish and the glass mug showcasing clear transparency. The background is a blurred kitchen countertop, adding context without distracting from the central action. The pouring motion is fluid and natural, emphasizing the interaction between the two cups.''}}
    \label{fig:video_wan13b_s02}
\end{figure}

\begin{figure}[!htb]
    \centering\lineskip=1pt
    {\small\textbf{Wan2.1-T2V-1.3B}}\\[1pt]
    \makebox[\linewidth][c]{\includegraphics[width=\vsframe]{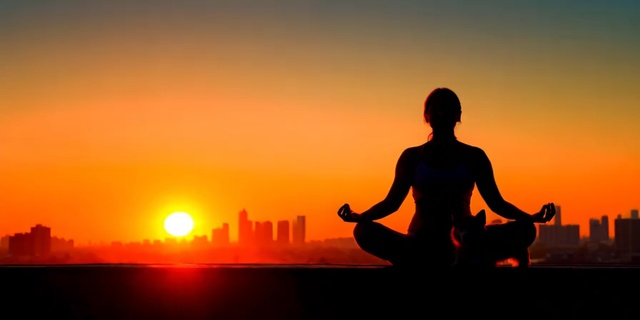}\hspace{\vsgap}\includegraphics[width=\vsframe]{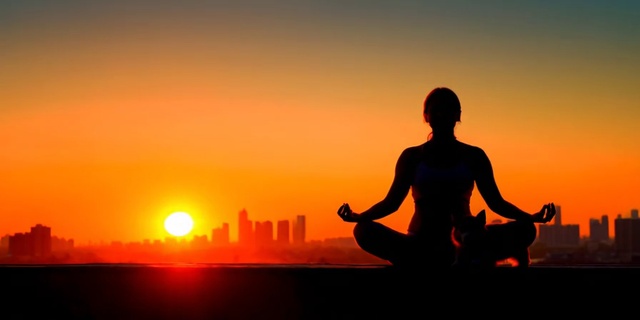}\hspace{\vsgap}\includegraphics[width=\vsframe]{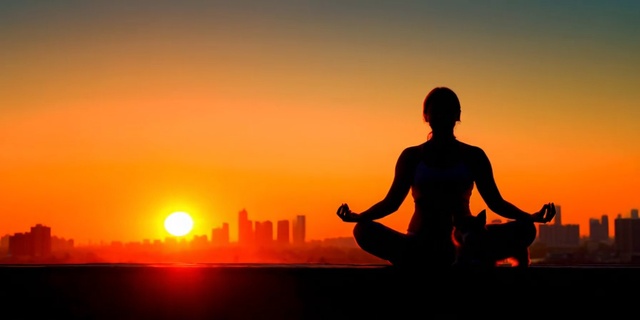}\hspace{\vsgap}\includegraphics[width=\vsframe]{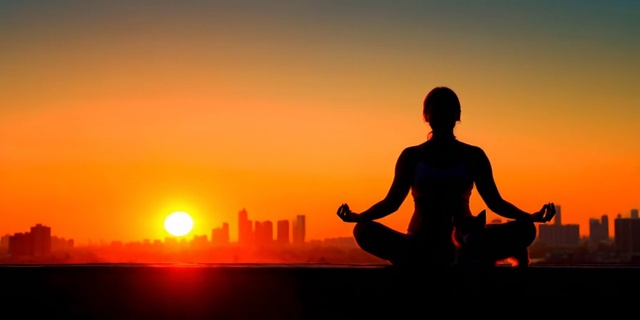}\hspace{\vsgap}\includegraphics[width=\vsframe]{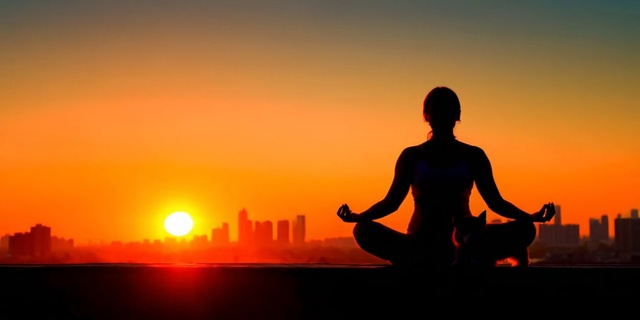}\hspace{\vsgap}\includegraphics[width=\vsframe]{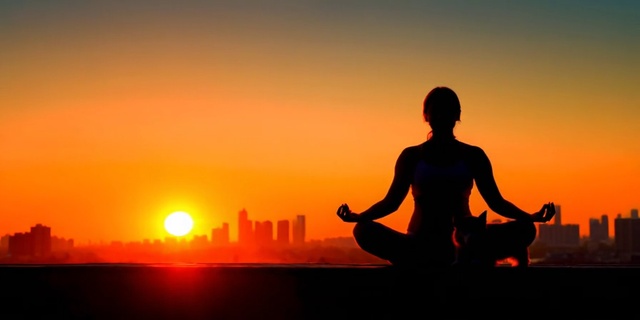}}\\
    [3pt]{\small\textbf{JetDiffusion (Ours)}}\\[1pt]
    \makebox[\linewidth][c]{\includegraphics[width=\vsframe]{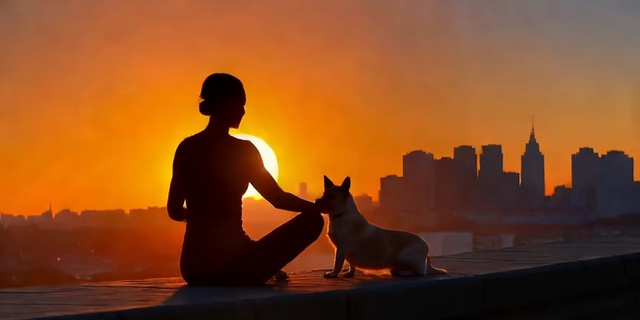}\hspace{\vsgap}\includegraphics[width=\vsframe]{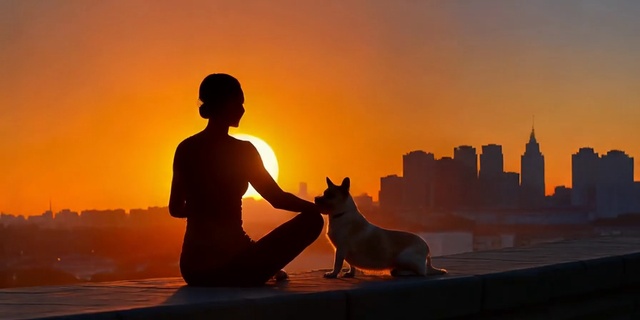}\hspace{\vsgap}\includegraphics[width=\vsframe]{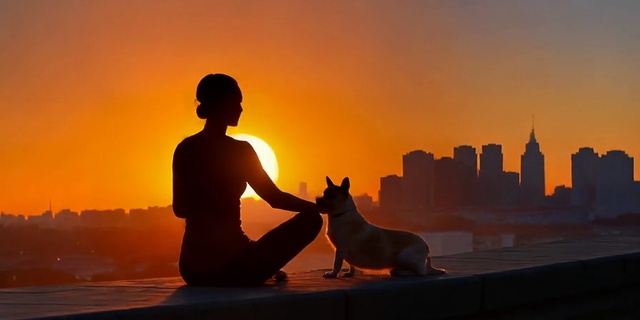}\hspace{\vsgap}\includegraphics[width=\vsframe]{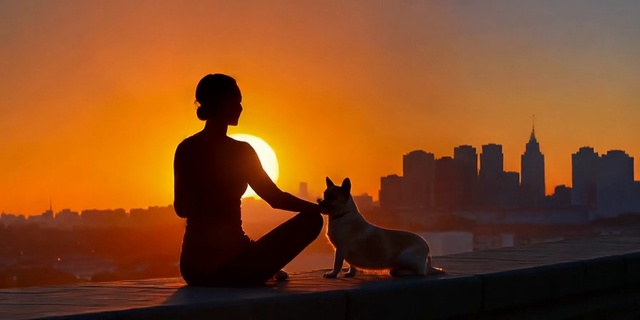}\hspace{\vsgap}\includegraphics[width=\vsframe]{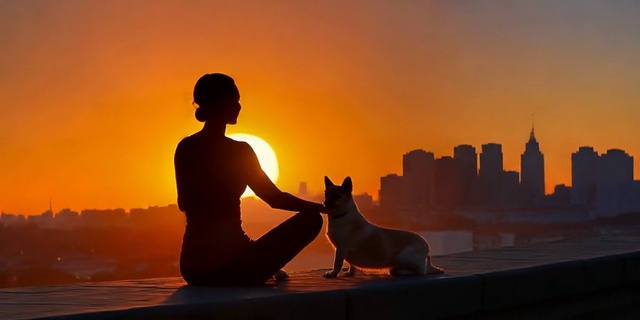}\hspace{\vsgap}\includegraphics[width=\vsframe]{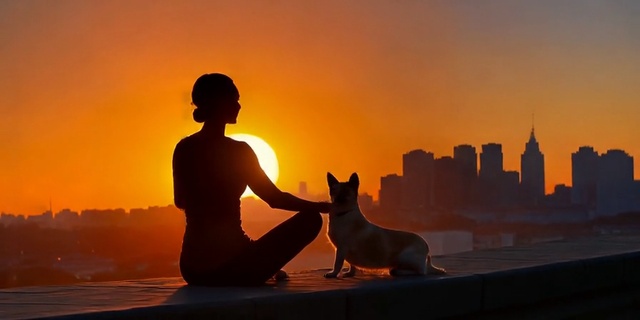}}\\
    \caption{Video generation with Wan2.1-T2V-1.3B. Top: the base model. Bottom: JetDiffusion.
    Text prompt: \textit{``In this serene video, a person sits cross-legged on a ledge in meditation, silhouetted against a vibrant orange sunset. They are captured in a moment of tranquility, with their eyes closed and hands resting on their knees in a mudra, as a small dog lies peacefully beside them. The background reveals a sprawling urban skyline, with buildings outlined against the glowing sun, which hangs low in the sky. The contrasting shadows beautifully highlight the figures, creating an atmosphere of peaceful coexistence between nature and city life. This captivating scene evokes a feeling of calm and reflection, perfectly encapsulated by the warm and harmonious colors of dusk.''}}
    \label{fig:video_wan13b_s03}
\end{figure}

\begin{figure}[!htb]
    \centering\lineskip=1pt
    {\small\textbf{Wan2.1-T2V-1.3B}}\\[1pt]
    \makebox[\linewidth][c]{\includegraphics[width=\vsframe]{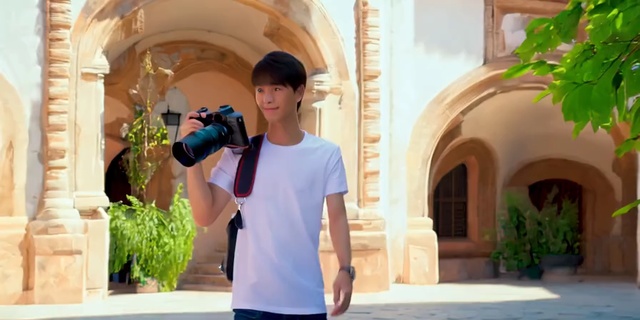}\hspace{\vsgap}\includegraphics[width=\vsframe]{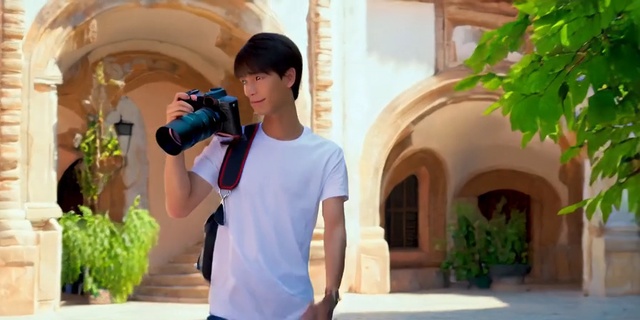}\hspace{\vsgap}\includegraphics[width=\vsframe]{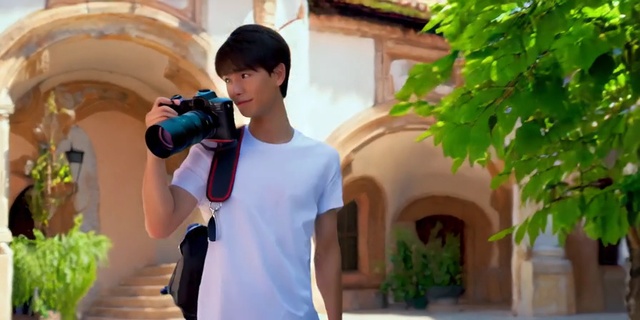}\hspace{\vsgap}\includegraphics[width=\vsframe]{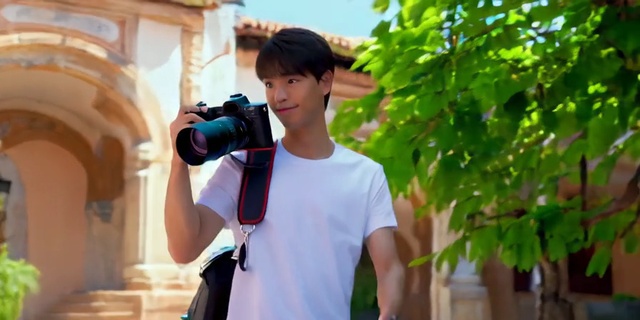}\hspace{\vsgap}\includegraphics[width=\vsframe]{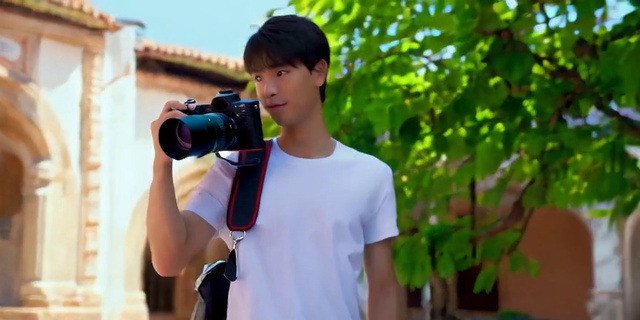}\hspace{\vsgap}\includegraphics[width=\vsframe]{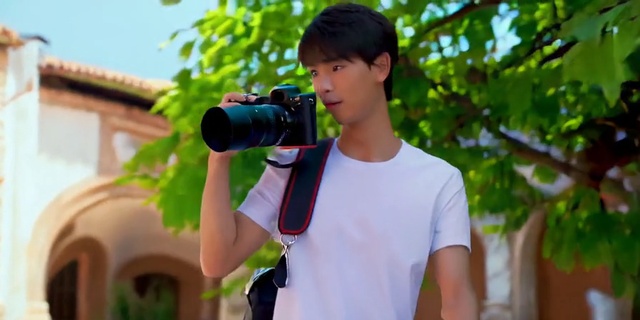}}\\
    [3pt]{\small\textbf{JetDiffusion (Ours)}}\\[1pt]
    \makebox[\linewidth][c]{\includegraphics[width=\vsframe]{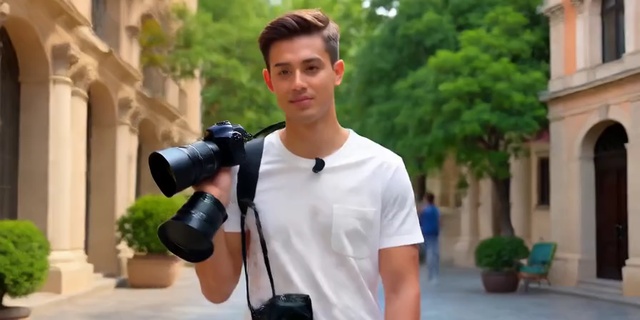}\hspace{\vsgap}\includegraphics[width=\vsframe]{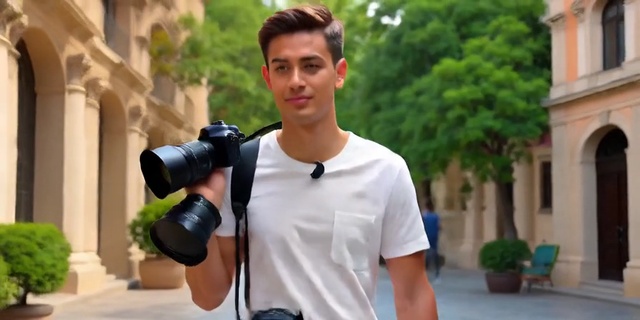}\hspace{\vsgap}\includegraphics[width=\vsframe]{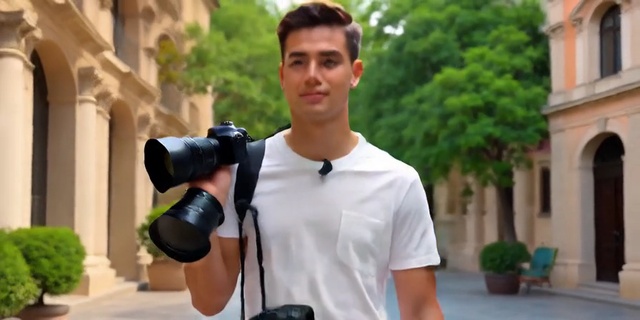}\hspace{\vsgap}\includegraphics[width=\vsframe]{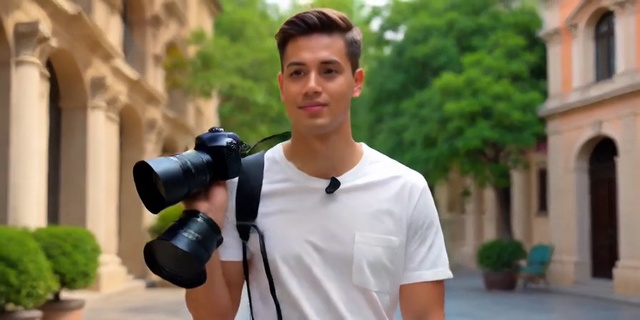}\hspace{\vsgap}\includegraphics[width=\vsframe]{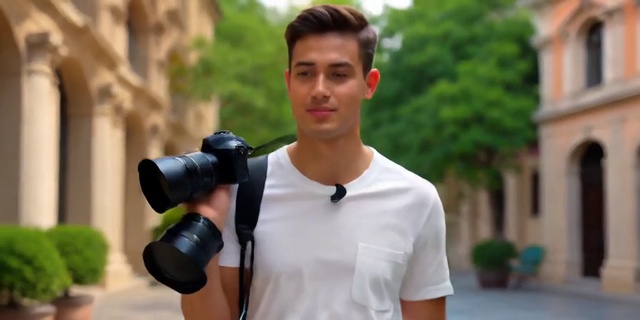}\hspace{\vsgap}\includegraphics[width=\vsframe]{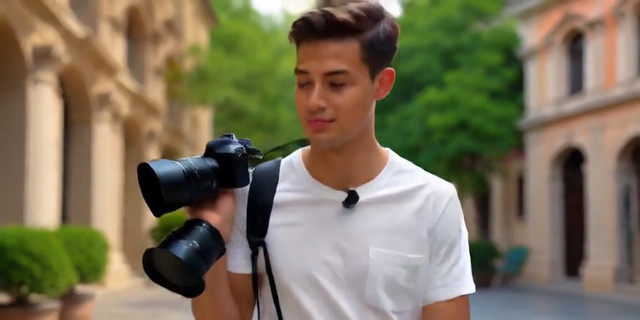}}\\
    \caption{Video generation with Wan2.1-T2V-1.3B. Top: the base model. Bottom: JetDiffusion.
    Text prompt: \textit{``A young man walks confidently through a picturesque plaza, clutching a professional camera over his shoulder. Dressed in a simple white t-shirt, his casual attire contrasts with the sophisticated camera equipment he carries, hinting at his expertise. The background is filled with charming historical architecture and lush greenery, suggesting a vibrant, bustling environment. The focus on the man's purposeful stride and the detailed lens of the camera creates an atmosphere of dynamic exploration and artistic intention.''}}
    \label{fig:video_wan13b_s04}
\end{figure}

\begin{figure}[!htb]
    \centering\lineskip=1pt
    {\small\textbf{Wan2.1-T2V-1.3B}}\\[1pt]
    \makebox[\linewidth][c]{\includegraphics[width=\vsframe]{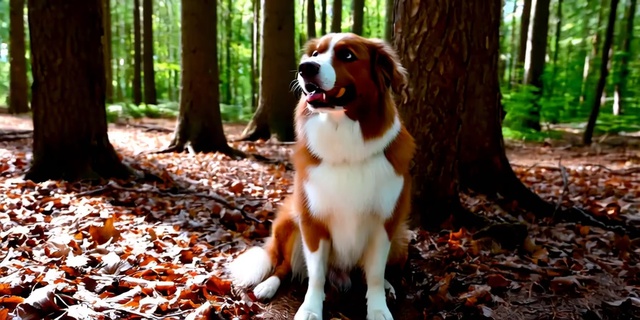}\hspace{\vsgap}\includegraphics[width=\vsframe]{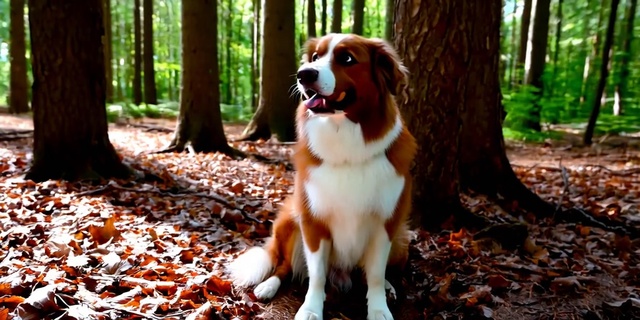}\hspace{\vsgap}\includegraphics[width=\vsframe]{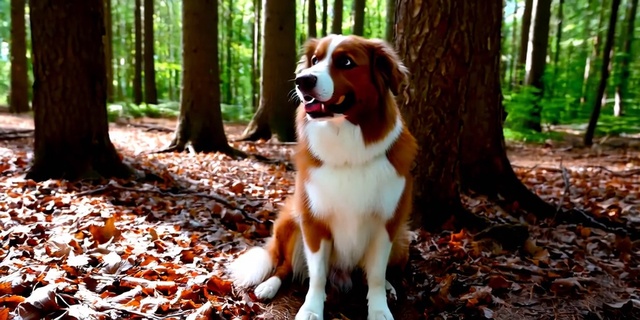}\hspace{\vsgap}\includegraphics[width=\vsframe]{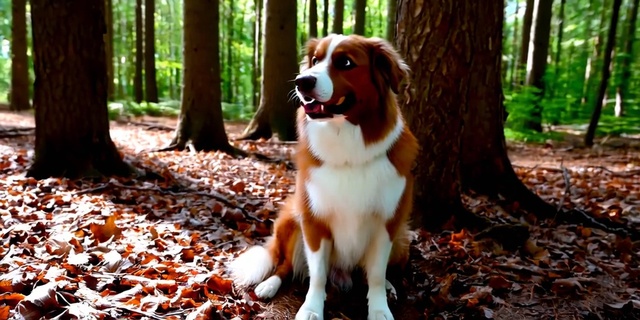}\hspace{\vsgap}\includegraphics[width=\vsframe]{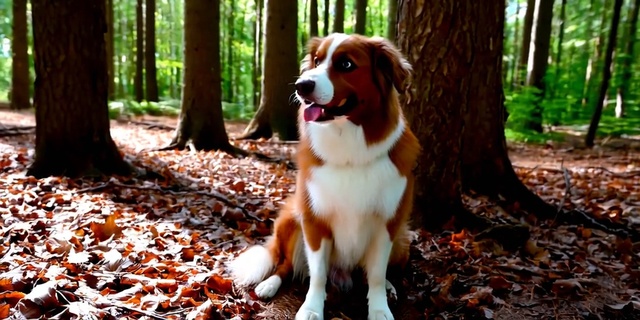}\hspace{\vsgap}\includegraphics[width=\vsframe]{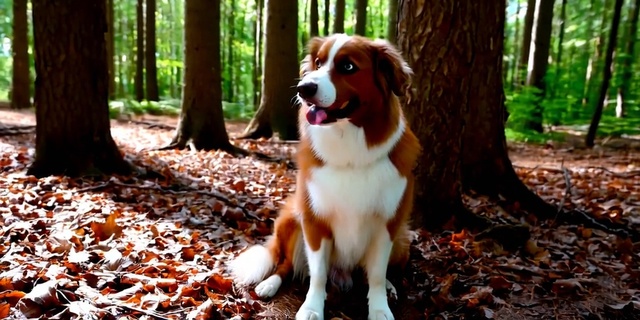}}\\
    [3pt]{\small\textbf{JetDiffusion (Ours)}}\\[1pt]
    \makebox[\linewidth][c]{\includegraphics[width=\vsframe]{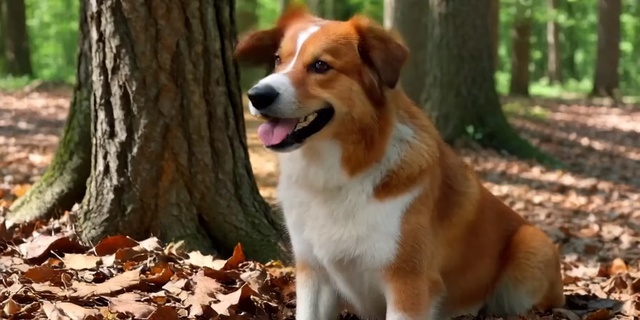}\hspace{\vsgap}\includegraphics[width=\vsframe]{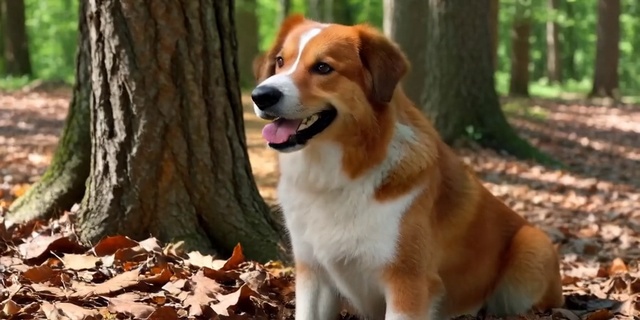}\hspace{\vsgap}\includegraphics[width=\vsframe]{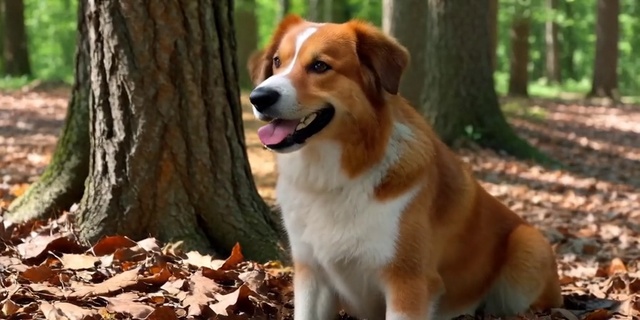}\hspace{\vsgap}\includegraphics[width=\vsframe]{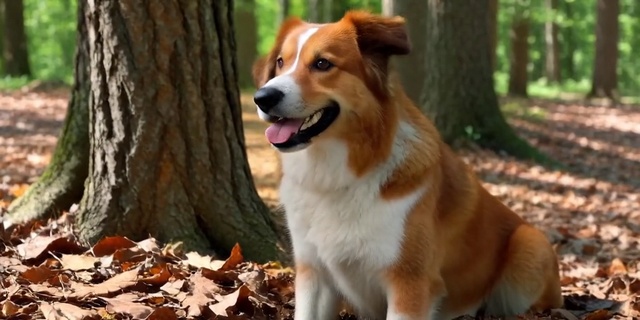}\hspace{\vsgap}\includegraphics[width=\vsframe]{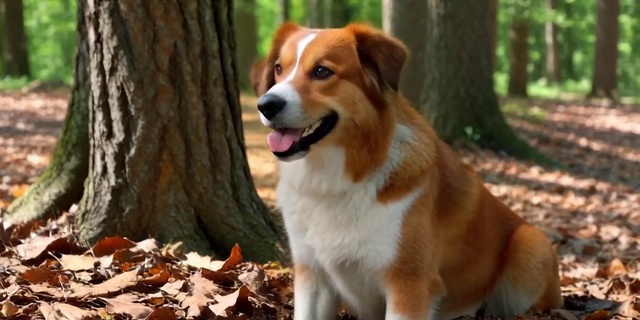}\hspace{\vsgap}\includegraphics[width=\vsframe]{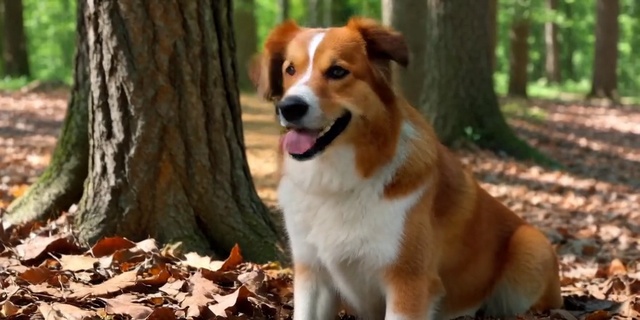}}\\
    \caption{Video generation with Wan2.1-T2V-1.3B. Top: the base model. Bottom: JetDiffusion.
    Text prompt: \textit{``In a serene forest setting, a brown and white dog with expressive eyes sits amidst a carpet of dry leaves. The tall trees create a peaceful and shaded ambiance, with sunlight filtering through the canopy. The dog is alert and attentive, as if listening to distant sounds, with its mouth slightly open revealing a playful expression. Its fur is fluffy and well-groomed, contrasting with the rough textures of the tree bark and forest floor. The surrounding tranquility enhances the focus on the dog, making it the central figure of this natural scene.''}}
    \label{fig:video_wan13b_s05}
\end{figure}

\begin{figure}[!htb]
    \centering\lineskip=1pt
    {\small\textbf{Wan2.1-T2V-1.3B}}\\[1pt]
    \makebox[\linewidth][c]{\includegraphics[width=\vsframe]{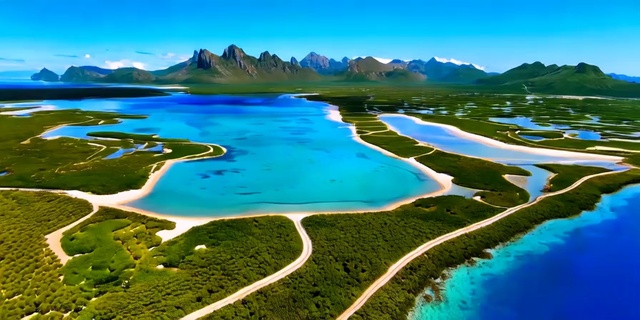}\hspace{\vsgap}\includegraphics[width=\vsframe]{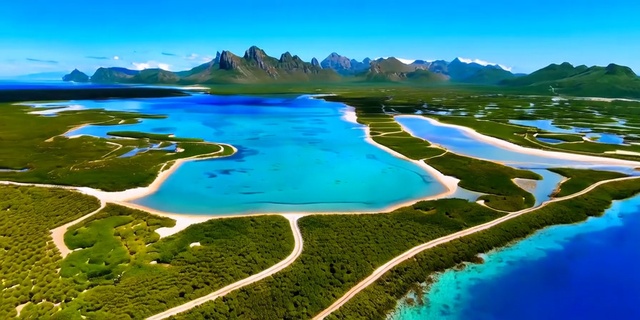}\hspace{\vsgap}\includegraphics[width=\vsframe]{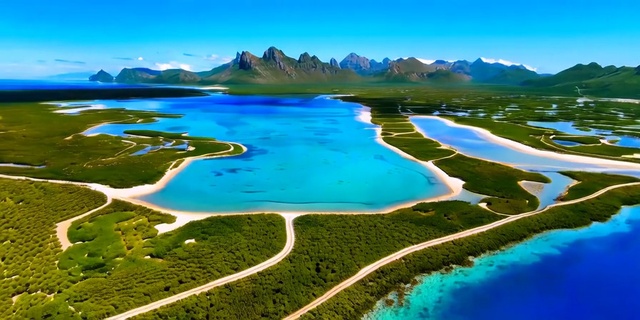}\hspace{\vsgap}\includegraphics[width=\vsframe]{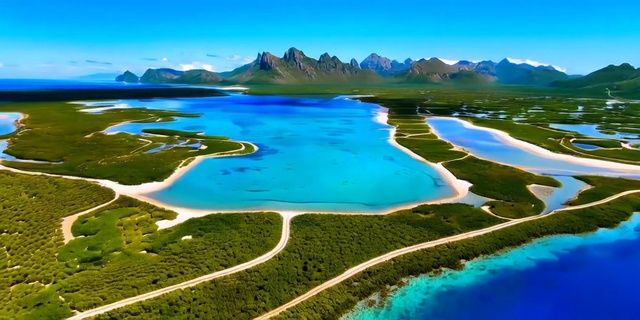}\hspace{\vsgap}\includegraphics[width=\vsframe]{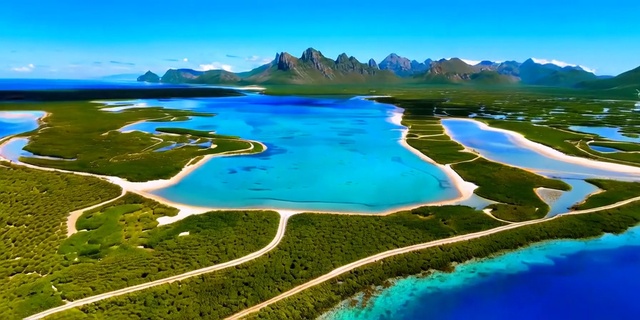}\hspace{\vsgap}\includegraphics[width=\vsframe]{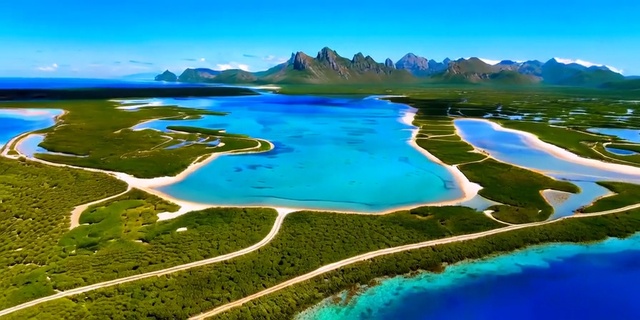}}\\
    [3pt]{\small\textbf{JetDiffusion (Ours)}}\\[1pt]
    \makebox[\linewidth][c]{\includegraphics[width=\vsframe]{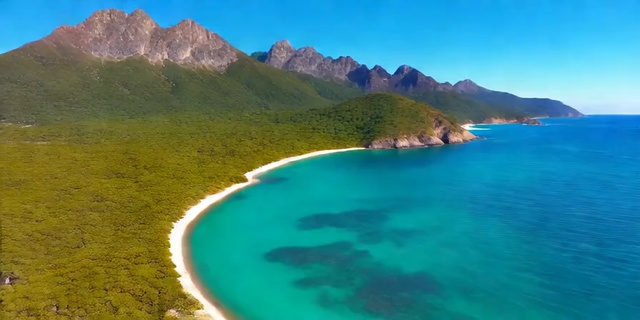}\hspace{\vsgap}\includegraphics[width=\vsframe]{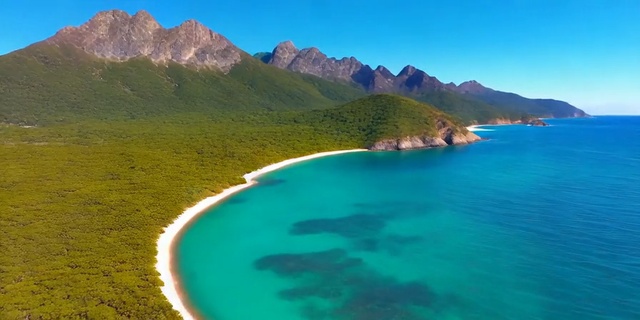}\hspace{\vsgap}\includegraphics[width=\vsframe]{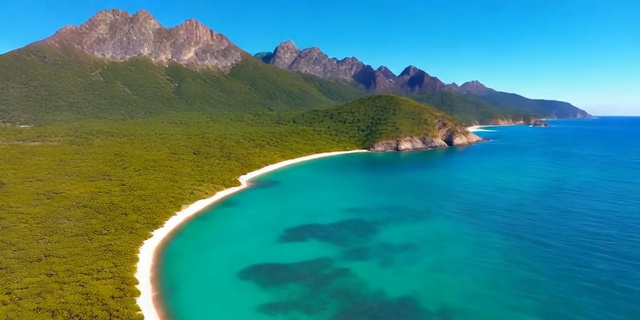}\hspace{\vsgap}\includegraphics[width=\vsframe]{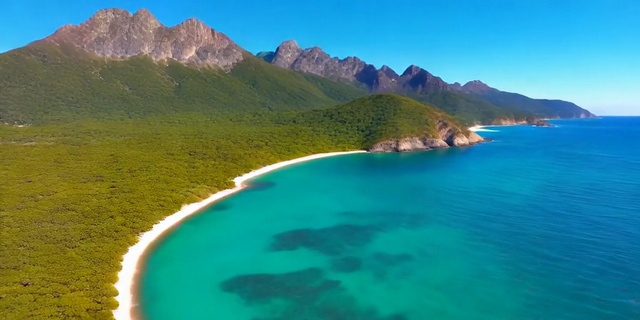}\hspace{\vsgap}\includegraphics[width=\vsframe]{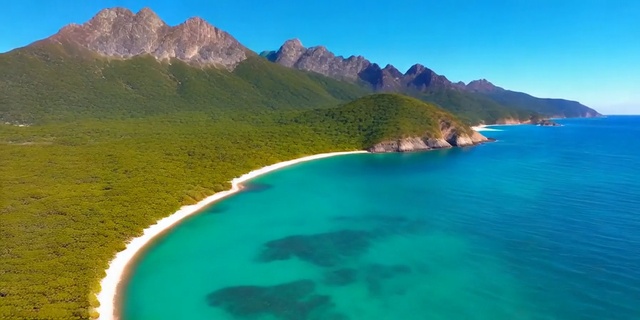}\hspace{\vsgap}\includegraphics[width=\vsframe]{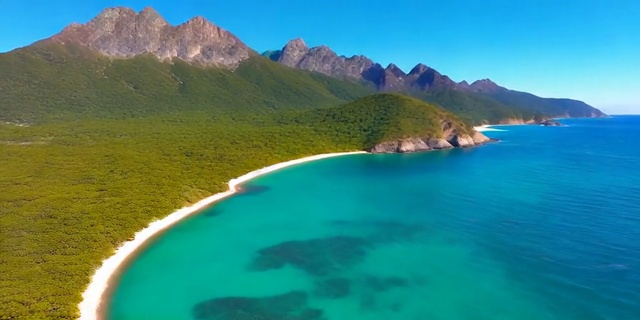}}\\
    \caption{Video generation with Wan2.1-T2V-1.3B. Top: the base model. Bottom: JetDiffusion.
    Text prompt: \textit{``Sweeping over a breathtaking coastal landscape, the video captures a stunning aerial view of a secluded bay embraced by rugged mountains. The turquoise waters gently lap against expansive sandy shores that display intricate patterns shaped by the tides. Surrounding the bay, lush green mangroves and winding inlets create a patchwork of rich textures and vibrant colors. In the background, majestic mountains rise with jagged peaks casting shadows over the serene landscape. The clear blue sky above adds a layer of tranquility, accentuating the untouched beauty of this remote paradise. The scene is a harmonious blend of natural elements, portraying a tranquil and isolated place that invites exploration and reflection.''}}
    \label{fig:video_wan13b_s06}
\end{figure}

\FloatBarrier

\subsection{Wan2.1-T2V-14B}

\begin{figure}[!htb]
    \centering\lineskip=1pt
    {\small\textbf{Wan2.1-T2V-14B}}\\[1pt]
    \makebox[\linewidth][c]{\includegraphics[width=\vsframe]{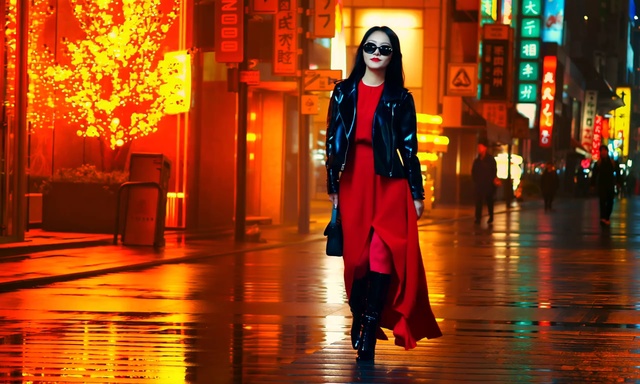}\hspace{\vsgap}\includegraphics[width=\vsframe]{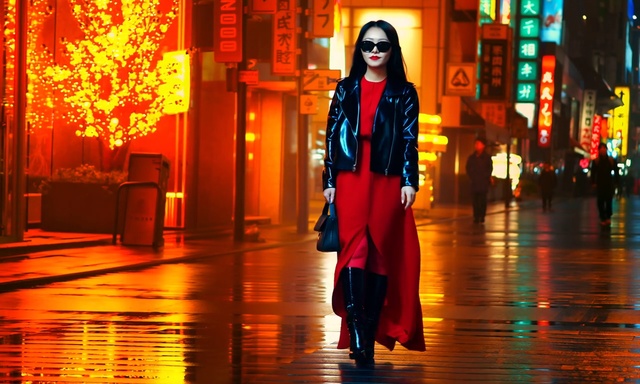}\hspace{\vsgap}\includegraphics[width=\vsframe]{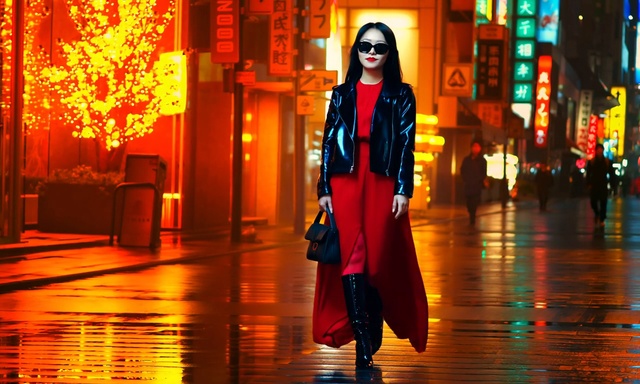}\hspace{\vsgap}\includegraphics[width=\vsframe]{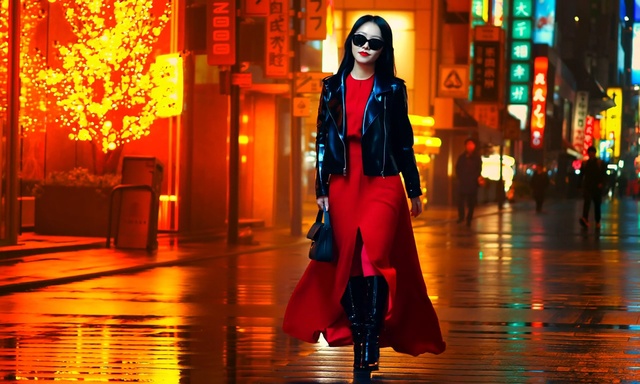}\hspace{\vsgap}\includegraphics[width=\vsframe]{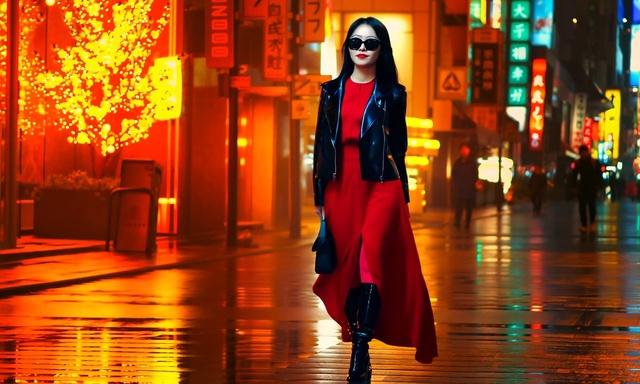}\hspace{\vsgap}\includegraphics[width=\vsframe]{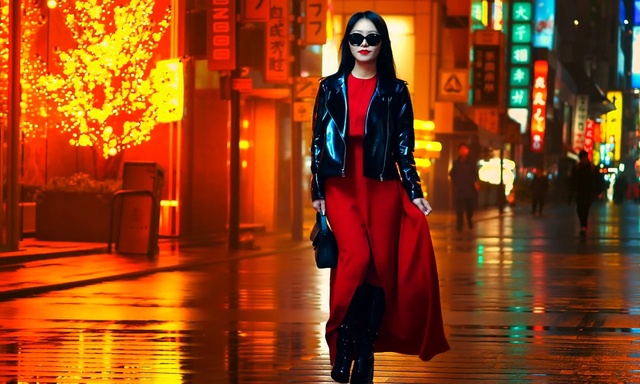}}\\
    [3pt]{\small\textbf{JetDiffusion (Ours)}}\\[1pt]
    \makebox[\linewidth][c]{\includegraphics[width=\vsframe]{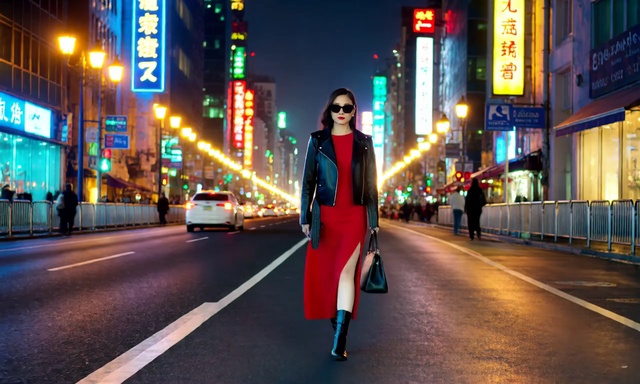}\hspace{\vsgap}\includegraphics[width=\vsframe]{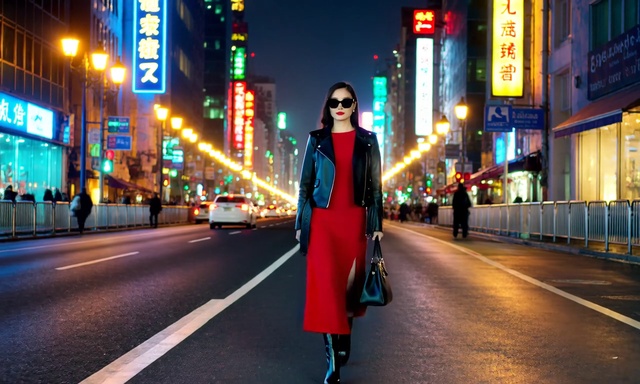}\hspace{\vsgap}\includegraphics[width=\vsframe]{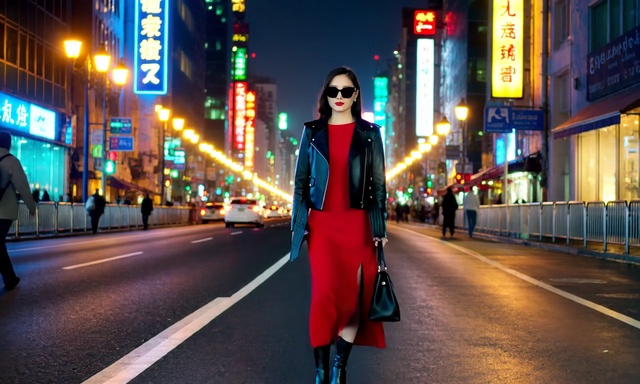}\hspace{\vsgap}\includegraphics[width=\vsframe]{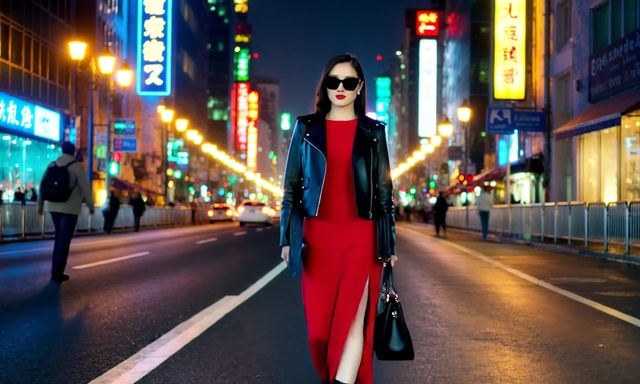}\hspace{\vsgap}\includegraphics[width=\vsframe]{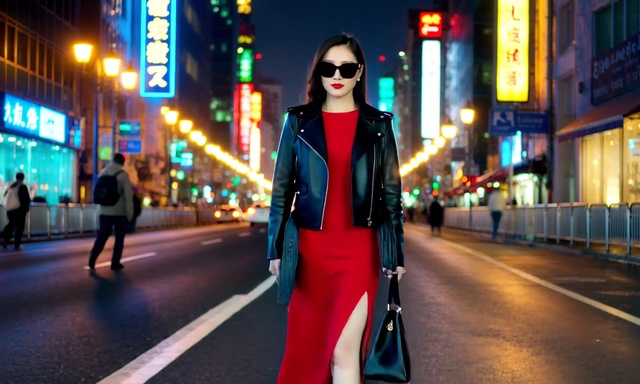}\hspace{\vsgap}\includegraphics[width=\vsframe]{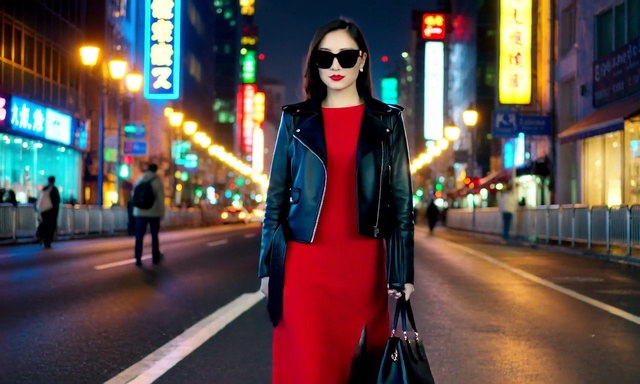}}\\
    \caption{Video generation with Wan2.1-T2V-14B. Top: the base model. Bottom: JetDiffusion.
    Text prompt: \textit{``A stylish woman walks down a Tokyo street filled with warm glowing neon and animated city signage. She wears a black leather jacket, a long red dress, and black boots, and carries a black purse. She wears sunglasses and red lipstick. She walks confidently and casually. The street is damp and reflective, creating a mirror effect of the colorful lights. Many pedestrians walk about.''}}
    \label{fig:video_wan14b_s01}
\end{figure}

\begin{figure}[!htb]
    \centering\lineskip=1pt
    {\small\textbf{Wan2.1-T2V-14B}}\\[1pt]
    \makebox[\linewidth][c]{\includegraphics[width=\vsframe]{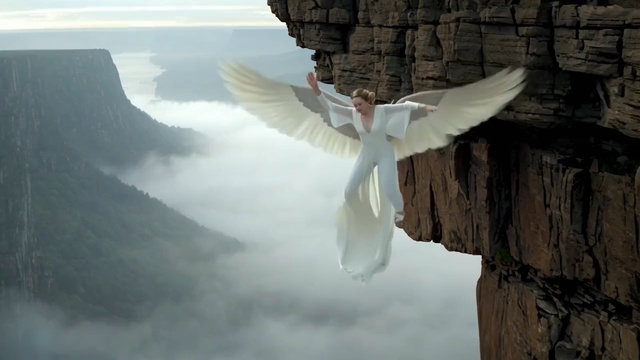}\hspace{\vsgap}\includegraphics[width=\vsframe]{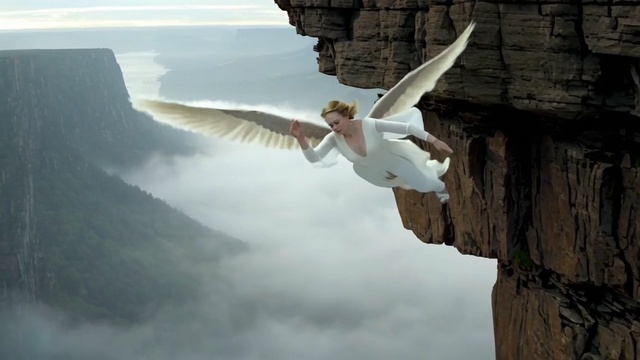}\hspace{\vsgap}\includegraphics[width=\vsframe]{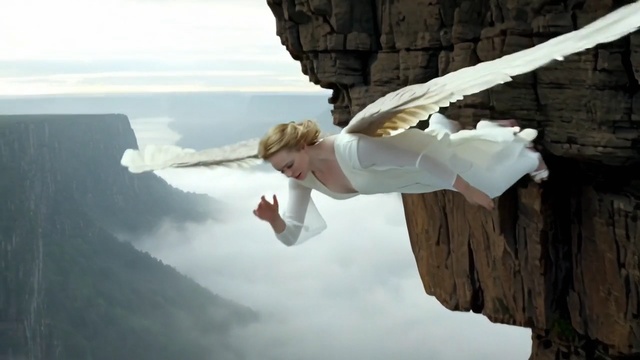}\hspace{\vsgap}\includegraphics[width=\vsframe]{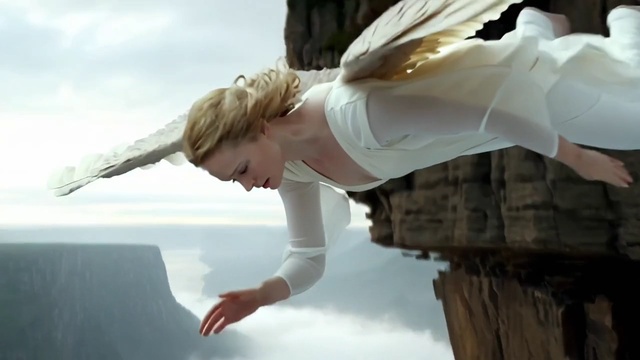}\hspace{\vsgap}\includegraphics[width=\vsframe]{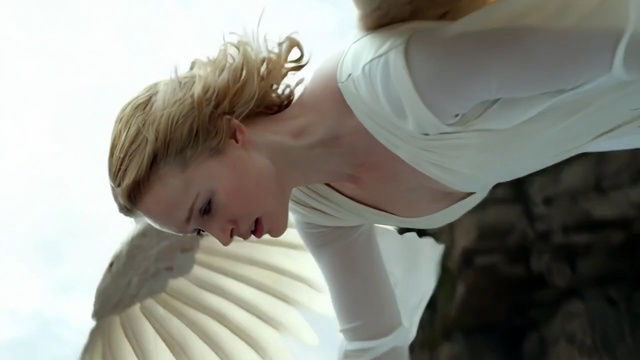}\hspace{\vsgap}\includegraphics[width=\vsframe]{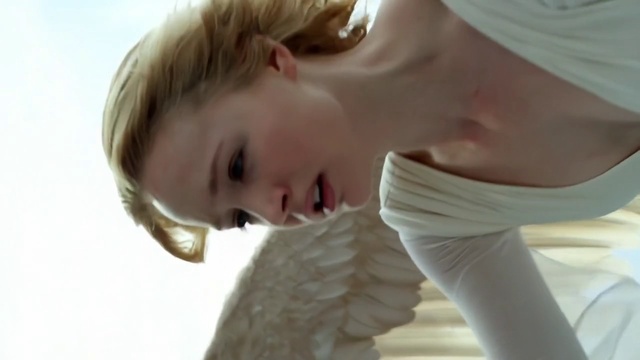}}\\
    [3pt]{\small\textbf{JetDiffusion (Ours)}}\\[1pt]
    \makebox[\linewidth][c]{\includegraphics[width=\vsframe]{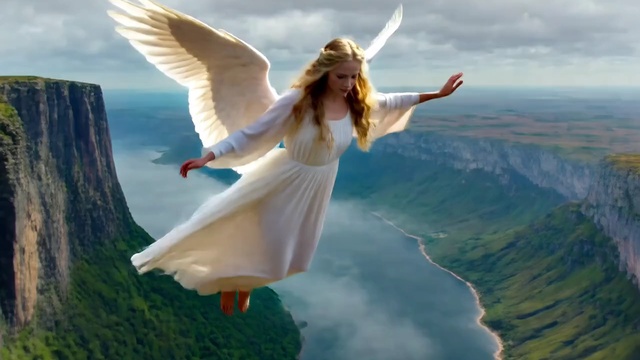}\hspace{\vsgap}\includegraphics[width=\vsframe]{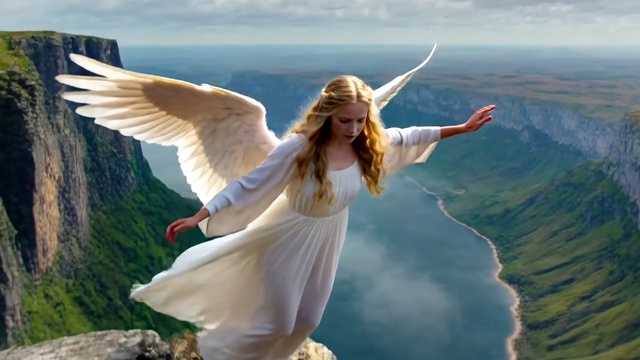}\hspace{\vsgap}\includegraphics[width=\vsframe]{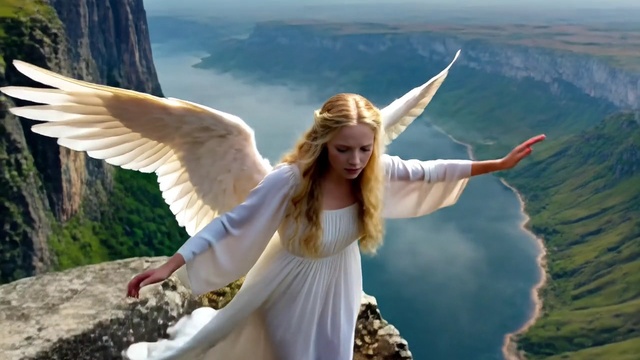}\hspace{\vsgap}\includegraphics[width=\vsframe]{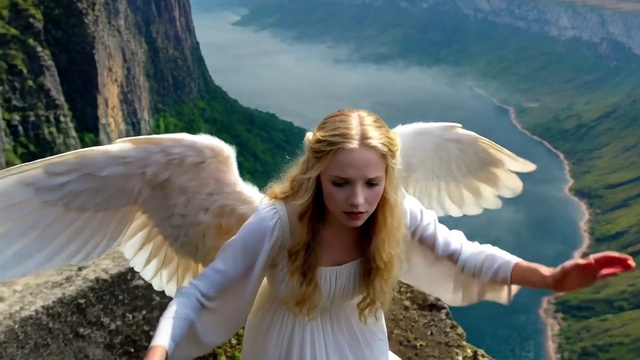}\hspace{\vsgap}\includegraphics[width=\vsframe]{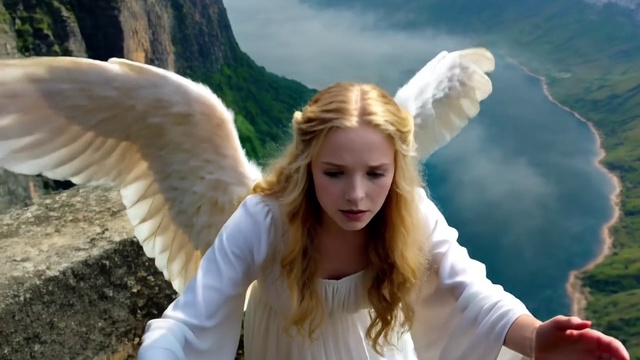}\hspace{\vsgap}\includegraphics[width=\vsframe]{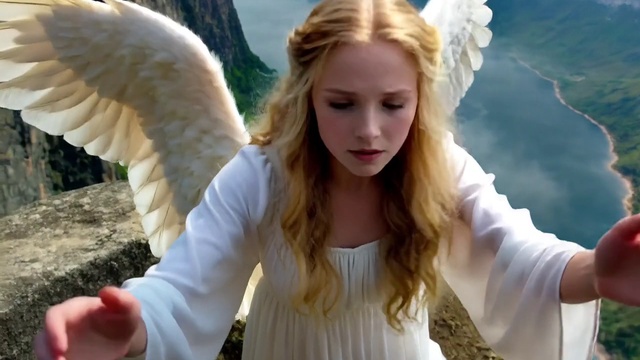}}\\
    \caption{Video generation with Wan2.1-T2V-14B. Top: the base model. Bottom: JetDiffusion.
    Text prompt: \textit{``A dramatic scene of an angelic figure falling off a cliff, with flowing white robes and ethereal wings. The angel has fair skin, blonde hair flowing freely, and a serene yet sorrowful expression. The background showcases rugged cliffs and a vast, misty valley below. The camera starts with a wide shot of the cliff edge and then smoothly zooms in on the angel as they begin their descent, capturing the intense emotion and the breathtaking landscape. Close-up on the angel's face and hands as they reach out, emphasizing the fall's gravity and the angel's vulnerability.''}}
    \label{fig:video_wan14b_s02}
\end{figure}

\begin{figure}[!htb]
    \centering\lineskip=1pt
    {\small\textbf{Wan2.1-T2V-14B}}\\[1pt]
    \makebox[\linewidth][c]{\includegraphics[width=\vsframe]{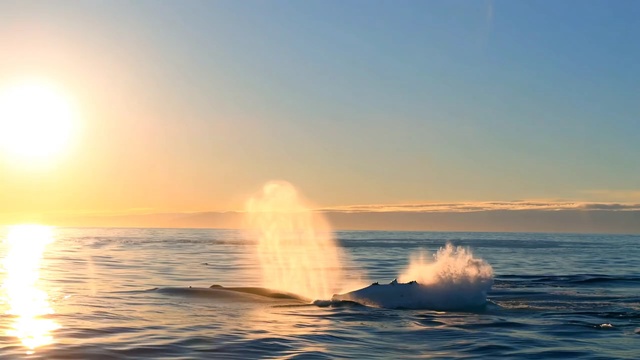}\hspace{\vsgap}\includegraphics[width=\vsframe]{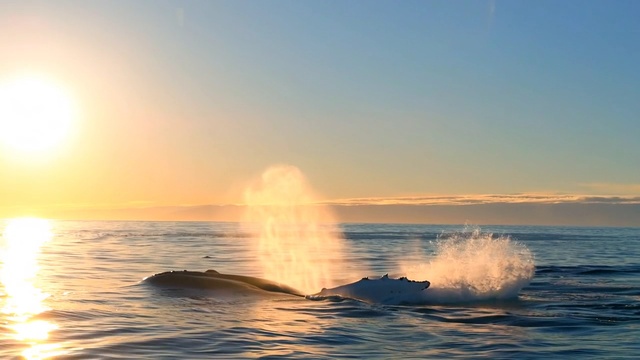}\hspace{\vsgap}\includegraphics[width=\vsframe]{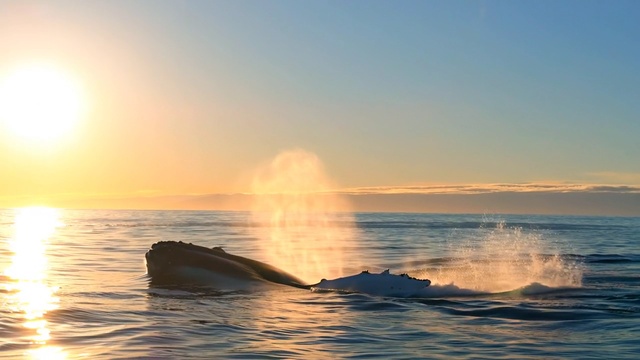}\hspace{\vsgap}\includegraphics[width=\vsframe]{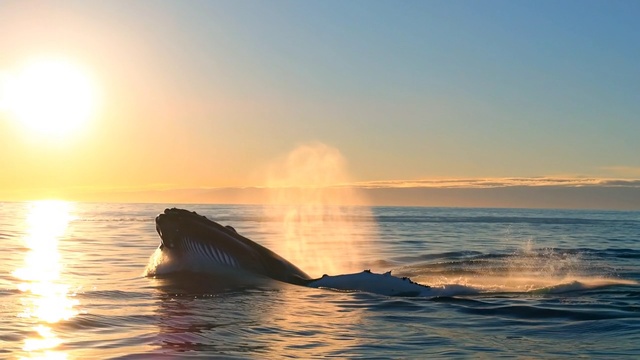}\hspace{\vsgap}\includegraphics[width=\vsframe]{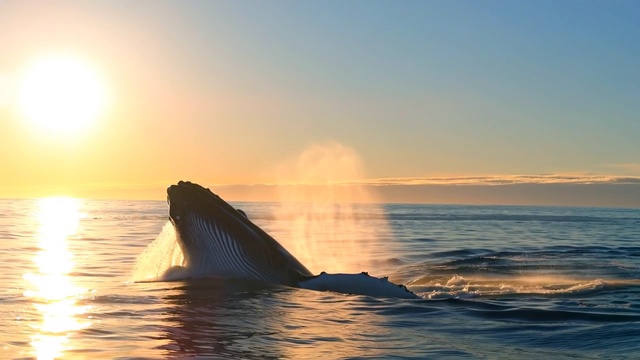}\hspace{\vsgap}\includegraphics[width=\vsframe]{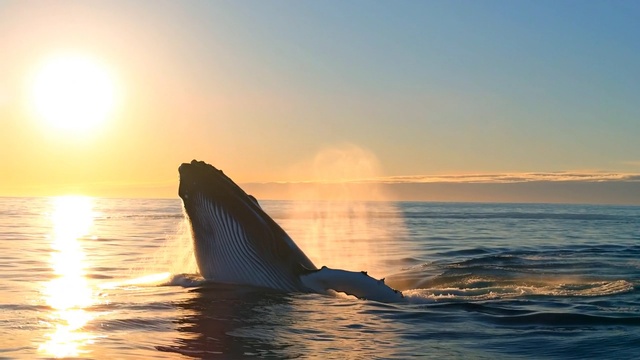}}\\
    [3pt]{\small\textbf{JetDiffusion (Ours)}}\\[1pt]
    \makebox[\linewidth][c]{\includegraphics[width=\vsframe]{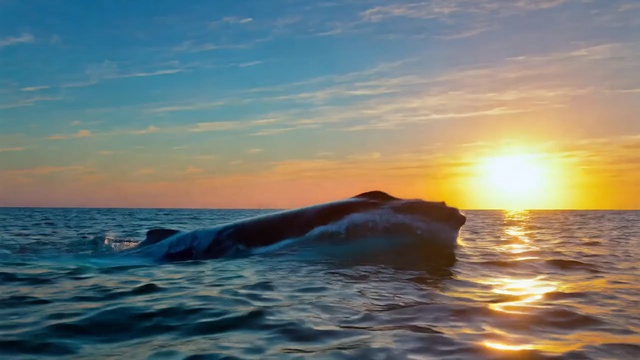}\hspace{\vsgap}\includegraphics[width=\vsframe]{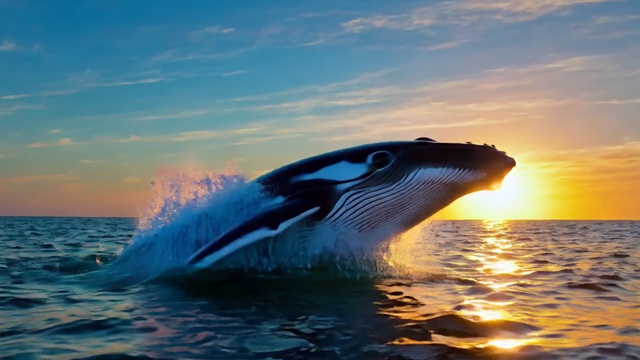}\hspace{\vsgap}\includegraphics[width=\vsframe]{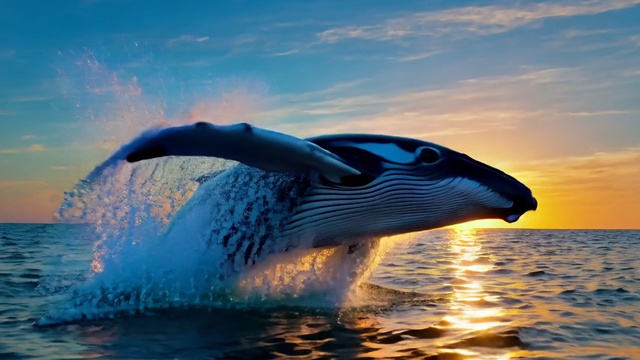}\hspace{\vsgap}\includegraphics[width=\vsframe]{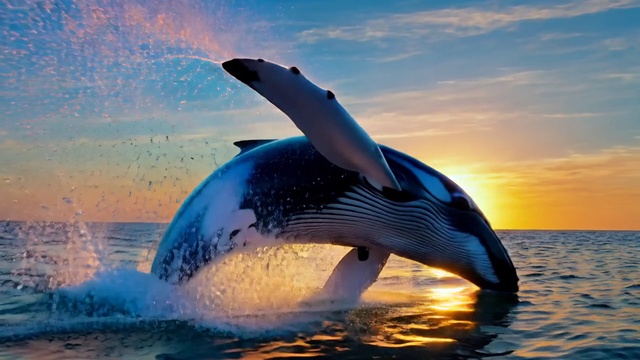}\hspace{\vsgap}\includegraphics[width=\vsframe]{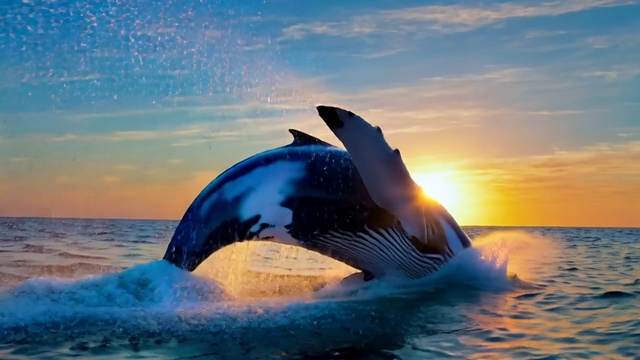}\hspace{\vsgap}\includegraphics[width=\vsframe]{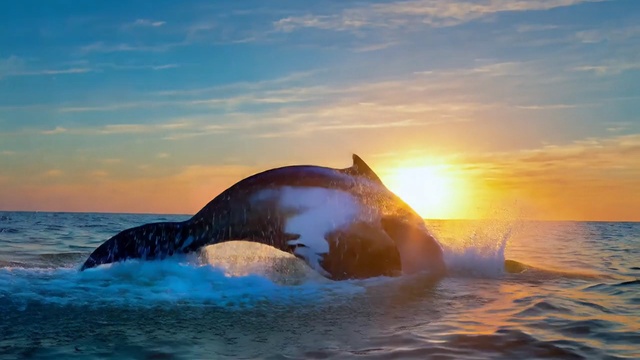}}\\
    \caption{Video generation with Wan2.1-T2V-14B. Top: the base model. Bottom: JetDiffusion.
    Text prompt: \textit{``A majestic humpback whale emerging from the ocean, breaching the surface with its massive body. The whale's sleek, gray body arcs gracefully as it leaps out of the water, creating a splash that sends droplets scattering into the air. The sun is setting, casting a warm golden glow over the ocean, reflecting off the waves and highlighting the spray from the whale's movement. The background shows a vast, open sea with distant rolling waves and a clear blue sky transitioning into shades of orange and pink. Wide shot, capturing the grandeur of the moment.''}}
    \label{fig:video_wan14b_s03}
\end{figure}

\begin{figure}[!htb]
    \centering\lineskip=1pt
    {\small\textbf{Wan2.1-T2V-14B}}\\[1pt]
    \makebox[\linewidth][c]{\includegraphics[width=\vsframe]{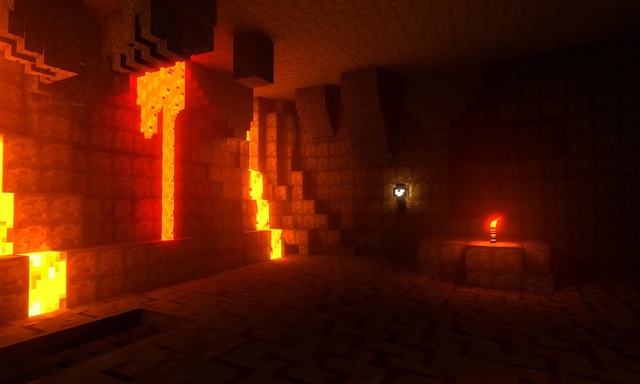}\hspace{\vsgap}\includegraphics[width=\vsframe]{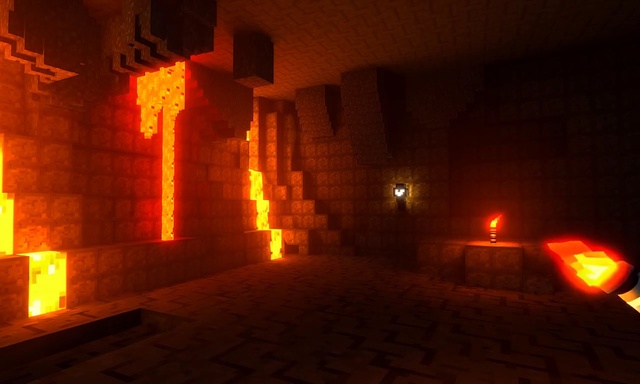}\hspace{\vsgap}\includegraphics[width=\vsframe]{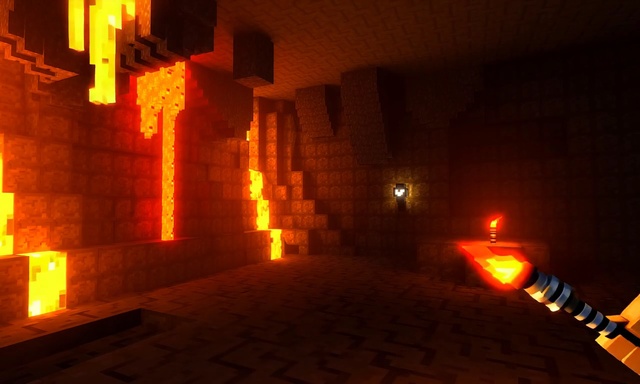}\hspace{\vsgap}\includegraphics[width=\vsframe]{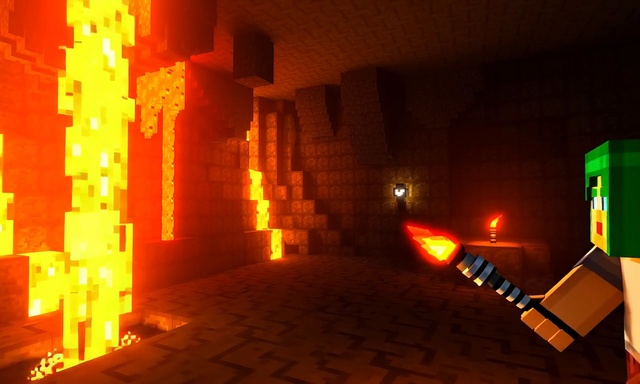}\hspace{\vsgap}\includegraphics[width=\vsframe]{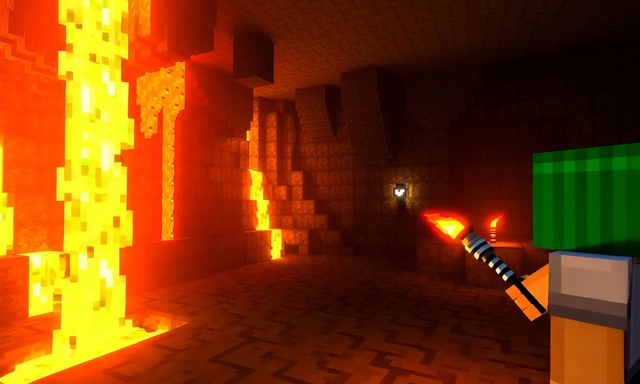}\hspace{\vsgap}\includegraphics[width=\vsframe]{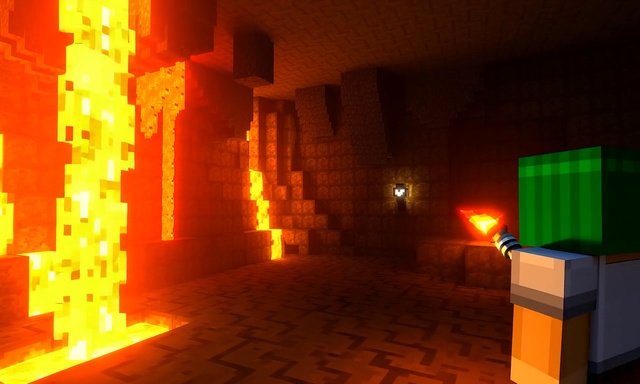}}\\
    [3pt]{\small\textbf{JetDiffusion (Ours)}}\\[1pt]
    \makebox[\linewidth][c]{\includegraphics[width=\vsframe]{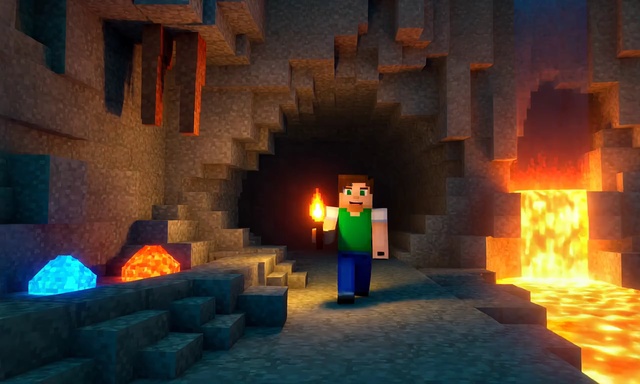}\hspace{\vsgap}\includegraphics[width=\vsframe]{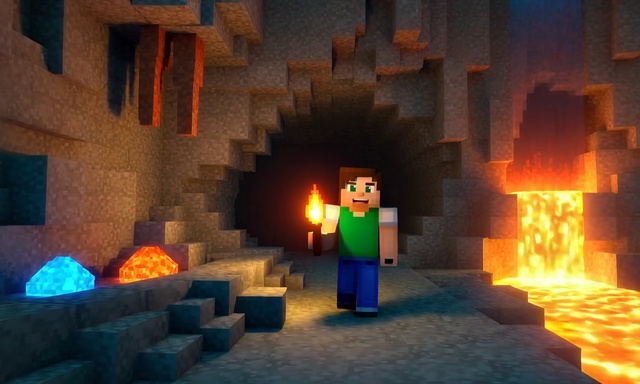}\hspace{\vsgap}\includegraphics[width=\vsframe]{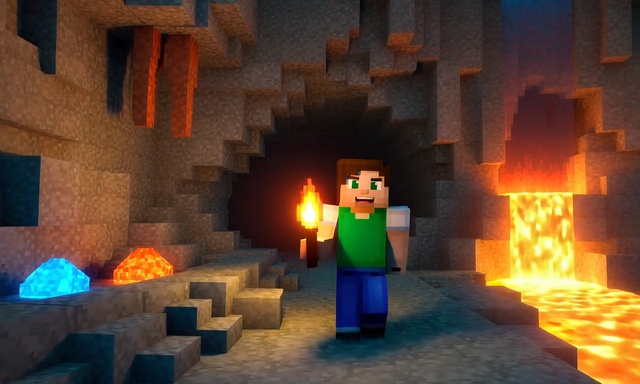}\hspace{\vsgap}\includegraphics[width=\vsframe]{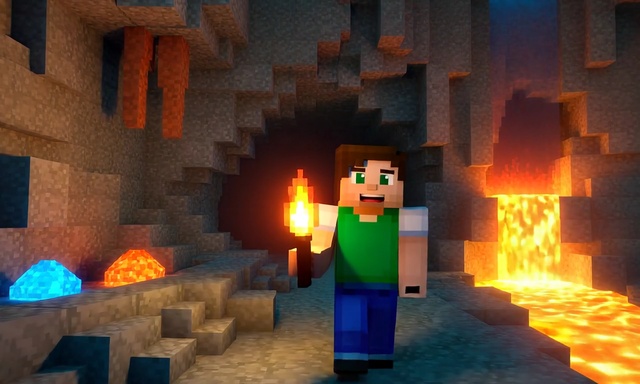}\hspace{\vsgap}\includegraphics[width=\vsframe]{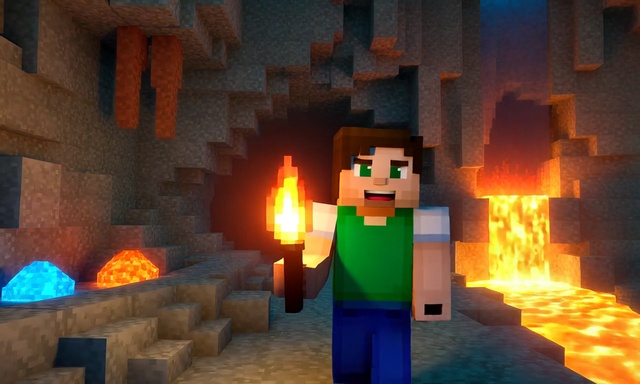}\hspace{\vsgap}\includegraphics[width=\vsframe]{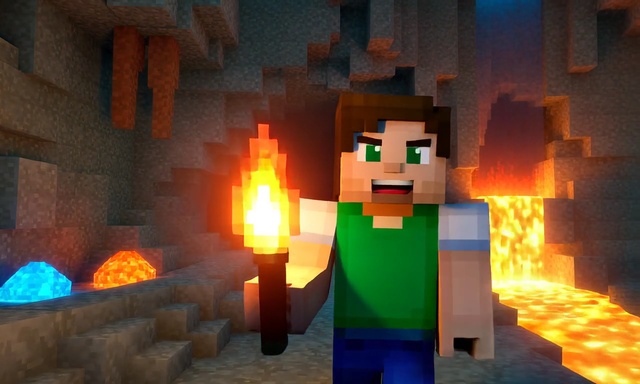}}\\
    \caption{Video generation with Wan2.1-T2V-14B. Top: the base model. Bottom: JetDiffusion.
    Text prompt: \textit{``A Minecraft player character holding a torch enters a massive underground cave. The torchlight flickers against jagged stone walls, illuminating patches of iron and diamond ores embedded in the rock. Stalactites hang from the ceiling, lava flows in glowing streams nearby, and the faint sound of water dripping echoes through the cavern.''}}
    \label{fig:video_wan14b_s04}
\end{figure}

\begin{figure}[!htb]
    \centering\lineskip=1pt
    {\small\textbf{Wan2.1-T2V-14B}}\\[1pt]
    \makebox[\linewidth][c]{\includegraphics[width=\vsframe]{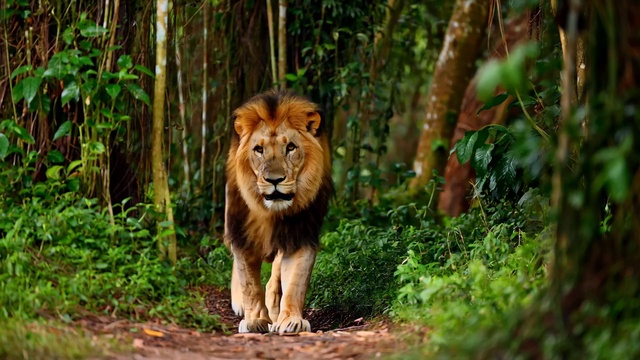}\hspace{\vsgap}\includegraphics[width=\vsframe]{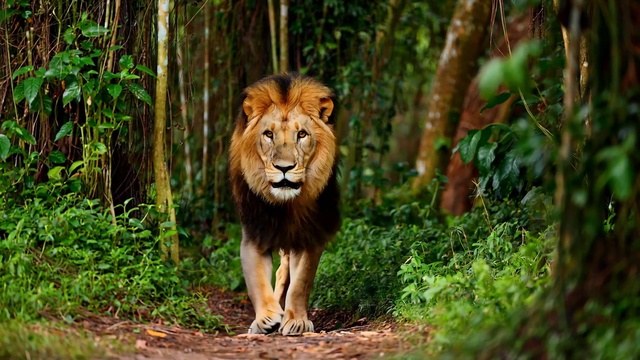}\hspace{\vsgap}\includegraphics[width=\vsframe]{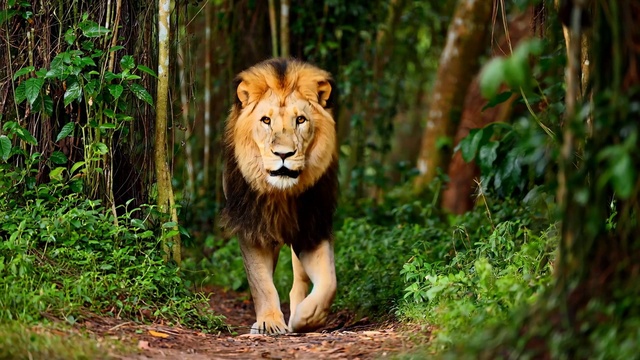}\hspace{\vsgap}\includegraphics[width=\vsframe]{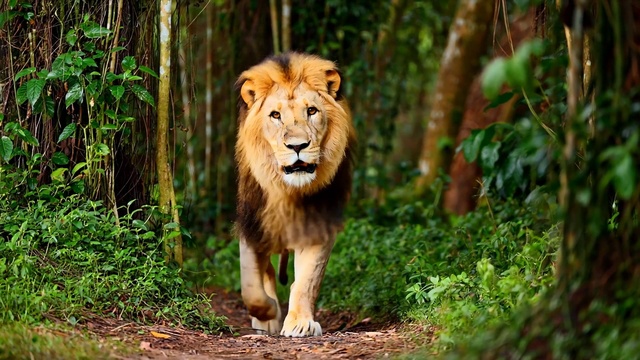}\hspace{\vsgap}\includegraphics[width=\vsframe]{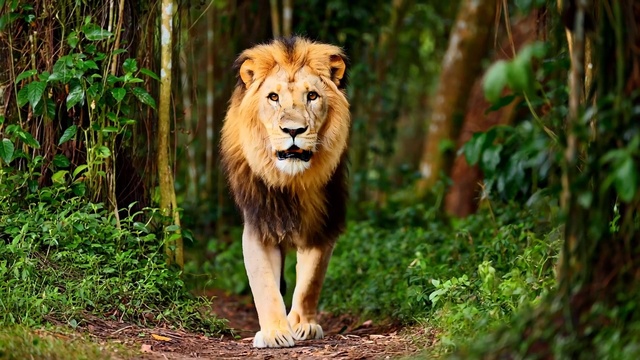}\hspace{\vsgap}\includegraphics[width=\vsframe]{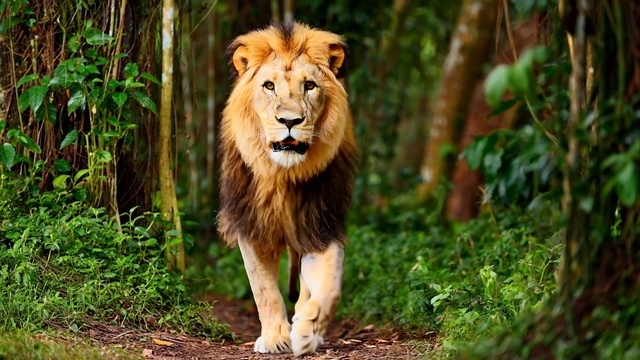}}\\
    [3pt]{\small\textbf{JetDiffusion (Ours)}}\\[1pt]
    \makebox[\linewidth][c]{\includegraphics[width=\vsframe]{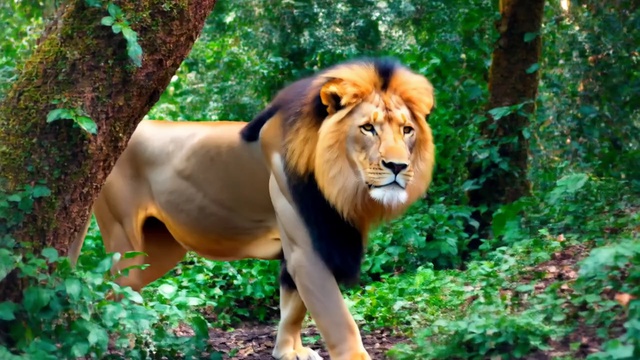}\hspace{\vsgap}\includegraphics[width=\vsframe]{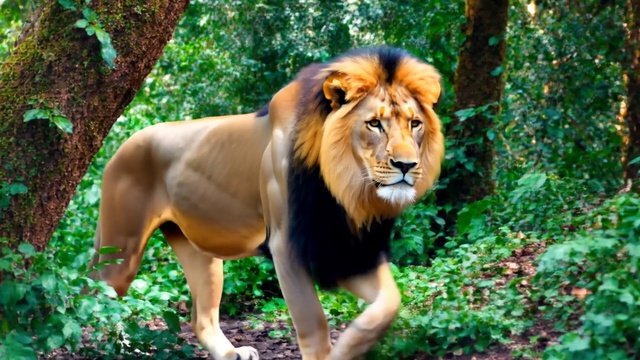}\hspace{\vsgap}\includegraphics[width=\vsframe]{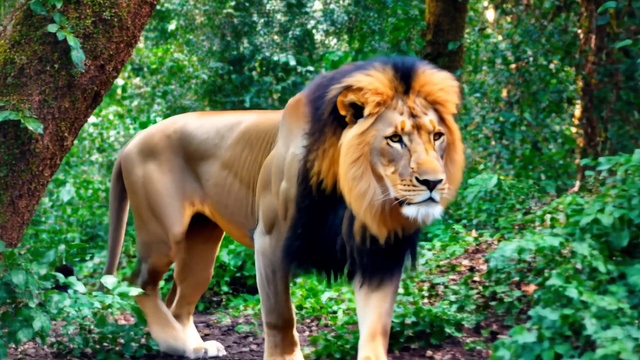}\hspace{\vsgap}\includegraphics[width=\vsframe]{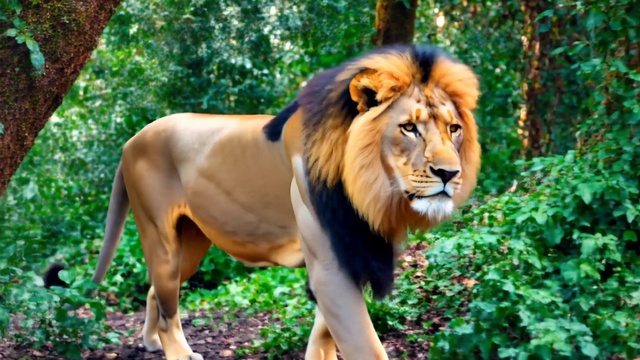}\hspace{\vsgap}\includegraphics[width=\vsframe]{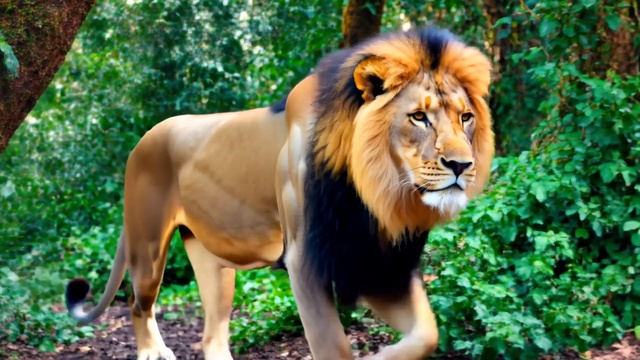}\hspace{\vsgap}\includegraphics[width=\vsframe]{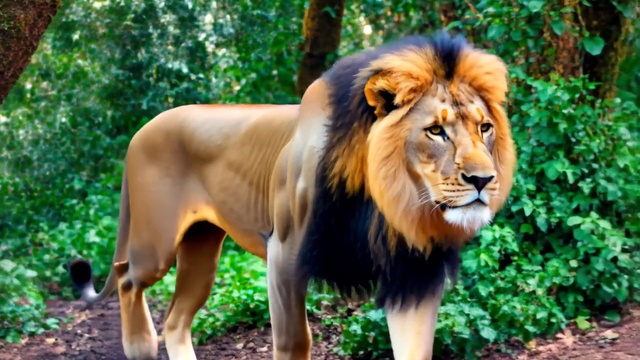}}\\
    \caption{Video generation with Wan2.1-T2V-14B. Top: the base model. Bottom: JetDiffusion.
    Text prompt: \textit{``A majestic African lion walking through a dense, lush forest jungle. The lion has a golden coat with a tawny mane, and its eyes are intense and focused as it moves gracefully through the underbrush. The forest is filled with tall trees, dappled sunlight filtering through the canopy, and various plants and vines that create a vibrant, verdant environment. The lion's muscles ripple under its fur as it strides confidently, embodying power and grace. Medium close-up shot, capturing the lion in motion against the rich backdrop of the jungle.''}}
    \label{fig:video_wan14b_s05}
\end{figure}

\begin{figure}[!htb]
    \centering\lineskip=1pt
    {\small\textbf{Wan2.1-T2V-14B}}\\[1pt]
    \makebox[\linewidth][c]{\includegraphics[width=\vsframe]{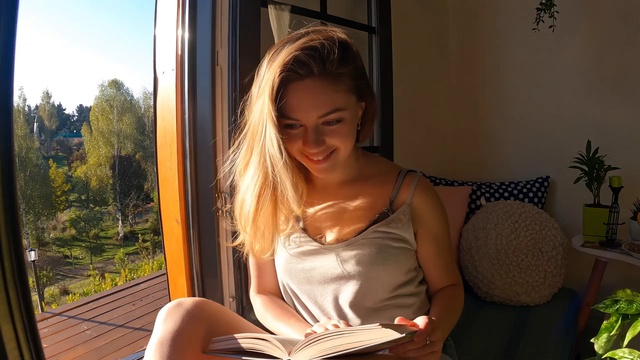}\hspace{\vsgap}\includegraphics[width=\vsframe]{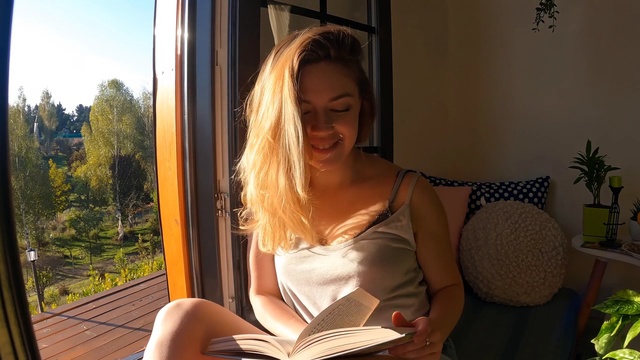}\hspace{\vsgap}\includegraphics[width=\vsframe]{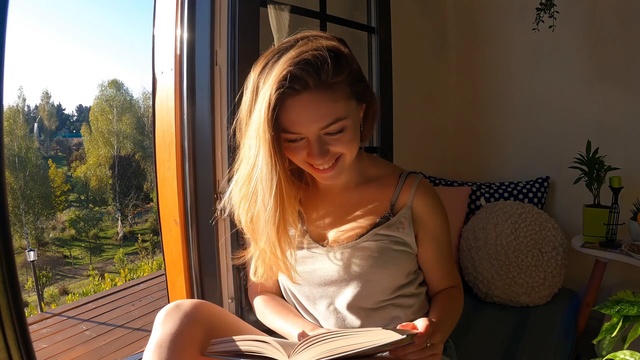}\hspace{\vsgap}\includegraphics[width=\vsframe]{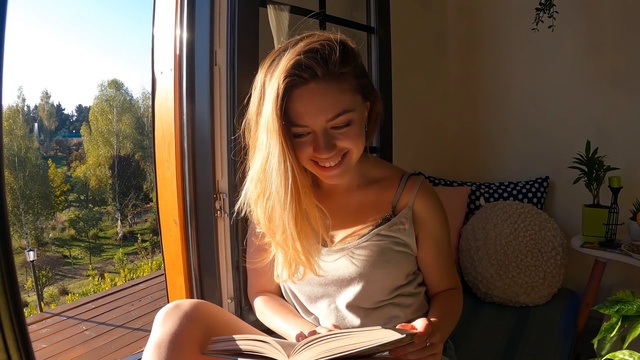}\hspace{\vsgap}\includegraphics[width=\vsframe]{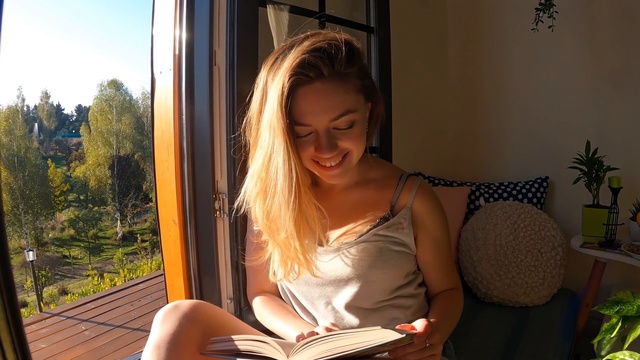}\hspace{\vsgap}\includegraphics[width=\vsframe]{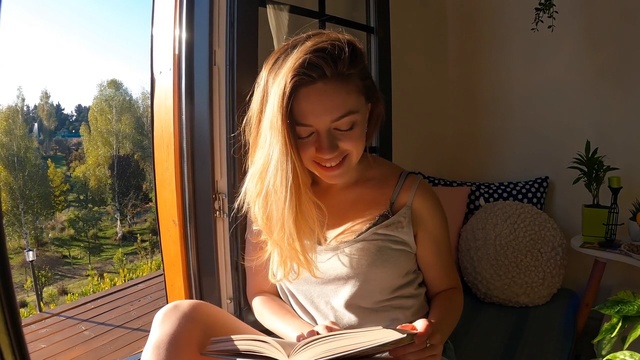}}\\
    [3pt]{\small\textbf{JetDiffusion (Ours)}}\\[1pt]
    \makebox[\linewidth][c]{\includegraphics[width=\vsframe]{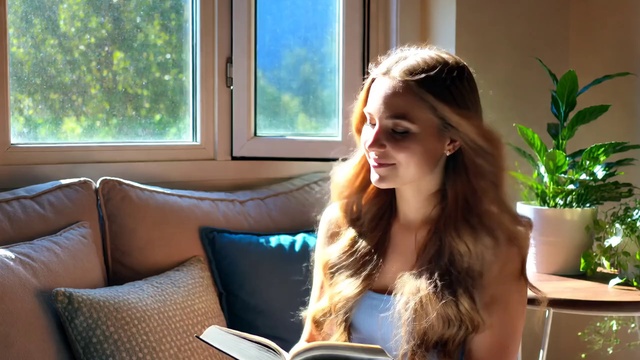}\hspace{\vsgap}\includegraphics[width=\vsframe]{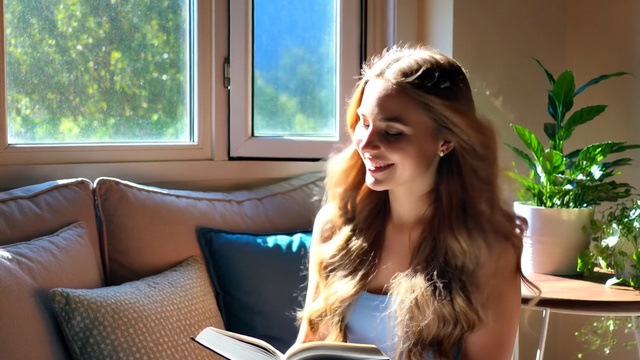}\hspace{\vsgap}\includegraphics[width=\vsframe]{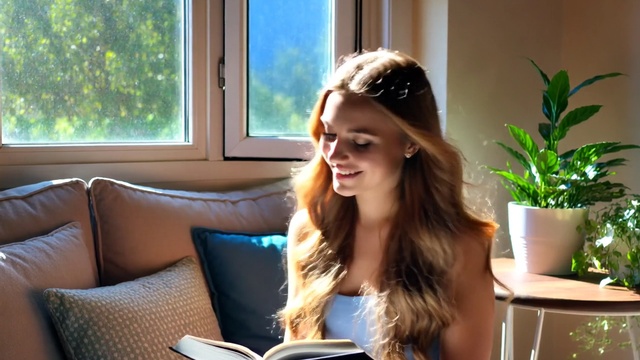}\hspace{\vsgap}\includegraphics[width=\vsframe]{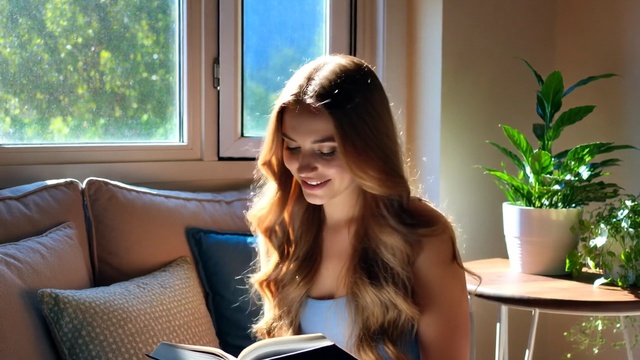}\hspace{\vsgap}\includegraphics[width=\vsframe]{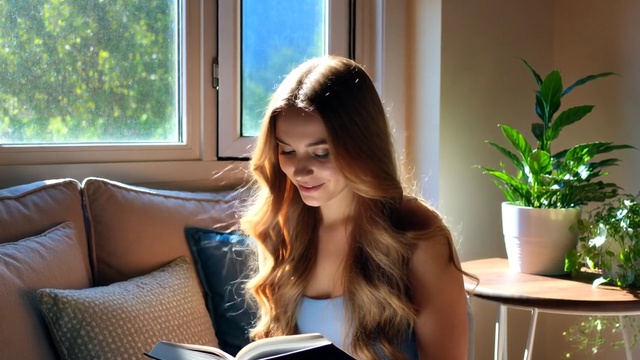}\hspace{\vsgap}\includegraphics[width=\vsframe]{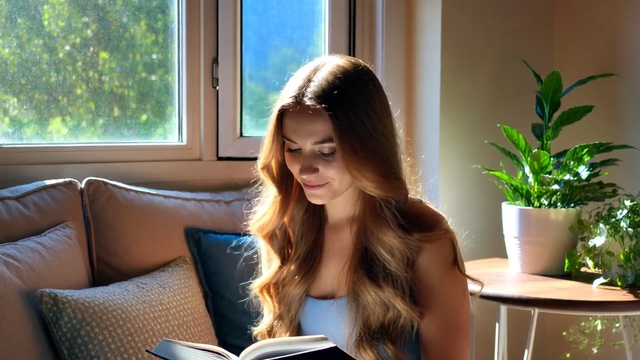}}\\
    \caption{Video generation with Wan2.1-T2V-14B. Top: the base model. Bottom: JetDiffusion.
    Text prompt: \textit{``A young woman with flowing hair sits comfortably next to an open window in a cozy room. Sunlight streams in, casting a warm glow on her. Her hair gently moves as a soft breeze blows through the open window. She is relaxed, with a gentle smile, and is reading a book. The room is decorated with plush cushions, a small table, and various houseplants. The background shows a picturesque outdoor scene with trees and a clear blue sky. Medium close-up shot, focusing on the girl and her interaction with her surroundings.''}}
    \label{fig:video_wan14b_s06}
\end{figure}

\begin{figure}[!htb]
    \centering\lineskip=1pt
    {\small\textbf{Wan2.1-T2V-14B}}\\[1pt]
    \makebox[\linewidth][c]{\includegraphics[width=\vsframe]{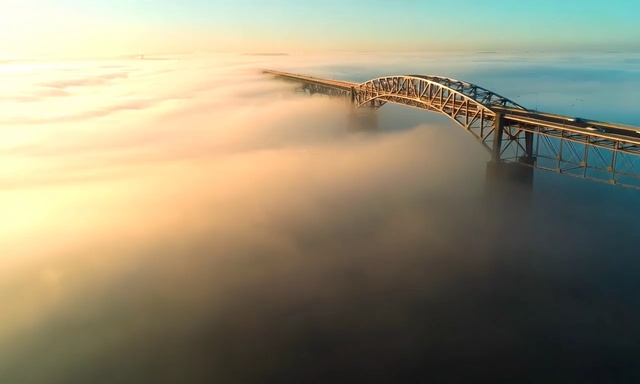}\hspace{\vsgap}\includegraphics[width=\vsframe]{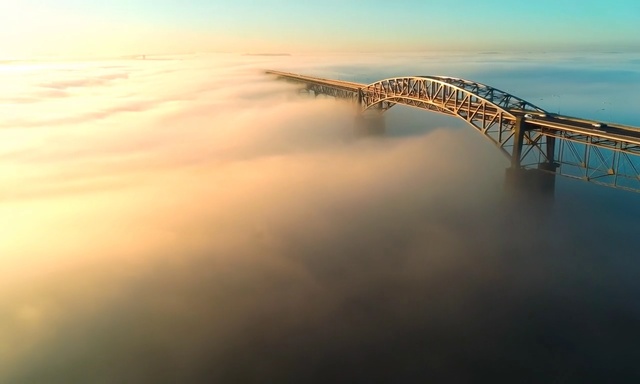}\hspace{\vsgap}\includegraphics[width=\vsframe]{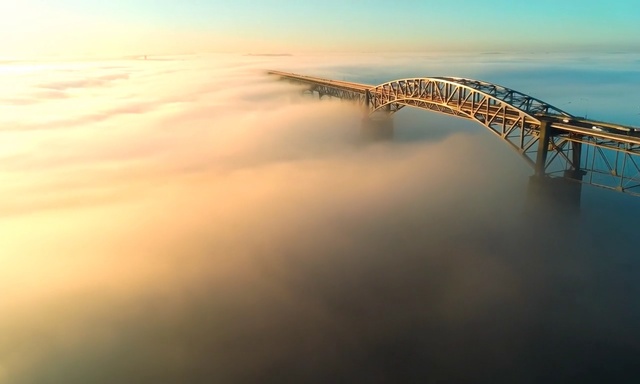}\hspace{\vsgap}\includegraphics[width=\vsframe]{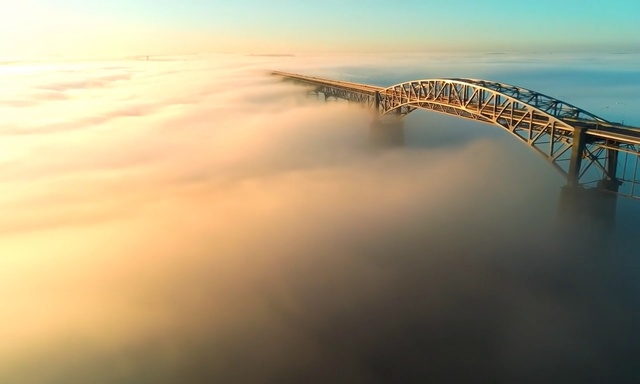}\hspace{\vsgap}\includegraphics[width=\vsframe]{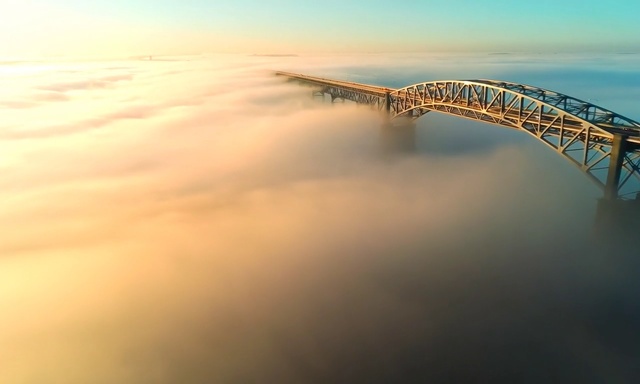}\hspace{\vsgap}\includegraphics[width=\vsframe]{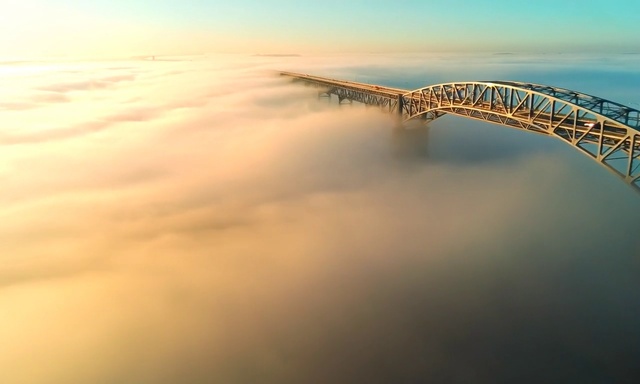}}\\
    [3pt]{\small\textbf{JetDiffusion (Ours)}}\\[1pt]
    \makebox[\linewidth][c]{\includegraphics[width=\vsframe]{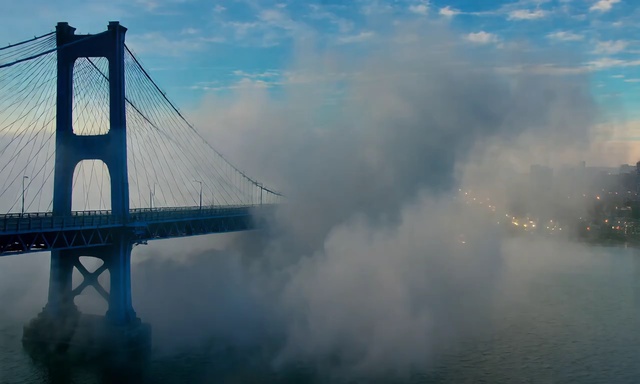}\hspace{\vsgap}\includegraphics[width=\vsframe]{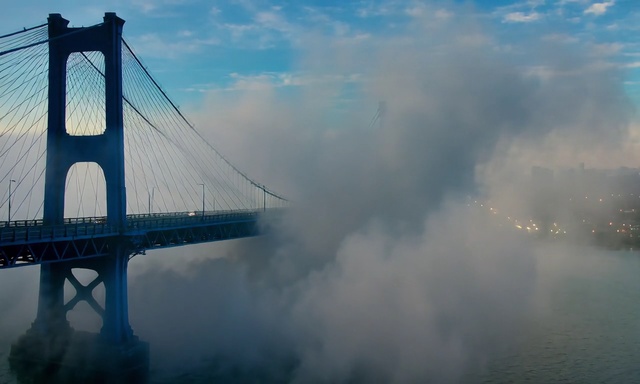}\hspace{\vsgap}\includegraphics[width=\vsframe]{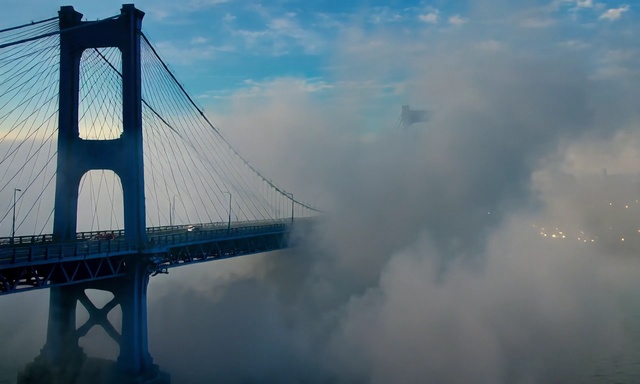}\hspace{\vsgap}\includegraphics[width=\vsframe]{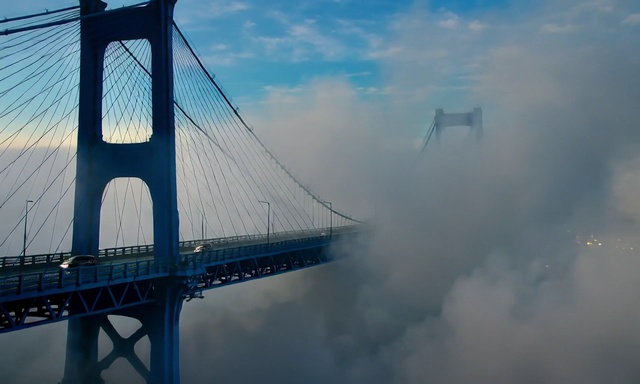}\hspace{\vsgap}\includegraphics[width=\vsframe]{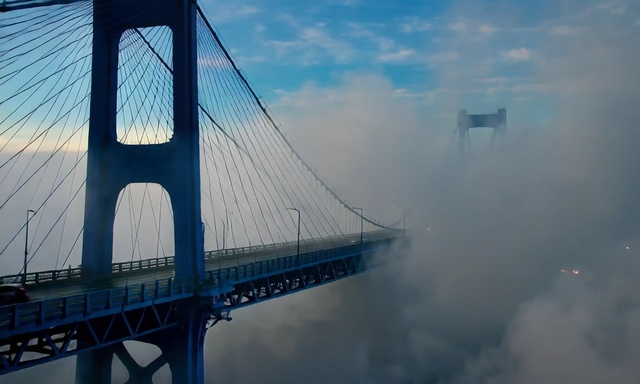}\hspace{\vsgap}\includegraphics[width=\vsframe]{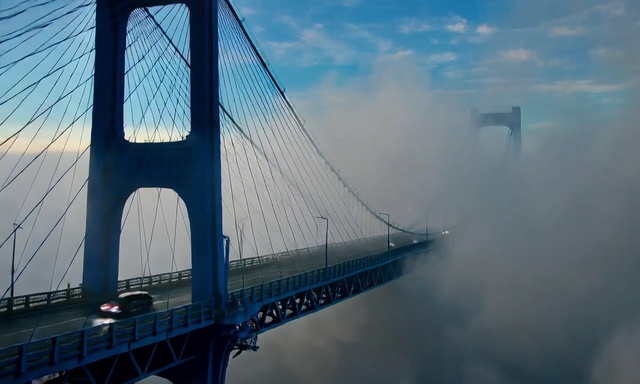}}\\
    \caption{Video generation with Wan2.1-T2V-14B. Top: the base model. Bottom: JetDiffusion.
    Text prompt: \textit{``A serene early morning scene with dense fog rolling over a long suspension bridge. The camera slowly pans along the length of the bridge, capturing the misty atmosphere as it gently moves across the structure. The bridge appears slightly ethereal, with only the occasional car light piercing through the fog. The rolling fog creates a dynamic and atmospheric effect, enhancing the sense of mystery and tranquility. Wide shot, slow horizontal pan.''}}
    \label{fig:video_wan14b_s07}
\end{figure}

\begin{figure}[!htb]
    \centering\lineskip=1pt
    {\small\textbf{Wan2.1-T2V-14B}}\\[1pt]
    \makebox[\linewidth][c]{\includegraphics[width=\vsframe]{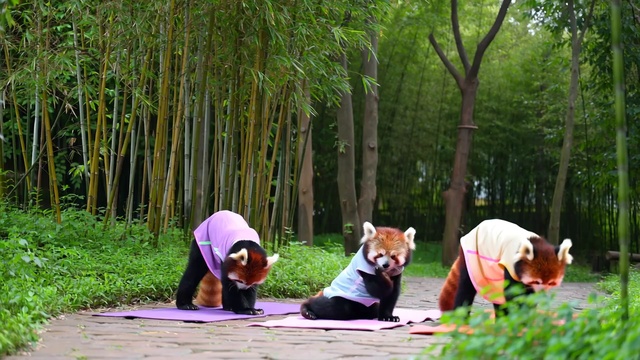}\hspace{\vsgap}\includegraphics[width=\vsframe]{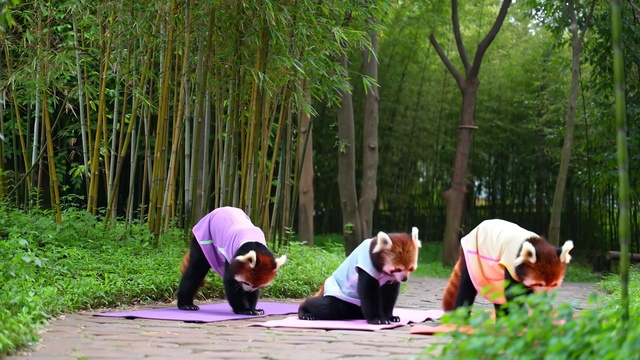}\hspace{\vsgap}\includegraphics[width=\vsframe]{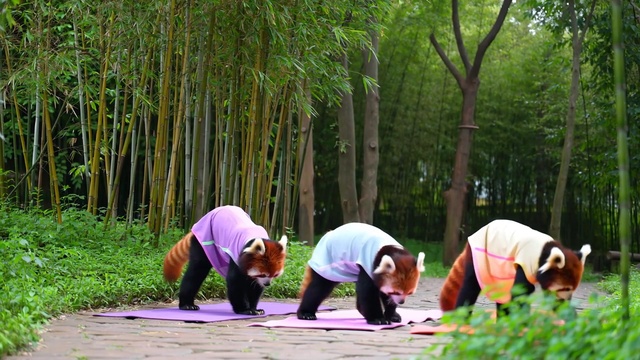}\hspace{\vsgap}\includegraphics[width=\vsframe]{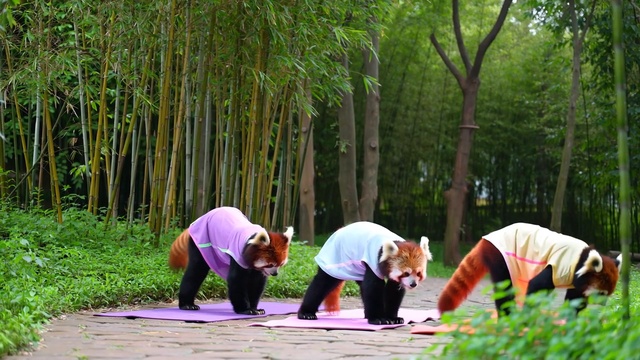}\hspace{\vsgap}\includegraphics[width=\vsframe]{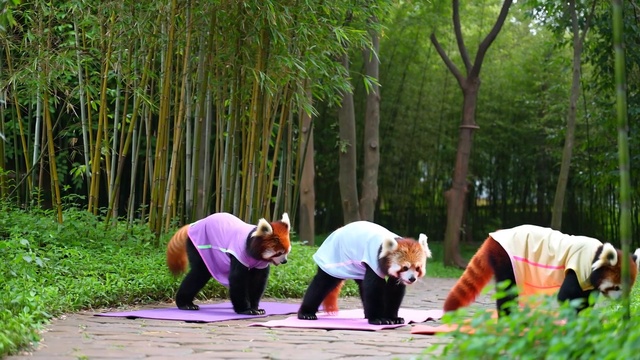}\hspace{\vsgap}\includegraphics[width=\vsframe]{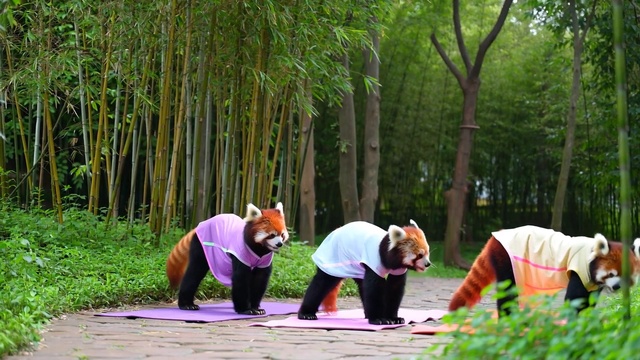}}\\
    [3pt]{\small\textbf{JetDiffusion (Ours)}}\\[1pt]
    \makebox[\linewidth][c]{\includegraphics[width=\vsframe]{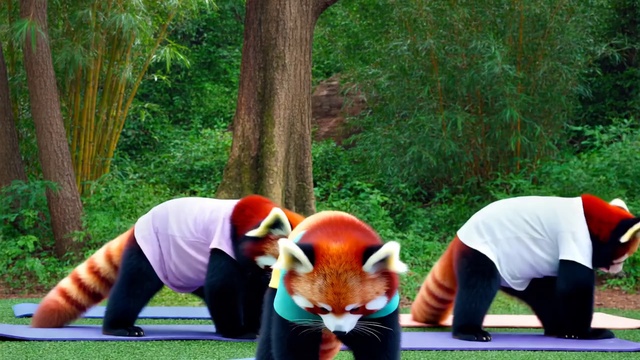}\hspace{\vsgap}\includegraphics[width=\vsframe]{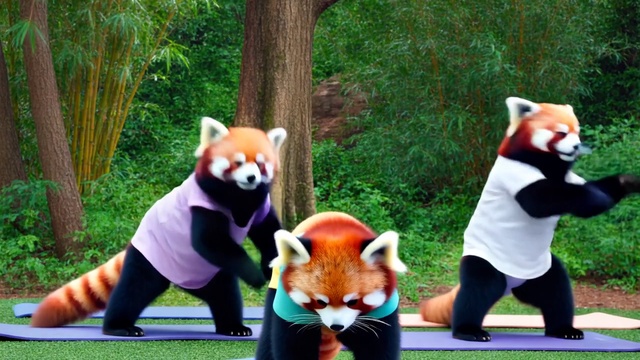}\hspace{\vsgap}\includegraphics[width=\vsframe]{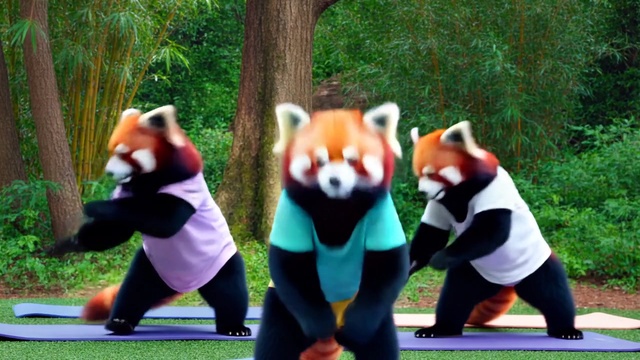}\hspace{\vsgap}\includegraphics[width=\vsframe]{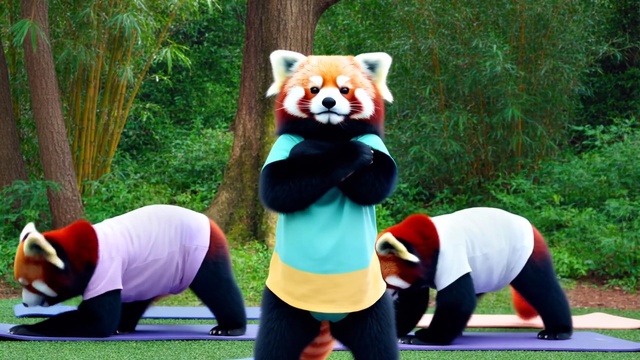}\hspace{\vsgap}\includegraphics[width=\vsframe]{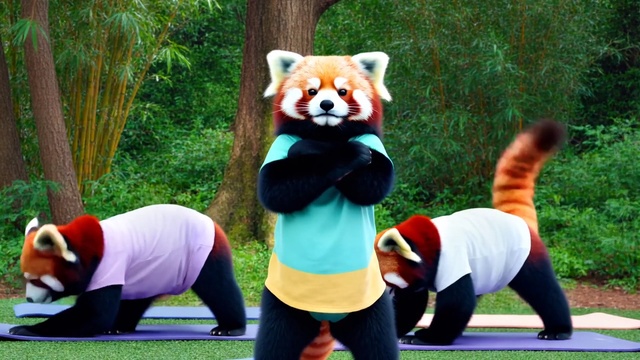}\hspace{\vsgap}\includegraphics[width=\vsframe]{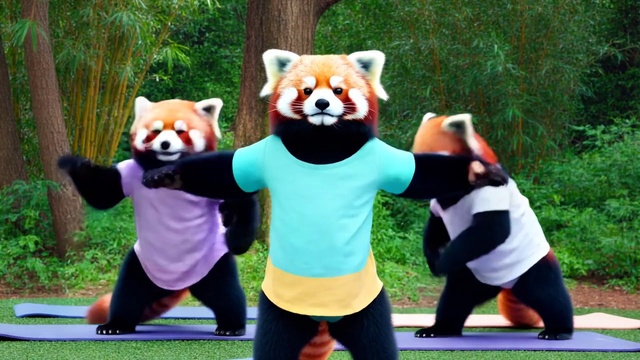}}\\
    \caption{Video generation with Wan2.1-T2V-14B. Top: the base model. Bottom: JetDiffusion.
    Text prompt: \textit{``Red pandas performing human-like yoga poses in a lush forest setting. The pandas are dressed in loose, colorful yoga outfits, each engaging in different poses such as downward dog, tree pose, and warrior pose. They are surrounded by tall bamboo trees and green foliage, creating a serene and natural environment. The camera captures medium shots of each panda, emphasizing their playful expressions and the fluidity of their movements. The scene is calm and tranquil, with soft lighting and a gentle background soundtrack.''}}
    \label{fig:video_wan14b_s08}
\end{figure}

\begin{figure}[!htb]
    \centering\lineskip=1pt
    {\small\textbf{Wan2.1-T2V-14B}}\\[1pt]
    \makebox[\linewidth][c]{\includegraphics[width=\vsframe]{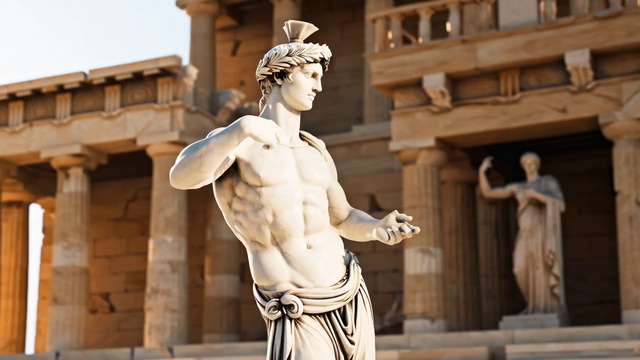}\hspace{\vsgap}\includegraphics[width=\vsframe]{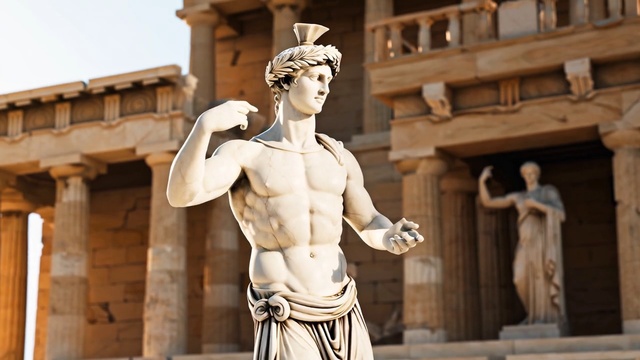}\hspace{\vsgap}\includegraphics[width=\vsframe]{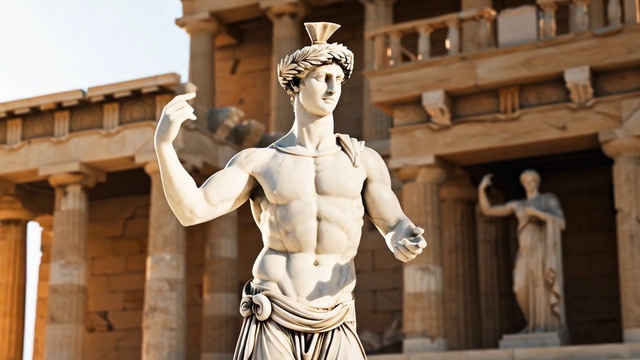}\hspace{\vsgap}\includegraphics[width=\vsframe]{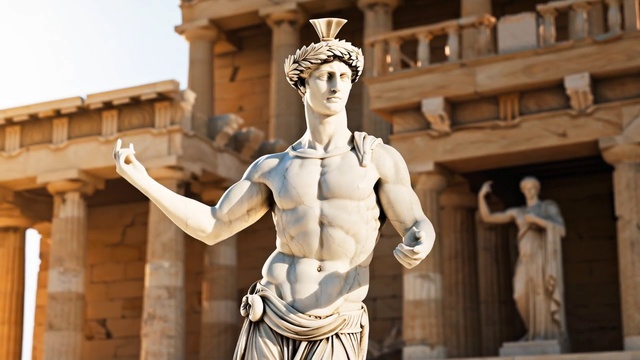}\hspace{\vsgap}\includegraphics[width=\vsframe]{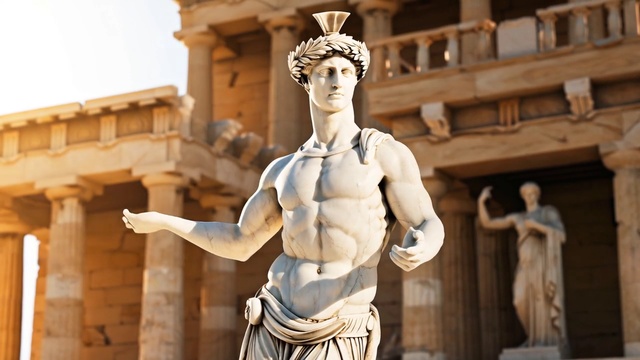}\hspace{\vsgap}\includegraphics[width=\vsframe]{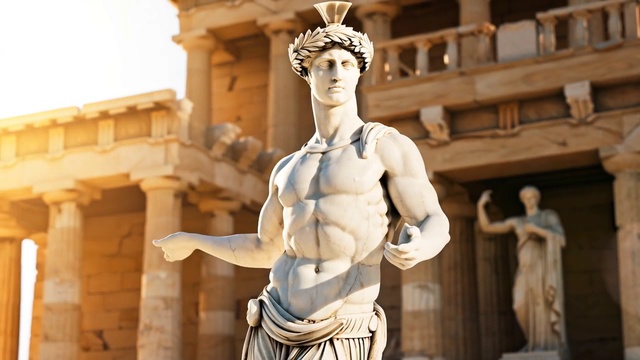}}\\
    [3pt]{\small\textbf{JetDiffusion (Ours)}}\\[1pt]
    \makebox[\linewidth][c]{\includegraphics[width=\vsframe]{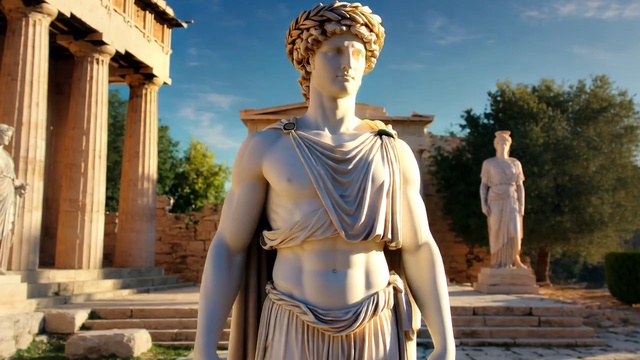}\hspace{\vsgap}\includegraphics[width=\vsframe]{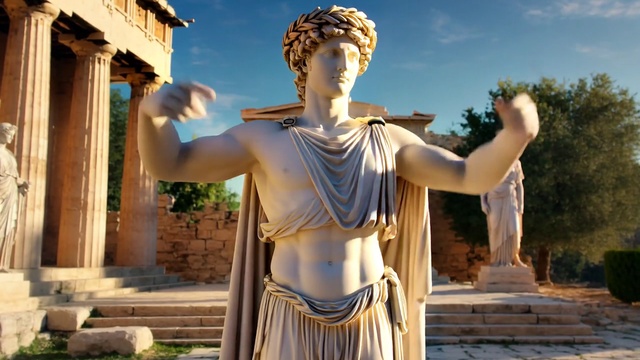}\hspace{\vsgap}\includegraphics[width=\vsframe]{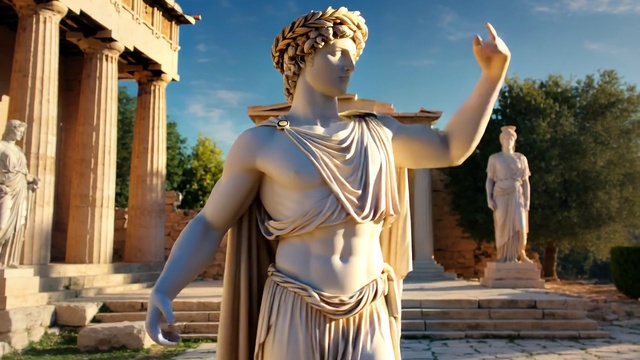}\hspace{\vsgap}\includegraphics[width=\vsframe]{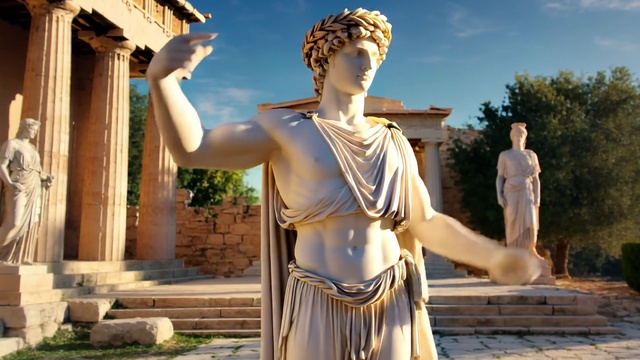}\hspace{\vsgap}\includegraphics[width=\vsframe]{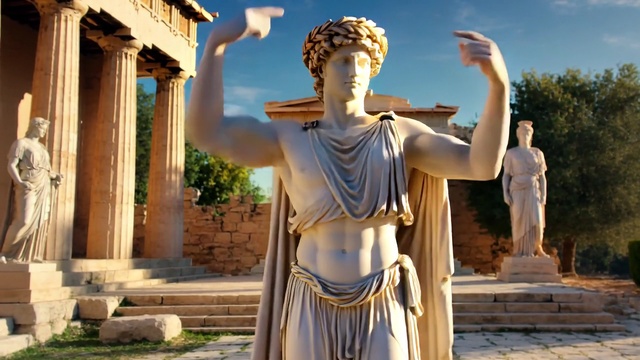}\hspace{\vsgap}\includegraphics[width=\vsframe]{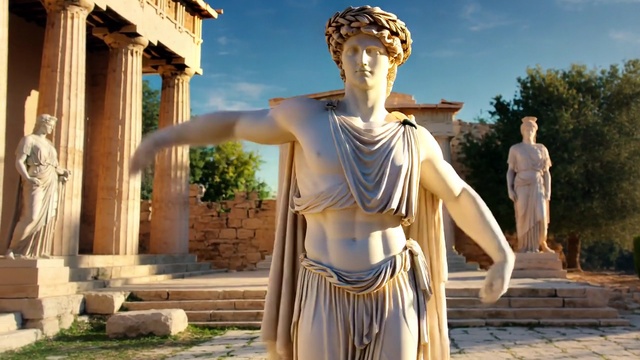}}\\
    \caption{Video generation with Wan2.1-T2V-14B. Top: the base model. Bottom: JetDiffusion.
    Text prompt: \textit{``A strong Greek statue coming to life and dancing gracefully. The statue, made of marble, has defined muscles and a chiseled physique. It wears a traditional Greek warrior's robe with intricate designs and a laurel wreath on its head. The statue moves with fluidity, performing classic dance poses and gestures. The background is an ancient Greek temple courtyard with columns and statues, bathed in warm sunlight. The scene is captured in a medium shot, emphasizing the statue's movement and expressions.''}}
    \label{fig:video_wan14b_s09}
\end{figure}

\begin{figure}[!htb]
    \centering\lineskip=1pt
    {\small\textbf{Wan2.1-T2V-14B}}\\[1pt]
    \makebox[\linewidth][c]{\includegraphics[width=\vsframe]{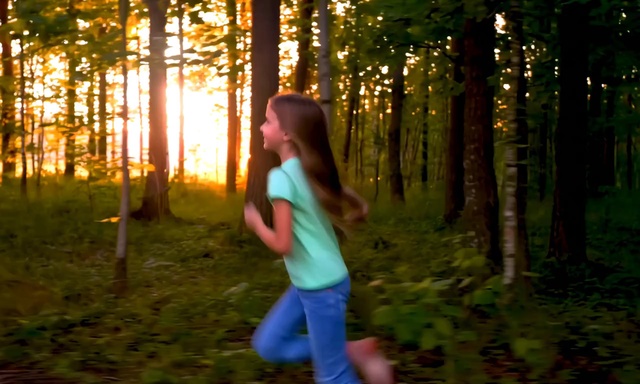}\hspace{\vsgap}\includegraphics[width=\vsframe]{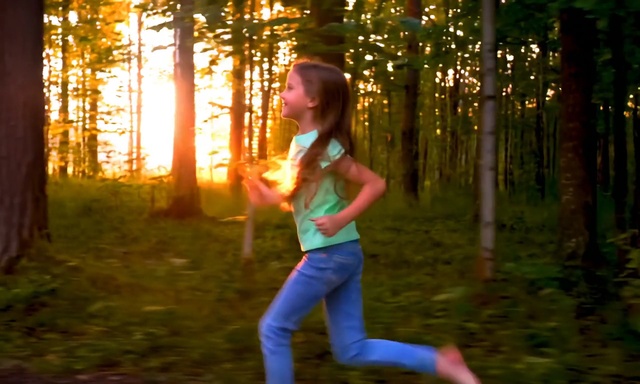}\hspace{\vsgap}\includegraphics[width=\vsframe]{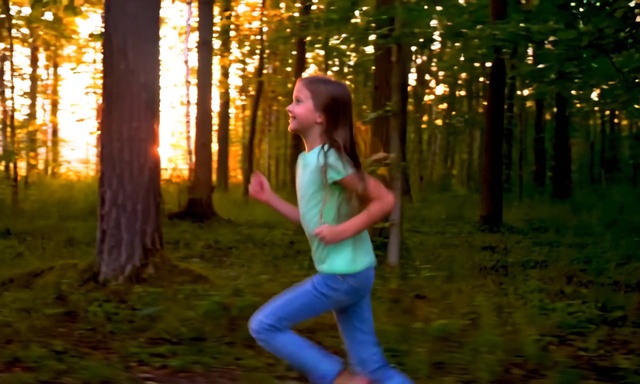}\hspace{\vsgap}\includegraphics[width=\vsframe]{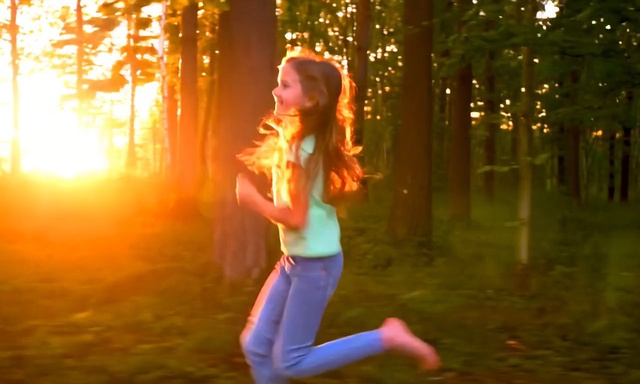}\hspace{\vsgap}\includegraphics[width=\vsframe]{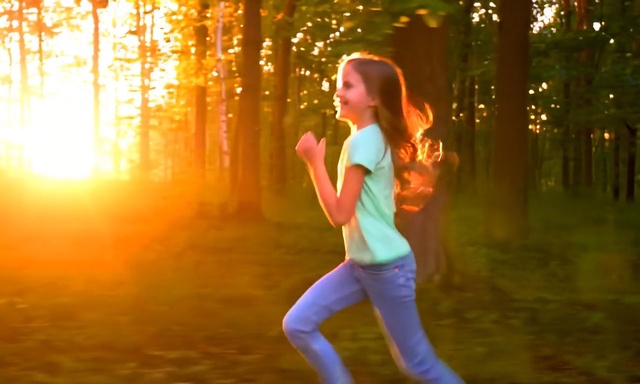}\hspace{\vsgap}\includegraphics[width=\vsframe]{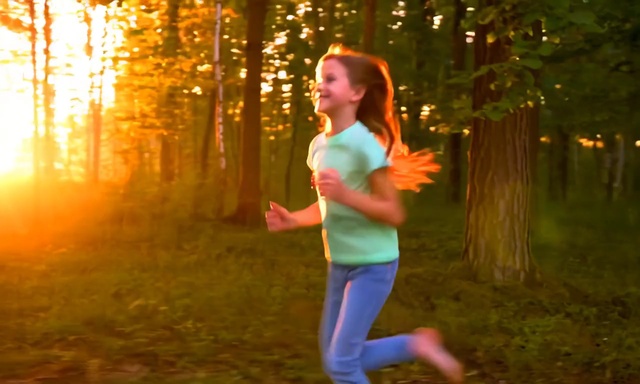}}\\
    [3pt]{\small\textbf{JetDiffusion (Ours)}}\\[1pt]
    \makebox[\linewidth][c]{\includegraphics[width=\vsframe]{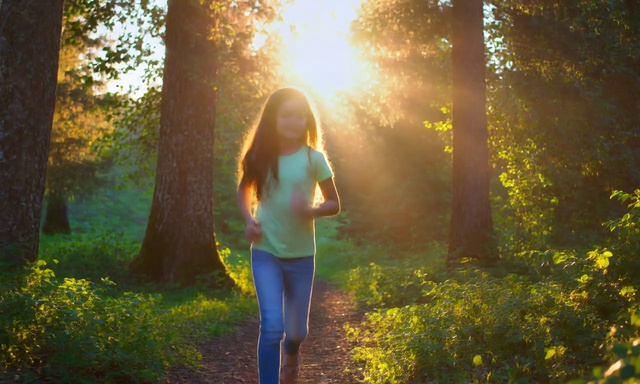}\hspace{\vsgap}\includegraphics[width=\vsframe]{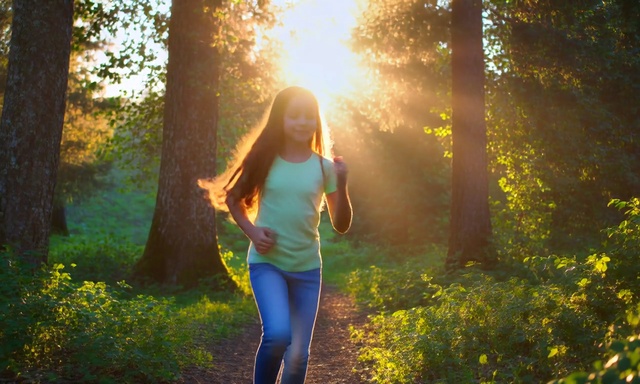}\hspace{\vsgap}\includegraphics[width=\vsframe]{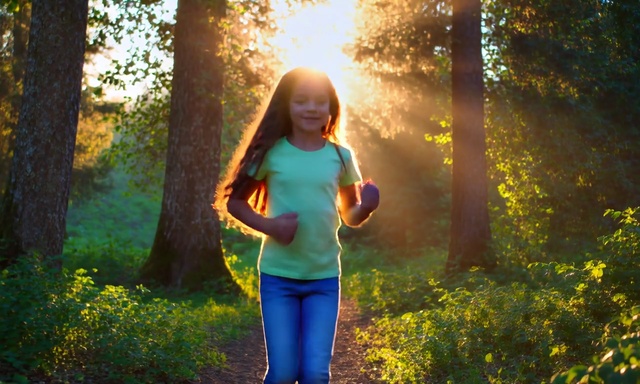}\hspace{\vsgap}\includegraphics[width=\vsframe]{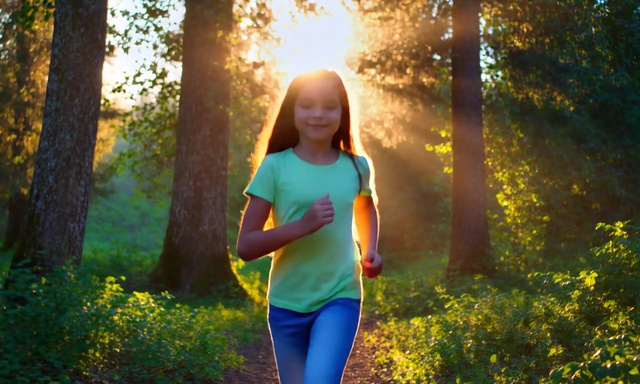}\hspace{\vsgap}\includegraphics[width=\vsframe]{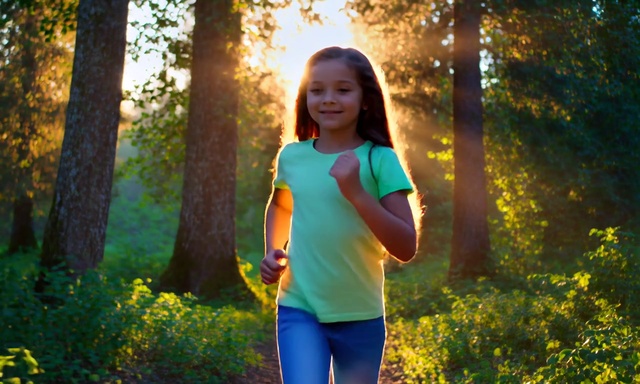}\hspace{\vsgap}\includegraphics[width=\vsframe]{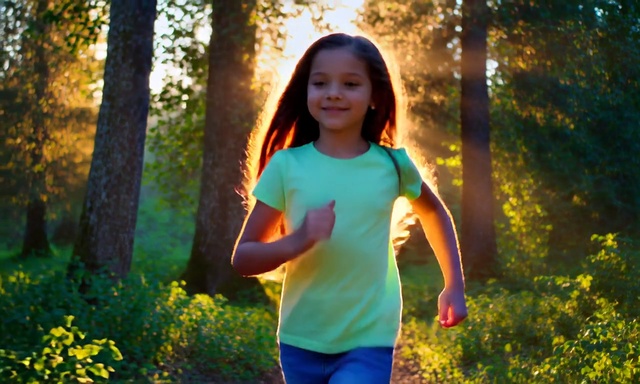}}\\
    \caption{Video generation with Wan2.1-T2V-14B. Top: the base model. Bottom: JetDiffusion.
    Text prompt: \textit{``A young girl with long brown hair running through a dense forest early in the morning. She is wearing a light green t-shirt and blue jeans, with bare feet, showcasing her adventurous spirit. The rising sun illuminates the scene, casting golden rays through the canopy of trees, creating a warm and vibrant atmosphere. The background shows tall trees with dappled sunlight filtering through the leaves. The girl's face is filled with joy and determination as she runs freely. Medium shot focusing on the girl from a side angle, capturing her motion and the surrounding nature.''}}
    \label{fig:video_wan14b_s10}
\end{figure}

\begin{figure}[!htb]
    \centering\lineskip=1pt
    {\small\textbf{Wan2.1-T2V-14B}}\\[1pt]
    \makebox[\linewidth][c]{\includegraphics[width=\vsframe]{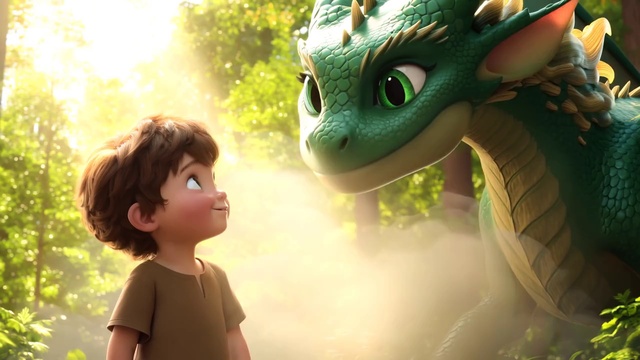}\hspace{\vsgap}\includegraphics[width=\vsframe]{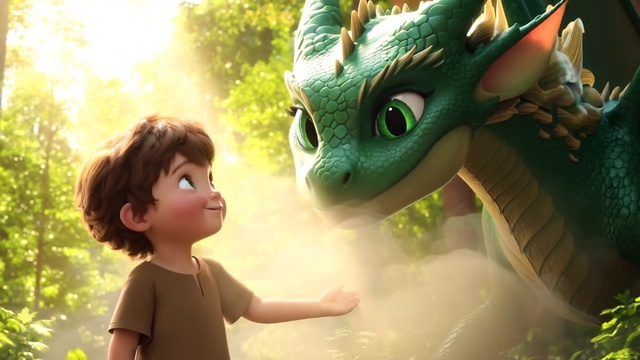}\hspace{\vsgap}\includegraphics[width=\vsframe]{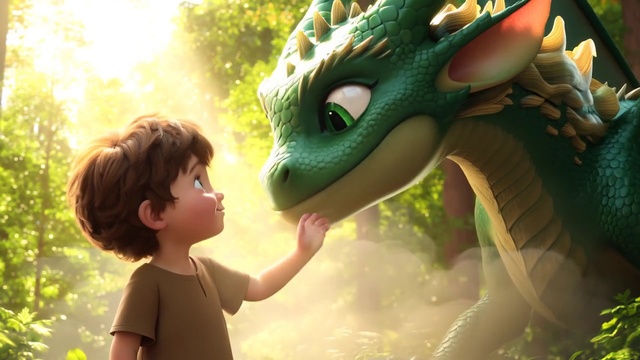}\hspace{\vsgap}\includegraphics[width=\vsframe]{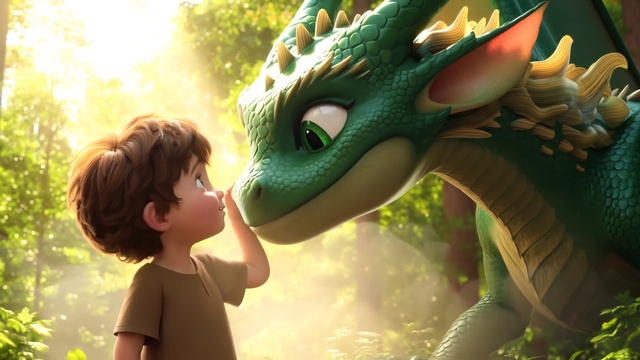}\hspace{\vsgap}\includegraphics[width=\vsframe]{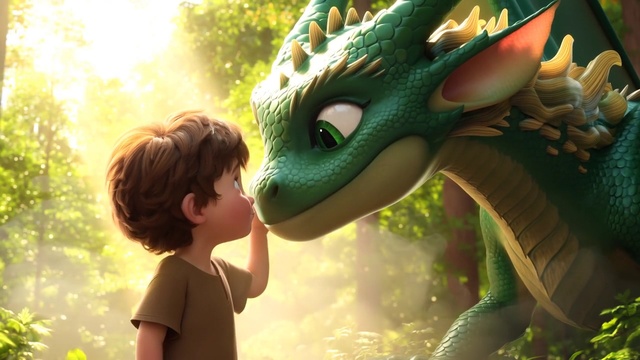}\hspace{\vsgap}\includegraphics[width=\vsframe]{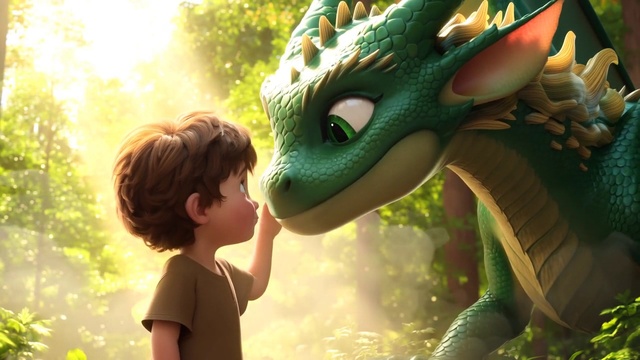}}\\
    [3pt]{\small\textbf{JetDiffusion (Ours)}}\\[1pt]
    \makebox[\linewidth][c]{\includegraphics[width=\vsframe]{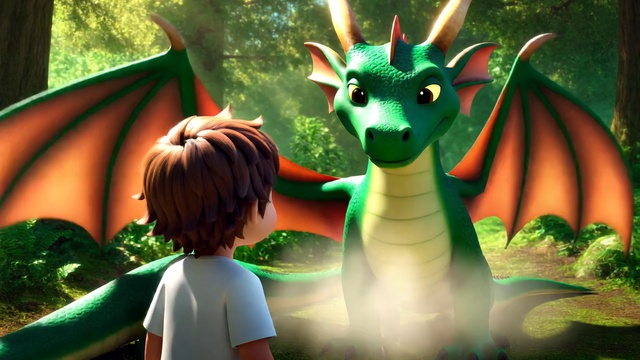}\hspace{\vsgap}\includegraphics[width=\vsframe]{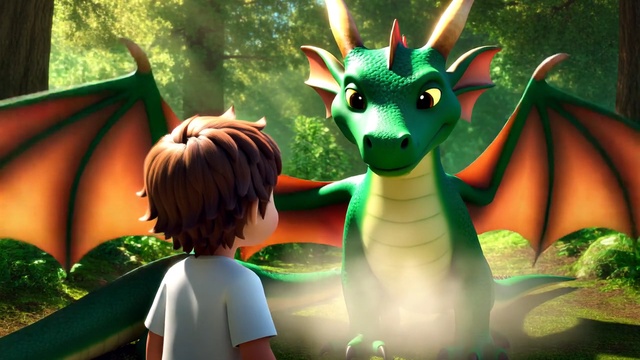}\hspace{\vsgap}\includegraphics[width=\vsframe]{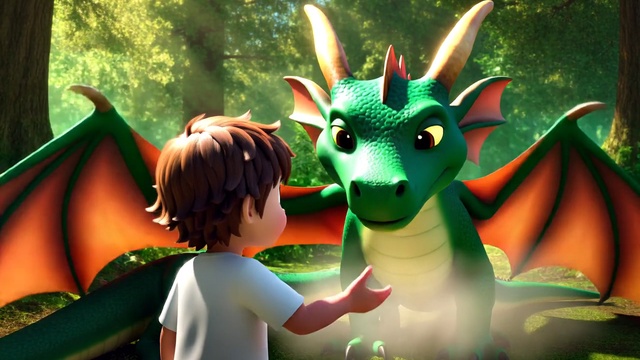}\hspace{\vsgap}\includegraphics[width=\vsframe]{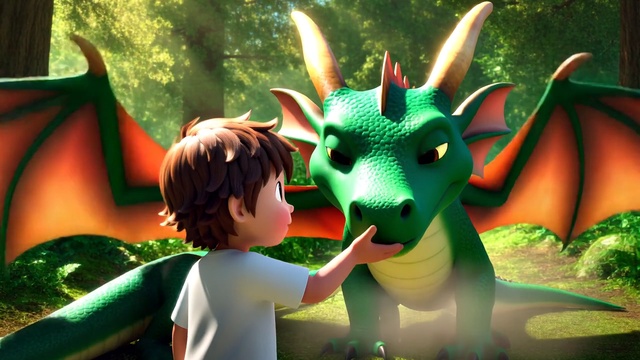}\hspace{\vsgap}\includegraphics[width=\vsframe]{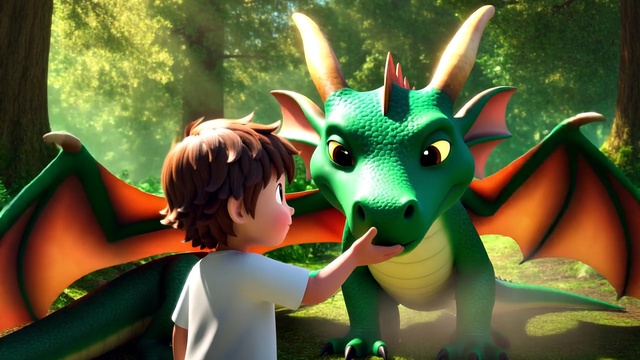}\hspace{\vsgap}\includegraphics[width=\vsframe]{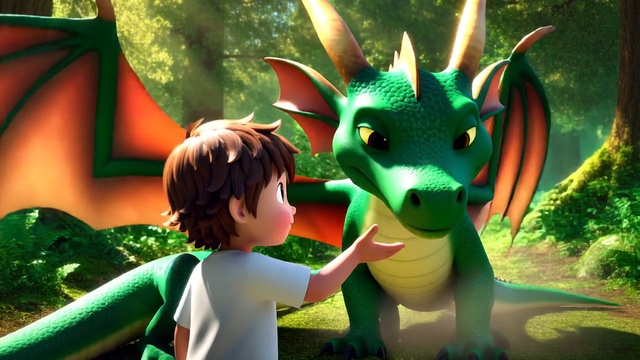}}\\
    \caption{Video generation with Wan2.1-T2V-14B. Top: the base model. Bottom: JetDiffusion.
    Text prompt: \textit{``A heartwarming scene from a fantasy book, featuring a young human protagonist meeting a friendly dragon in a lush forest clearing. The dragon is depicted as having emerald green scales, large expressive eyes, and a gentle demeanor. It breathes a soft, warm mist that illuminates the surrounding area. The human, a curious child with messy brown hair and wide-eyed wonder, reaches out tentatively towards the dragon. The dragon gently nudges the child with its snout, showing affection. The background showcases a vibrant forest with dappled sunlight filtering through the leaves. The scene captures a moment of mutual trust and friendship between species. Mid-shot, capturing both the child and the dragon in full.''}}
    \label{fig:video_wan14b_s11}
\end{figure}

\begin{figure}[!htb]
    \centering\lineskip=1pt
    {\small\textbf{Wan2.1-T2V-14B}}\\[1pt]
    \makebox[\linewidth][c]{\includegraphics[width=\vsframe]{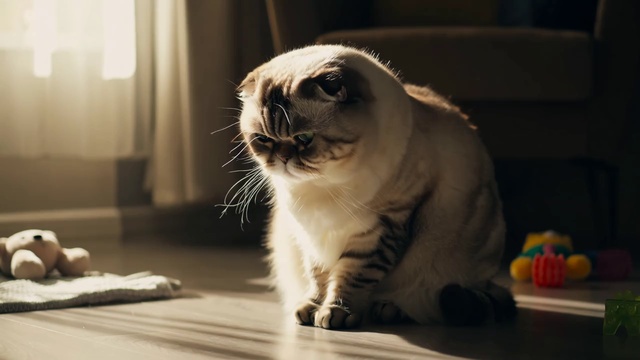}\hspace{\vsgap}\includegraphics[width=\vsframe]{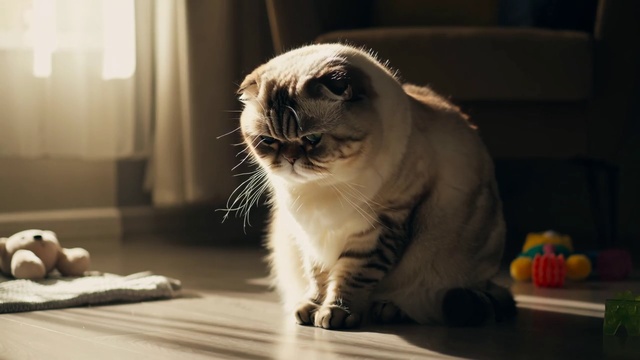}\hspace{\vsgap}\includegraphics[width=\vsframe]{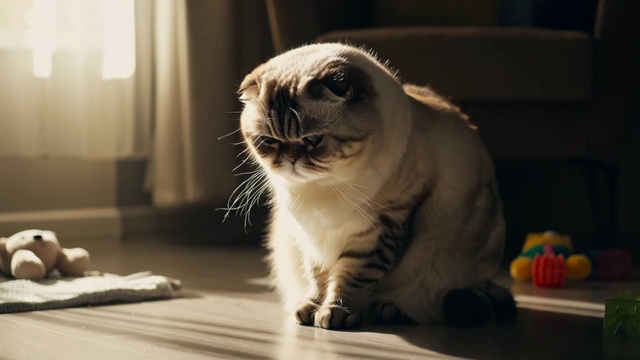}\hspace{\vsgap}\includegraphics[width=\vsframe]{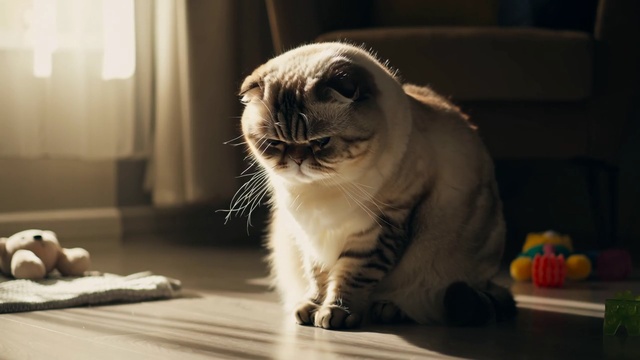}\hspace{\vsgap}\includegraphics[width=\vsframe]{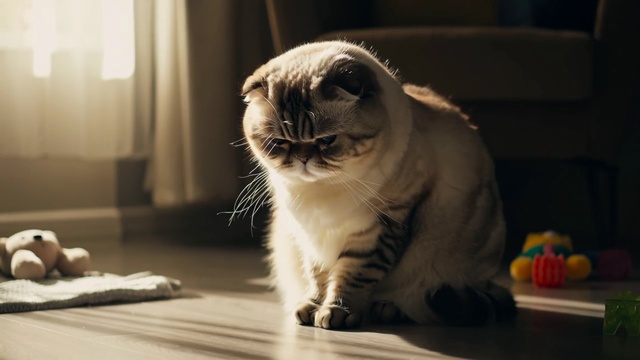}\hspace{\vsgap}\includegraphics[width=\vsframe]{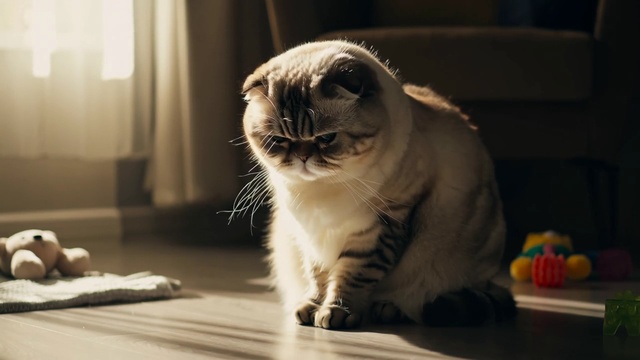}}\\
    [3pt]{\small\textbf{JetDiffusion (Ours)}}\\[1pt]
    \makebox[\linewidth][c]{\includegraphics[width=\vsframe]{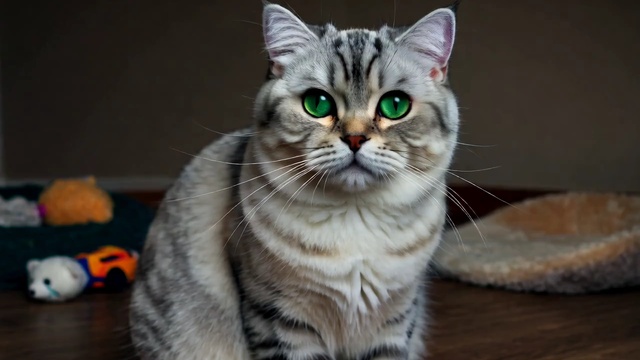}\hspace{\vsgap}\includegraphics[width=\vsframe]{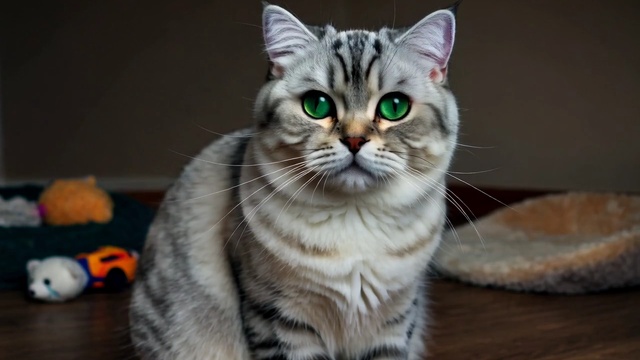}\hspace{\vsgap}\includegraphics[width=\vsframe]{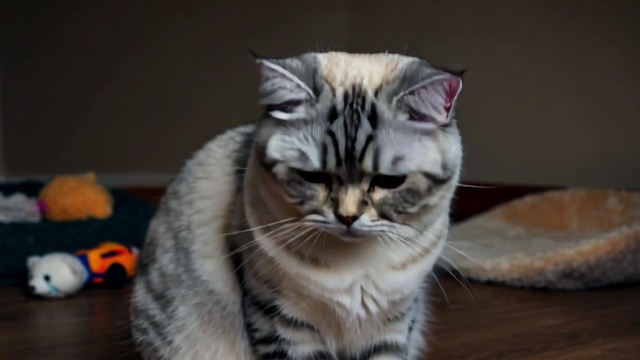}\hspace{\vsgap}\includegraphics[width=\vsframe]{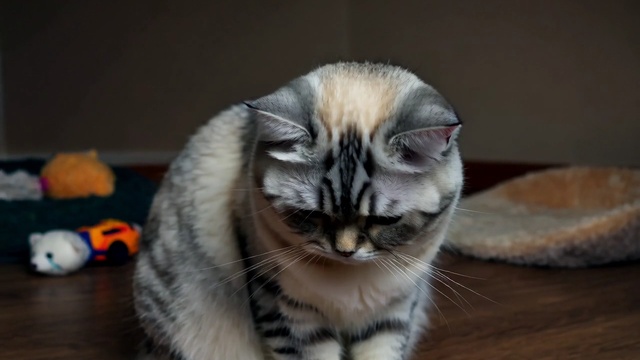}\hspace{\vsgap}\includegraphics[width=\vsframe]{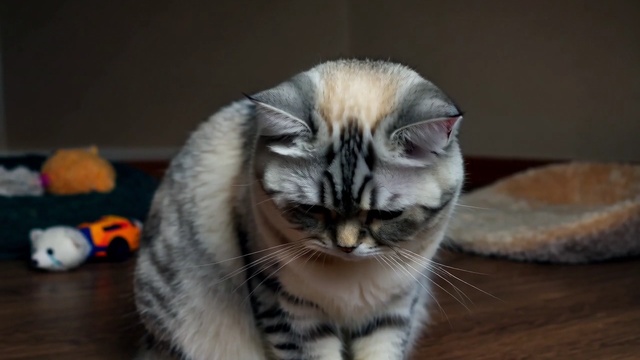}\hspace{\vsgap}\includegraphics[width=\vsframe]{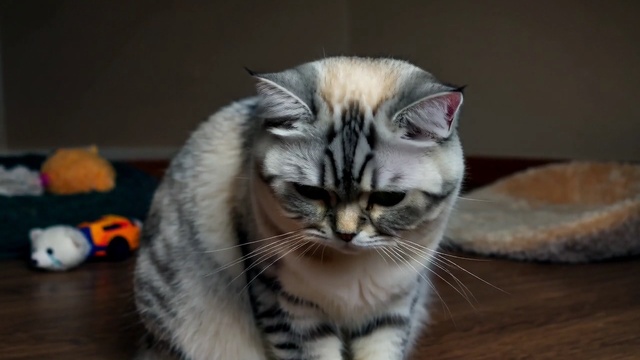}}\\
    \caption{Video generation with Wan2.1-T2V-14B. Top: the base model. Bottom: JetDiffusion.
    Text prompt: \textit{``A realistic video shows a solitary domestic cat sitting in a dimly lit room, its fur sleek and soft. The cat has large, expressive eyes filled with sadness and tears streaming down its face. It is hunched over, with its ears flattened against its head, conveying deep sorrow. The lighting is soft and melancholic, casting gentle shadows across the room. The background includes a few scattered toys and a small blanket, adding to the sense of loneliness. The cat's body language and facial expressions emphasize its emotional state. Static shot, medium close-up.''}}
    \label{fig:video_wan14b_s12}
\end{figure}

\begin{figure}[!htb]
    \centering\lineskip=1pt
    {\small\textbf{Wan2.1-T2V-14B}}\\[1pt]
    \makebox[\linewidth][c]{\includegraphics[width=\vsframe]{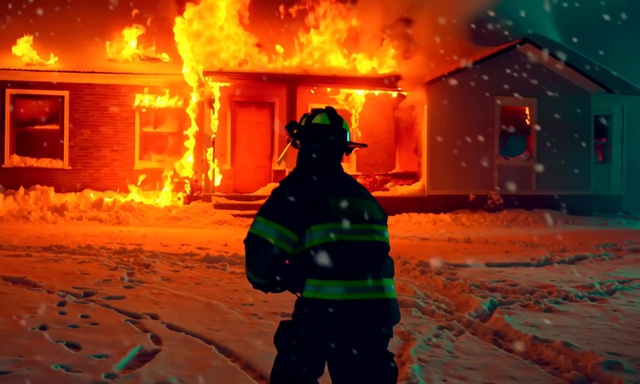}\hspace{\vsgap}\includegraphics[width=\vsframe]{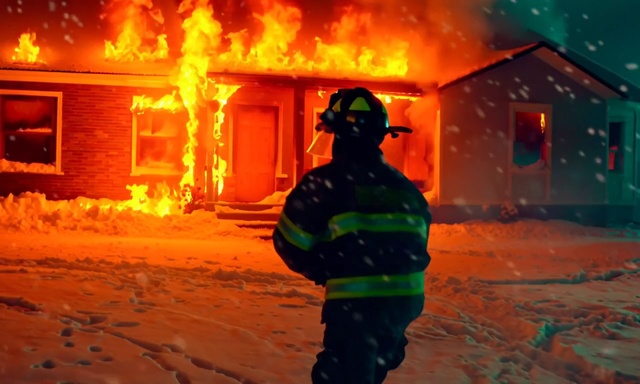}\hspace{\vsgap}\includegraphics[width=\vsframe]{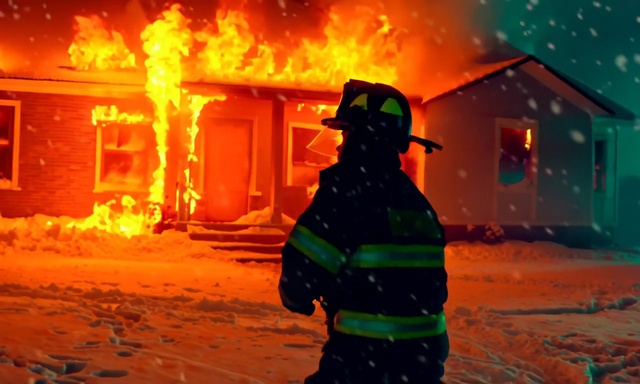}\hspace{\vsgap}\includegraphics[width=\vsframe]{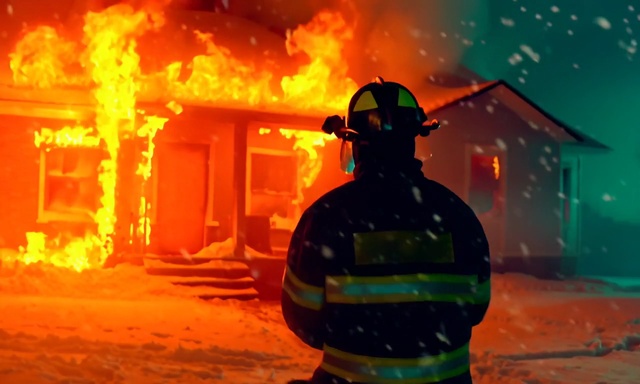}\hspace{\vsgap}\includegraphics[width=\vsframe]{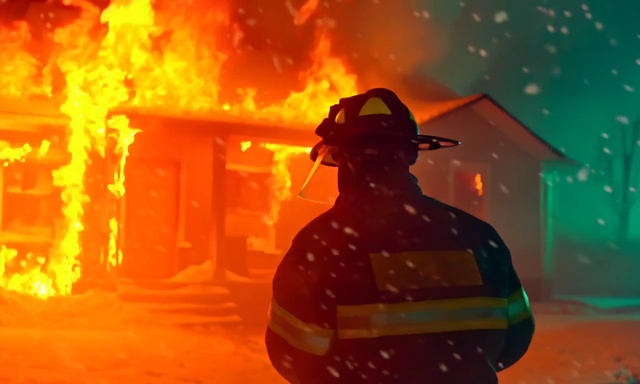}\hspace{\vsgap}\includegraphics[width=\vsframe]{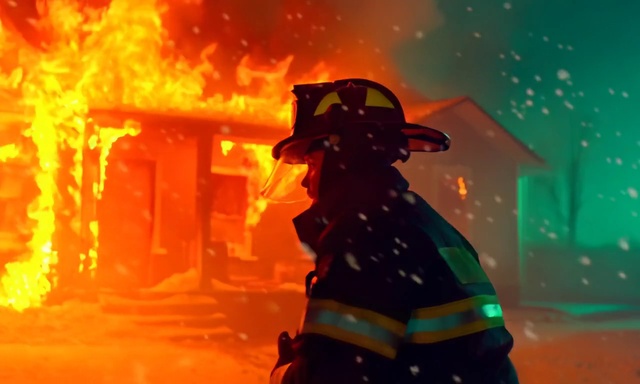}}\\
    [3pt]{\small\textbf{JetDiffusion (Ours)}}\\[1pt]
    \makebox[\linewidth][c]{\includegraphics[width=\vsframe]{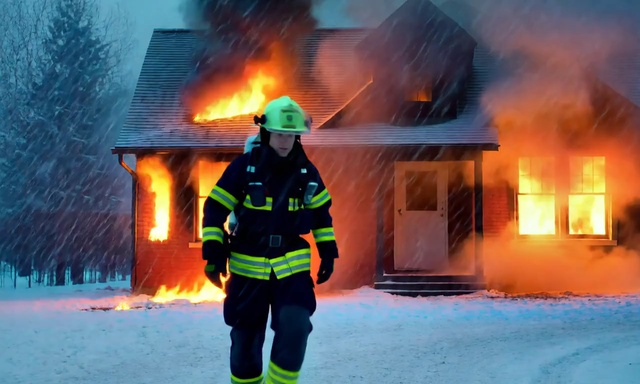}\hspace{\vsgap}\includegraphics[width=\vsframe]{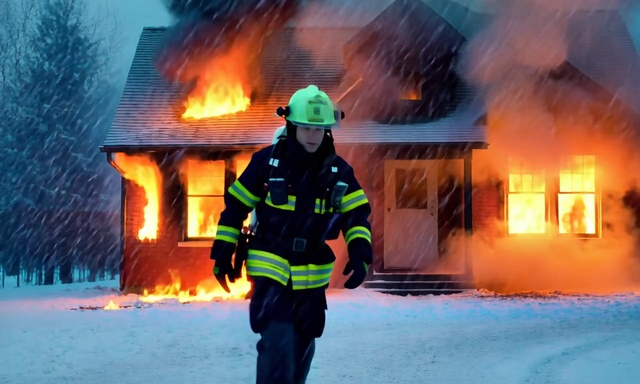}\hspace{\vsgap}\includegraphics[width=\vsframe]{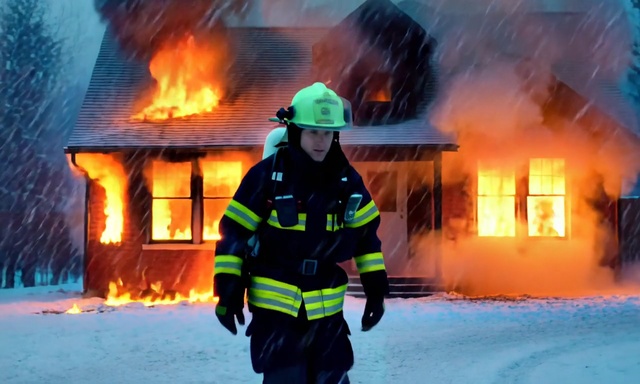}\hspace{\vsgap}\includegraphics[width=\vsframe]{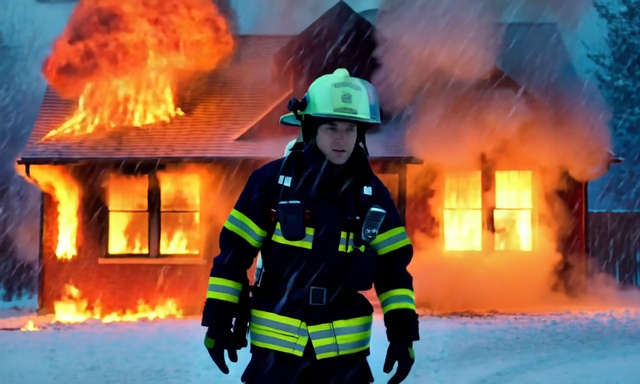}\hspace{\vsgap}\includegraphics[width=\vsframe]{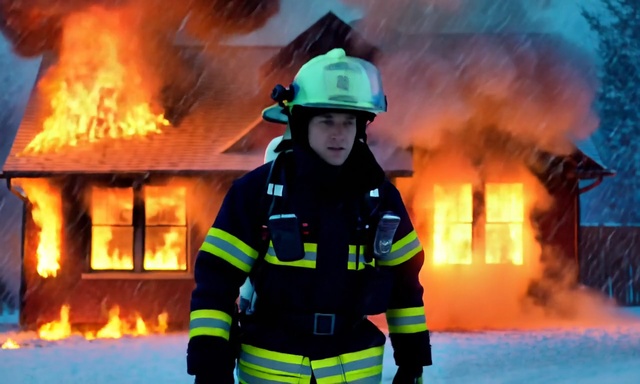}\hspace{\vsgap}\includegraphics[width=\vsframe]{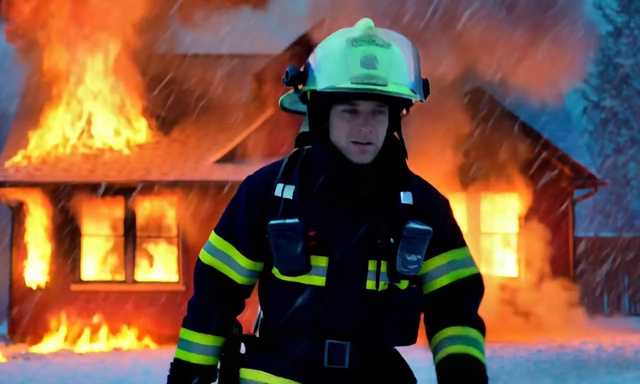}}\\
    \caption{Video generation with Wan2.1-T2V-14B. Top: the base model. Bottom: JetDiffusion.
    Text prompt: \textit{``A firefighter in full gear moves dynamically towards a burning house, his helmet reflecting the intense flames. Snowflakes gently fall around him, creating a stark contrast between the cold winter environment and the raging fire. The house is engulfed in bright orange and red flames, with smoke billowing out from various windows and the roof. The scene captures the tension and urgency of the moment, with the camera focused closely on the firefighter as he advances, and the flames roaring behind him. Close-up view with a slight pan to follow the firefighter's movement.''}}
    \label{fig:video_wan14b_s13}
\end{figure}

\begin{figure}[!htb]
    \centering\lineskip=1pt
    {\small\textbf{Wan2.1-T2V-14B}}\\[1pt]
    \makebox[\linewidth][c]{\includegraphics[width=\vsframe]{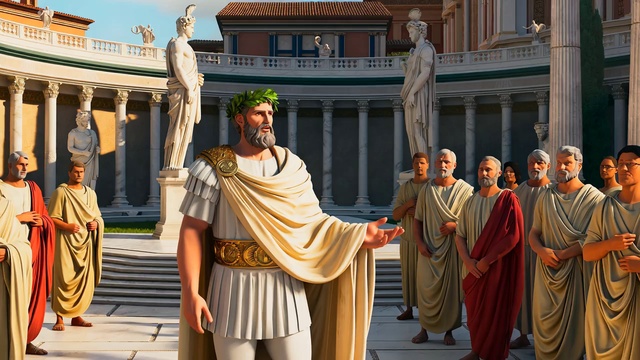}\hspace{\vsgap}\includegraphics[width=\vsframe]{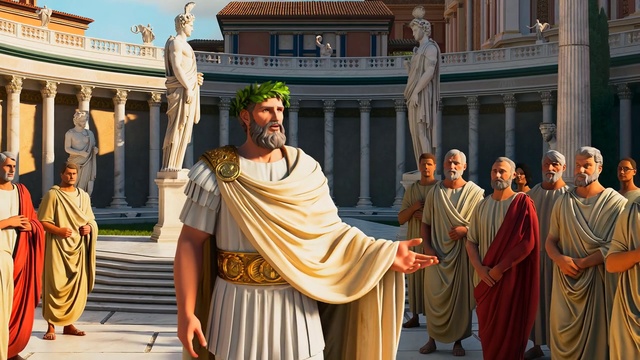}\hspace{\vsgap}\includegraphics[width=\vsframe]{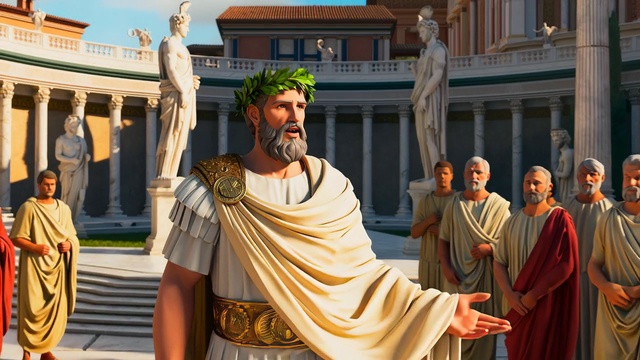}\hspace{\vsgap}\includegraphics[width=\vsframe]{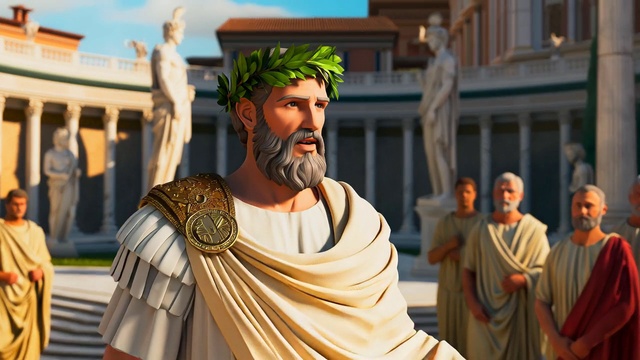}\hspace{\vsgap}\includegraphics[width=\vsframe]{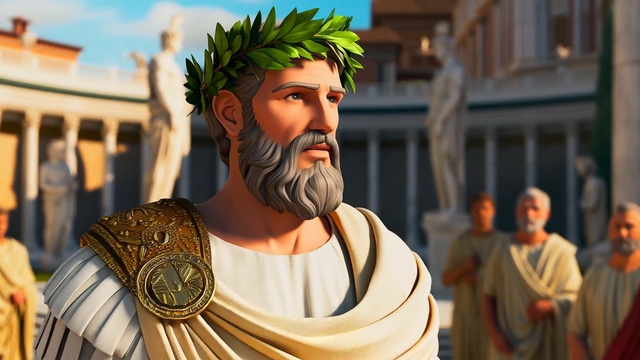}\hspace{\vsgap}\includegraphics[width=\vsframe]{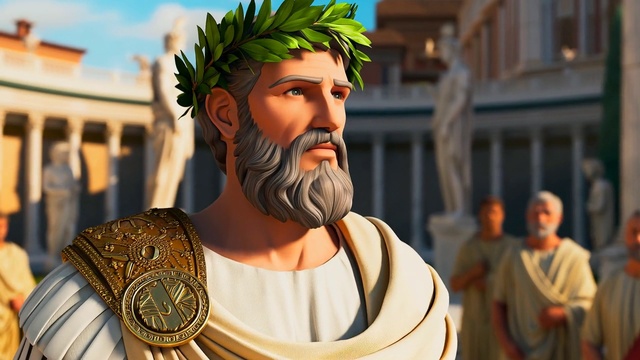}}\\
    [3pt]{\small\textbf{JetDiffusion (Ours)}}\\[1pt]
    \makebox[\linewidth][c]{\includegraphics[width=\vsframe]{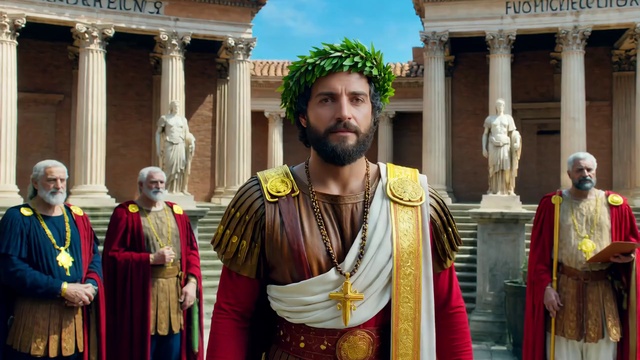}\hspace{\vsgap}\includegraphics[width=\vsframe]{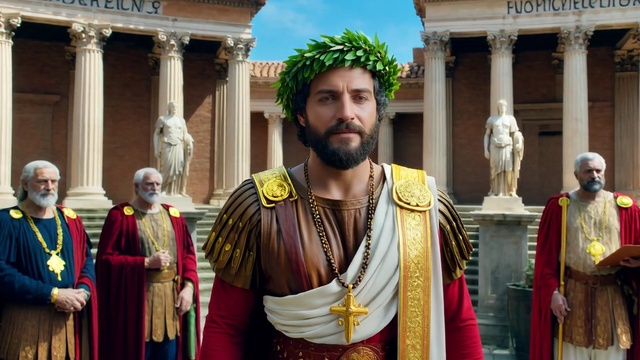}\hspace{\vsgap}\includegraphics[width=\vsframe]{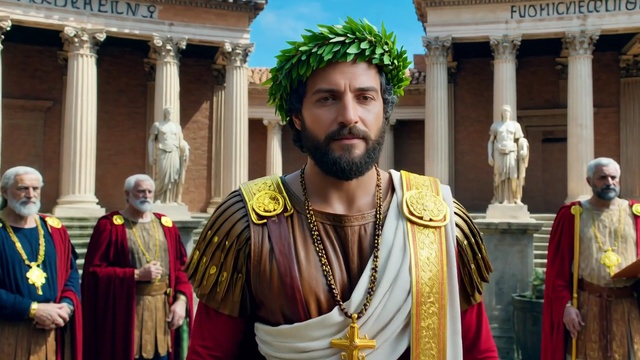}\hspace{\vsgap}\includegraphics[width=\vsframe]{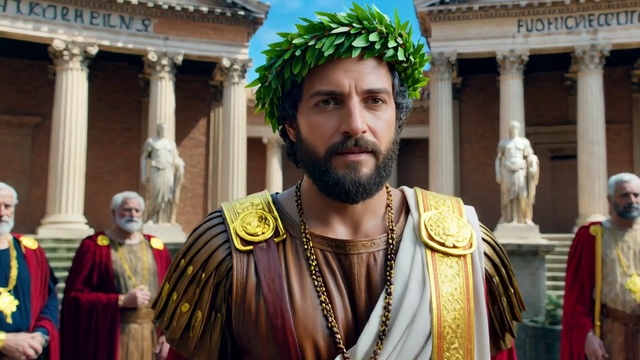}\hspace{\vsgap}\includegraphics[width=\vsframe]{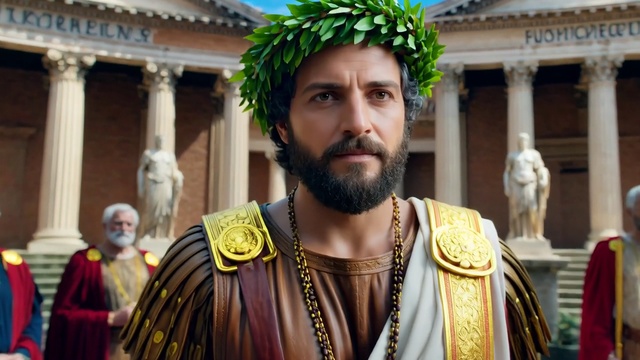}\hspace{\vsgap}\includegraphics[width=\vsframe]{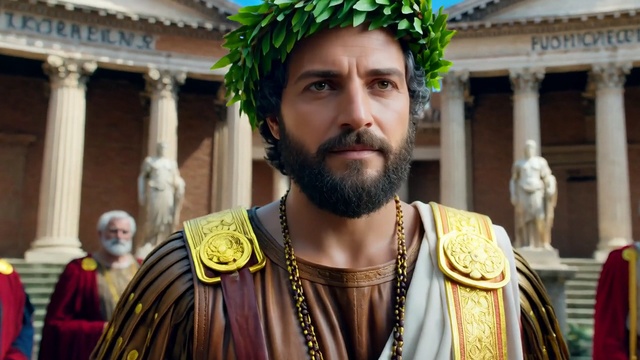}}\\
    \caption{Video generation with Wan2.1-T2V-14B. Top: the base model. Bottom: JetDiffusion.
    Text prompt: \textit{``A Roman Emperor, depicted as a stoic philosopher, is speaking to a group of people in a grand, ancient Roman courtyard. He wears a toga and laurel wreath, exuding wisdom and authority. His facial expression is calm and composed, with a serious yet contemplative gaze. The scene includes several well-dressed senators and scholars, listening intently. The background showcases columns, statues, and other architectural elements typical of Roman architecture. As he speaks, the camera slowly zooms in on the Emperor, highlighting his powerful presence and the depth of his thoughts. Medium shot transitioning to close-up.''}}
    \label{fig:video_wan14b_s14}
\end{figure}

\begin{figure}[!htb]
    \centering\lineskip=1pt
    {\small\textbf{Wan2.1-T2V-14B}}\\[1pt]
    \makebox[\linewidth][c]{\includegraphics[width=\vsframe]{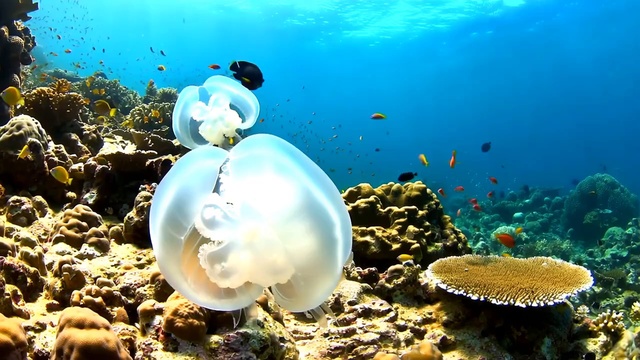}\hspace{\vsgap}\includegraphics[width=\vsframe]{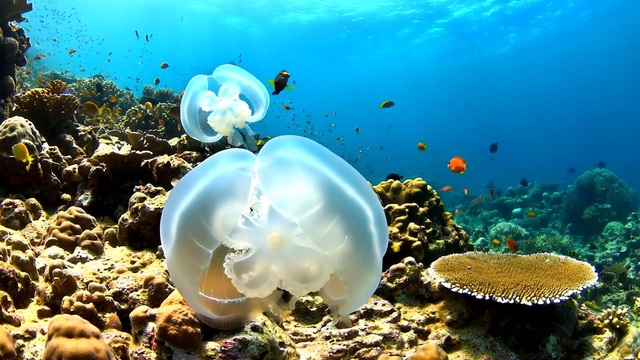}\hspace{\vsgap}\includegraphics[width=\vsframe]{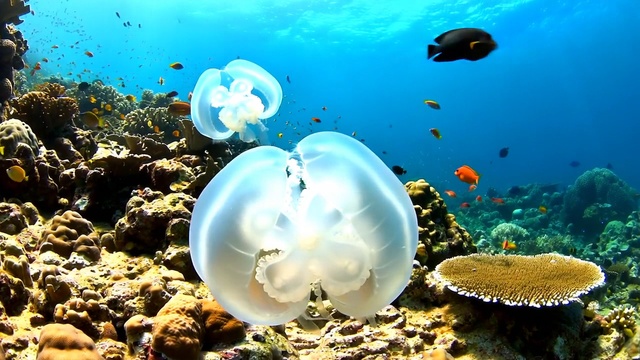}\hspace{\vsgap}\includegraphics[width=\vsframe]{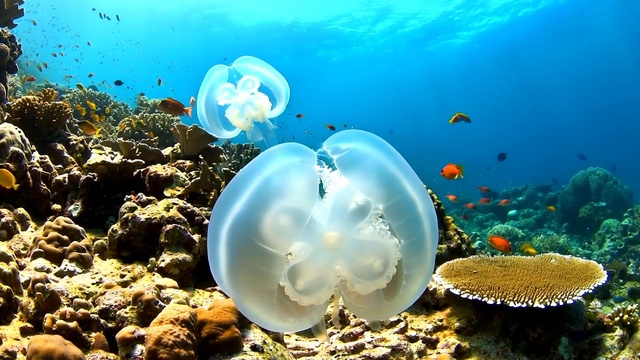}\hspace{\vsgap}\includegraphics[width=\vsframe]{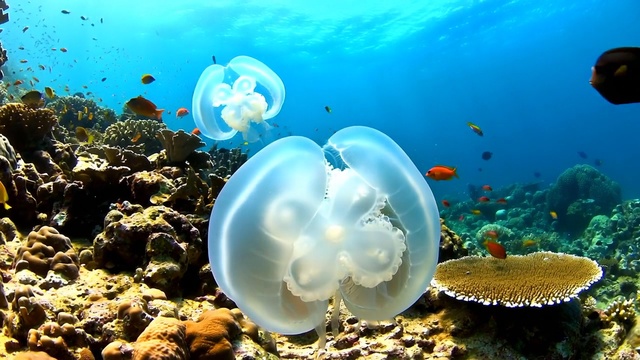}\hspace{\vsgap}\includegraphics[width=\vsframe]{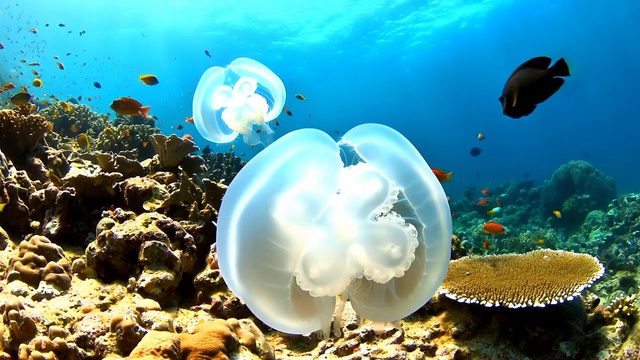}}\\
    [3pt]{\small\textbf{JetDiffusion (Ours)}}\\[1pt]
    \makebox[\linewidth][c]{\includegraphics[width=\vsframe]{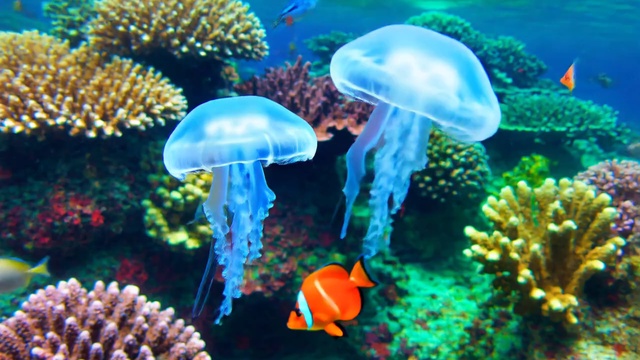}\hspace{\vsgap}\includegraphics[width=\vsframe]{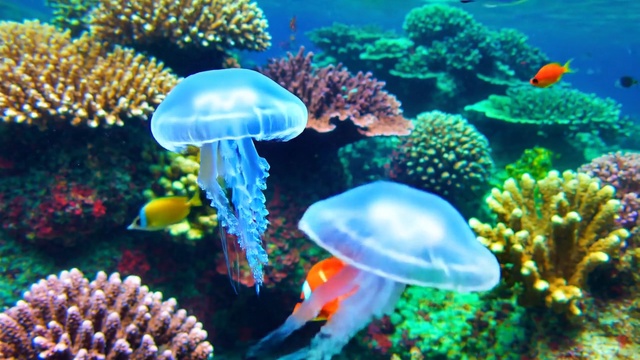}\hspace{\vsgap}\includegraphics[width=\vsframe]{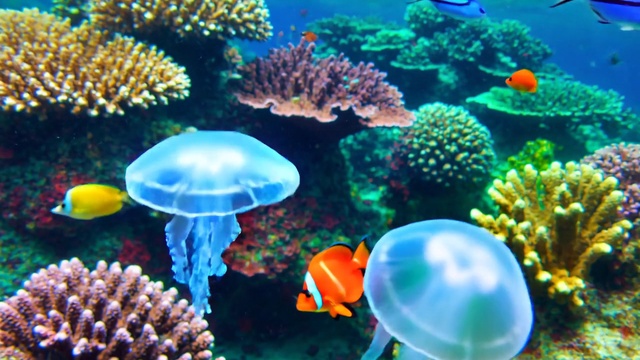}\hspace{\vsgap}\includegraphics[width=\vsframe]{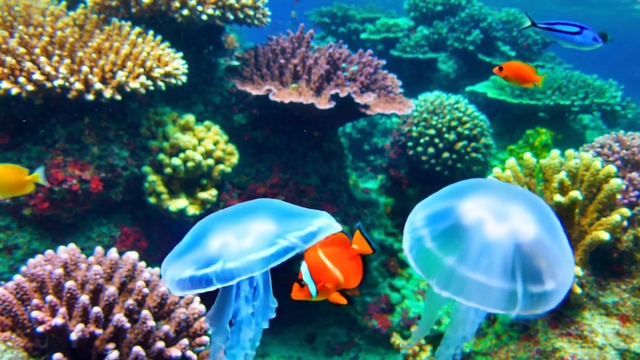}\hspace{\vsgap}\includegraphics[width=\vsframe]{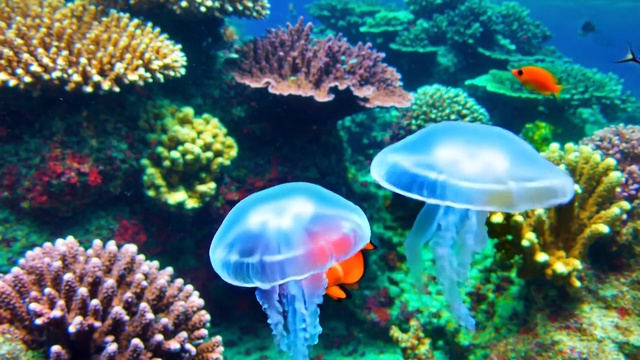}\hspace{\vsgap}\includegraphics[width=\vsframe]{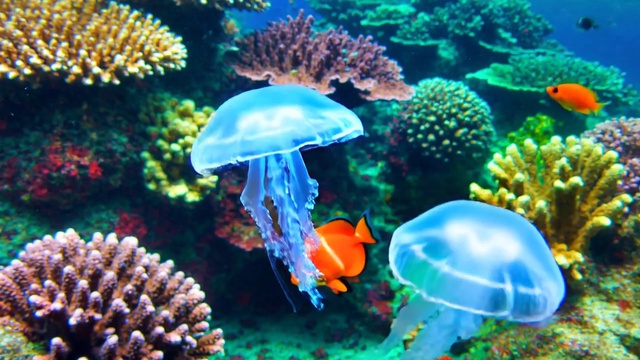}}\\
    \caption{Video generation with Wan2.1-T2V-14B. Top: the base model. Bottom: JetDiffusion.
    Text prompt: \textit{``Underwater jellyfish-like creatures perform the jelly dance in a vibrant coral reef. These translucent beings pulse and sway gracefully, their tentacles flowing rhythmically as they move to an unseen beat. Brightly colored fish and schools of smaller jellyfish swim around them, adding to the lively atmosphere. The background showcases a variety of coral formations and underwater plant life, creating a dynamic and colorful environment. The scene is a close-up, capturing the fluid motions and intricate details of the dancing jellyfish.''}}
    \label{fig:video_wan14b_s15}
\end{figure}

\begin{figure}[!htb]
    \centering\lineskip=1pt
    {\small\textbf{Wan2.1-T2V-14B}}\\[1pt]
    \makebox[\linewidth][c]{\includegraphics[width=\vsframe]{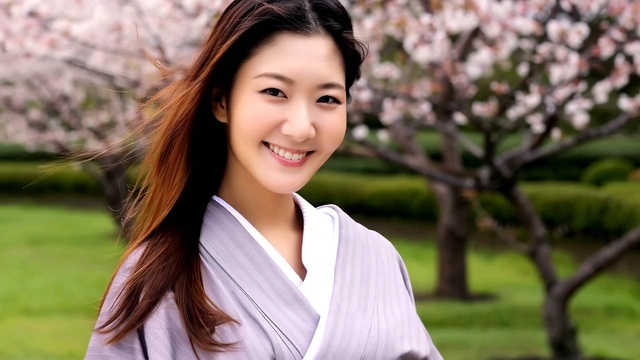}\hspace{\vsgap}\includegraphics[width=\vsframe]{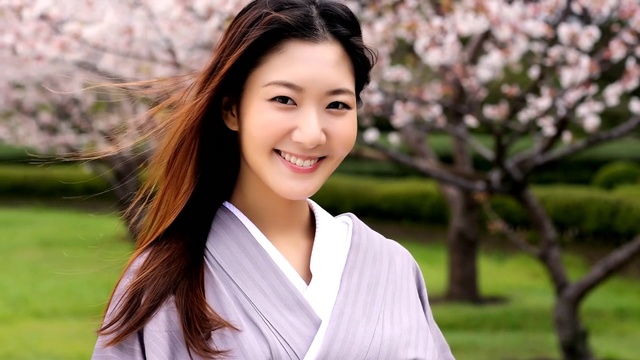}\hspace{\vsgap}\includegraphics[width=\vsframe]{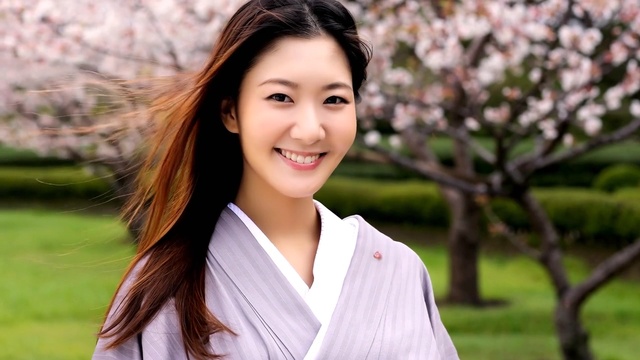}\hspace{\vsgap}\includegraphics[width=\vsframe]{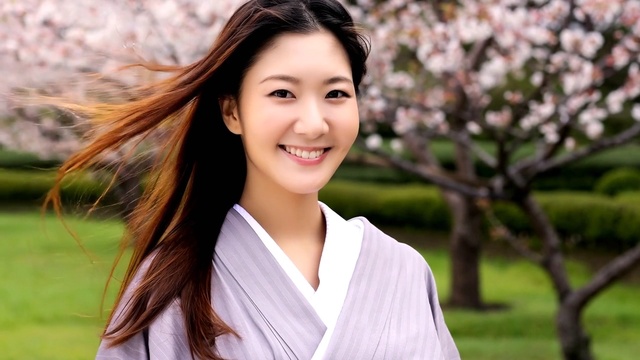}\hspace{\vsgap}\includegraphics[width=\vsframe]{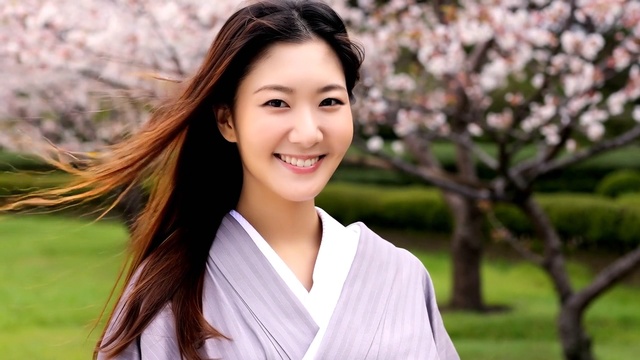}\hspace{\vsgap}\includegraphics[width=\vsframe]{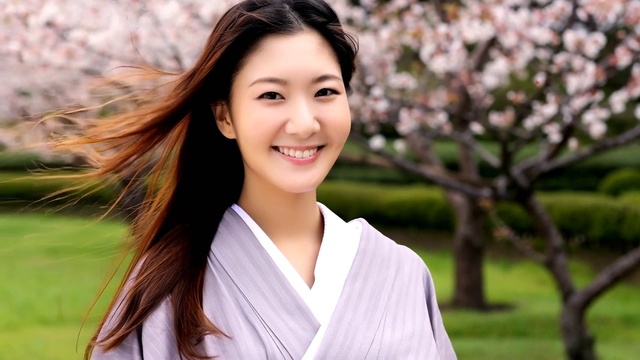}}\\
    [3pt]{\small\textbf{JetDiffusion (Ours)}}\\[1pt]
    \makebox[\linewidth][c]{\includegraphics[width=\vsframe]{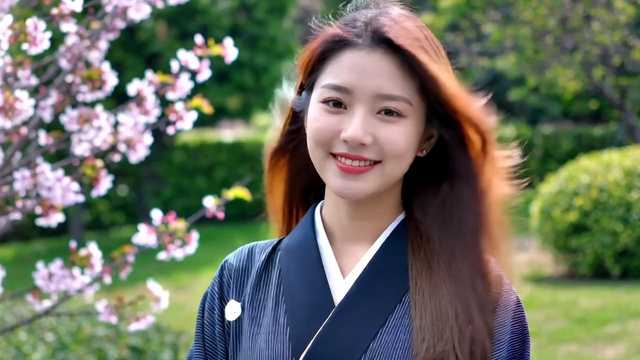}\hspace{\vsgap}\includegraphics[width=\vsframe]{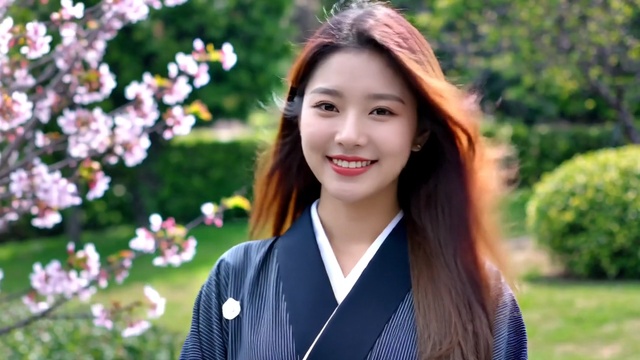}\hspace{\vsgap}\includegraphics[width=\vsframe]{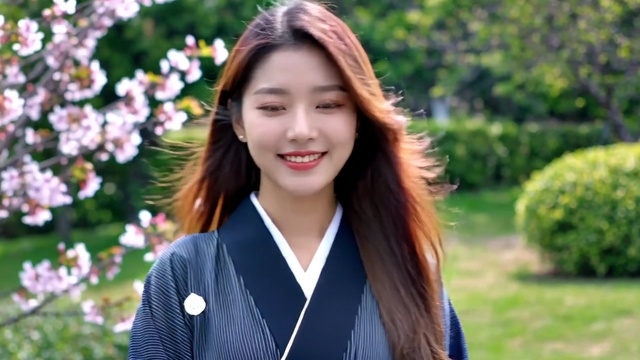}\hspace{\vsgap}\includegraphics[width=\vsframe]{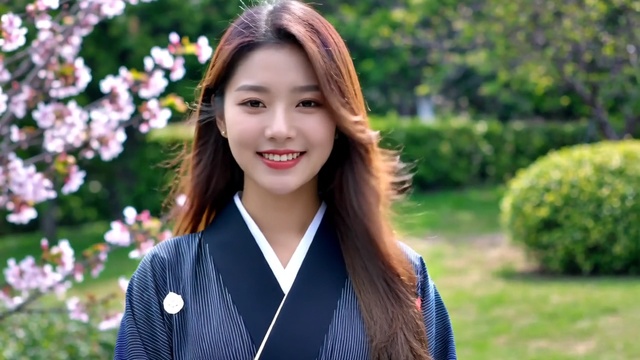}\hspace{\vsgap}\includegraphics[width=\vsframe]{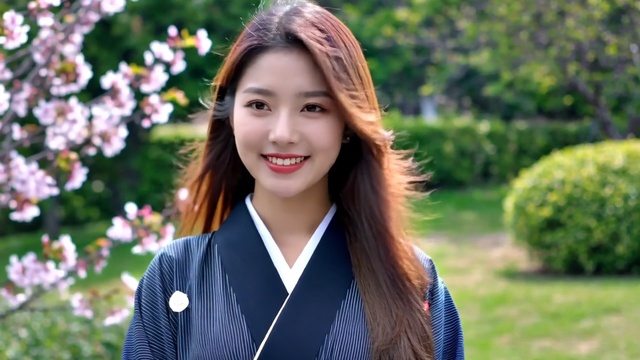}\hspace{\vsgap}\includegraphics[width=\vsframe]{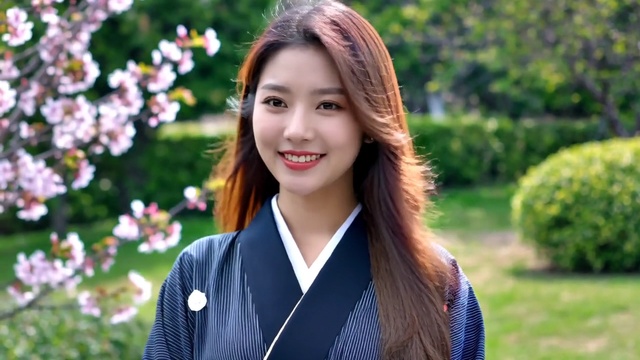}}\\
    \caption{Video generation with Wan2.1-T2V-14B. Top: the base model. Bottom: JetDiffusion.
    Text prompt: \textit{``A video showcasing an East Asian beauty with a warm smile as she stands gracefully in a gentle breeze. She has long flowing hair that gently sways with the wind, accentuating her serene expression. Her attire consists of a traditional yet modern kimono, complementing her elegant posture. The background features a picturesque garden with cherry blossoms blooming in full splendor, adding a touch of nature's beauty. The scene captures her smiling face and flowing hair in a medium close-up, emphasizing the natural beauty and tranquility of the moment.''}}
    \label{fig:video_wan14b_s16}
\end{figure}

\begin{figure}[!htb]
    \centering\lineskip=1pt
    {\small\textbf{Wan2.1-T2V-14B}}\\[1pt]
    \makebox[\linewidth][c]{\includegraphics[width=\vsframe]{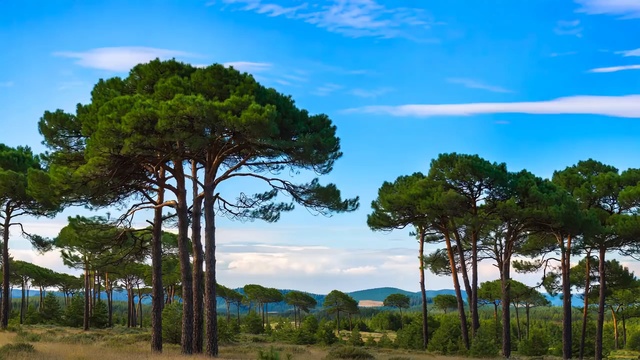}\hspace{\vsgap}\includegraphics[width=\vsframe]{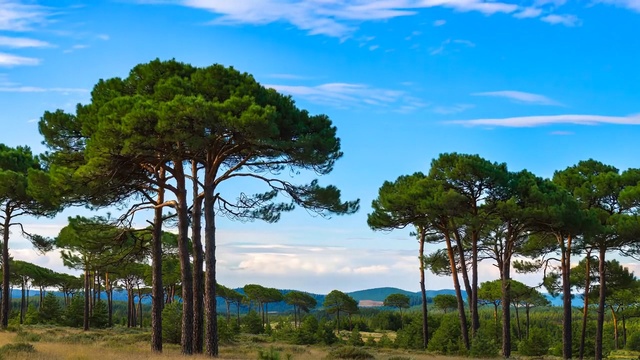}\hspace{\vsgap}\includegraphics[width=\vsframe]{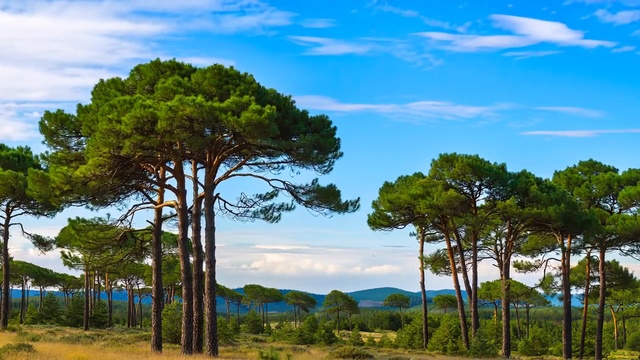}\hspace{\vsgap}\includegraphics[width=\vsframe]{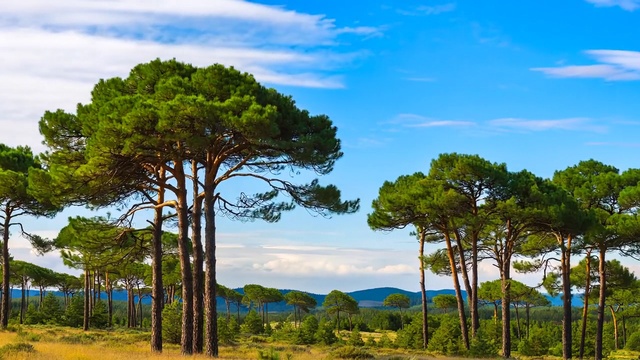}\hspace{\vsgap}\includegraphics[width=\vsframe]{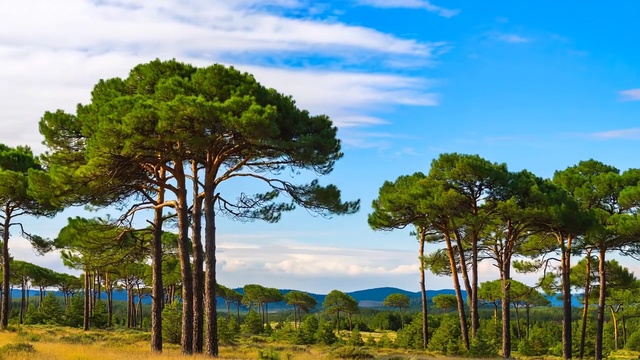}\hspace{\vsgap}\includegraphics[width=\vsframe]{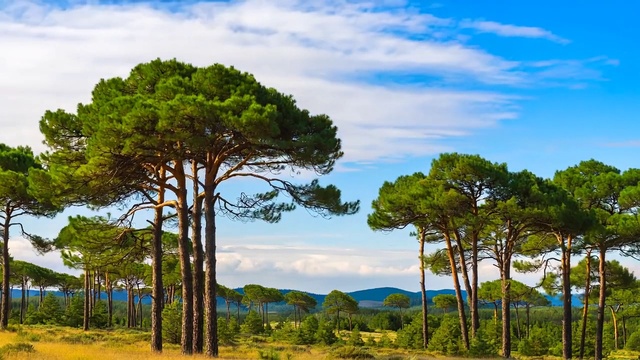}}\\
    [3pt]{\small\textbf{JetDiffusion (Ours)}}\\[1pt]
    \makebox[\linewidth][c]{\includegraphics[width=\vsframe]{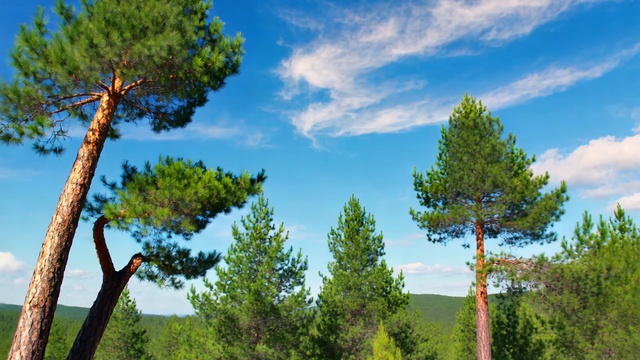}\hspace{\vsgap}\includegraphics[width=\vsframe]{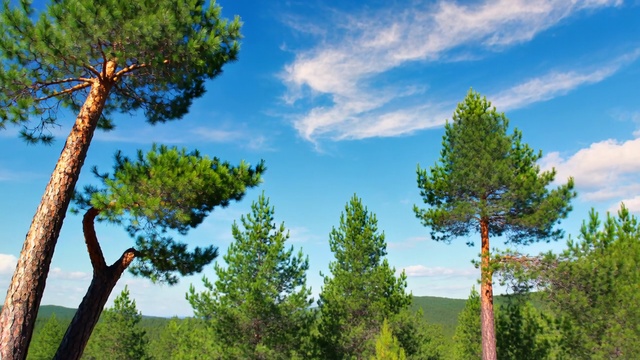}\hspace{\vsgap}\includegraphics[width=\vsframe]{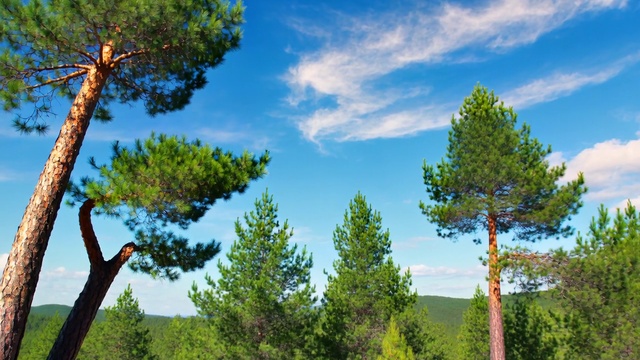}\hspace{\vsgap}\includegraphics[width=\vsframe]{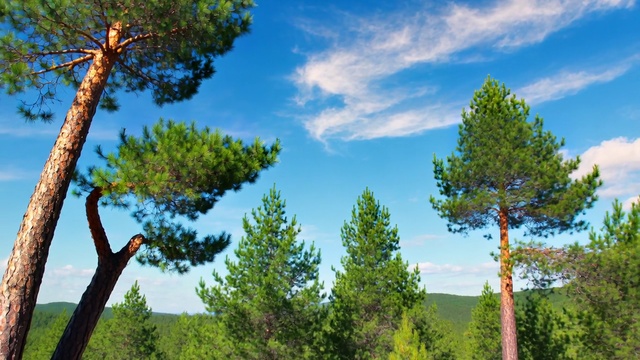}\hspace{\vsgap}\includegraphics[width=\vsframe]{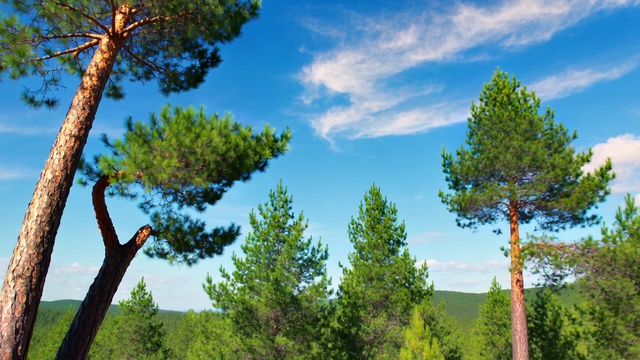}\hspace{\vsgap}\includegraphics[width=\vsframe]{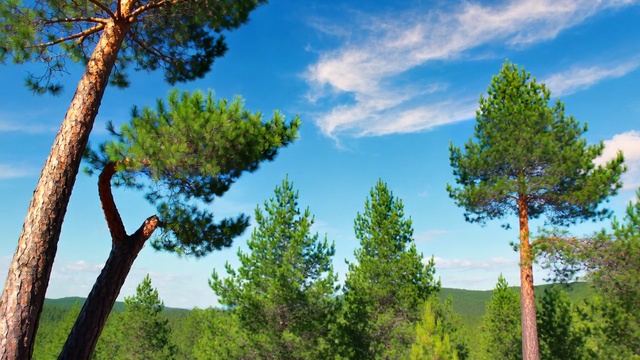}}\\
    \caption{Video generation with Wan2.1-T2V-14B. Top: the base model. Bottom: JetDiffusion.
    Text prompt: \textit{``A serene landscape featuring tall pine trees swaying gently in the breeze on a clear blue sky day with wispy clouds drifting across the sky. The trees have lush green needles and rugged trunks, their branches bending gracefully under the wind's touch. The scene is bathed in soft sunlight, casting dappled shadows on the forest floor. In the background, a few distant hills and a bright blue sky with scattered clouds create a peaceful atmosphere. The camera captures this tranquil environment from a mid-shot perspective, emphasizing the natural beauty and tranquility of the scene.''}}
    \label{fig:video_wan14b_s17}
\end{figure}

\begin{figure}[!htb]
    \centering\lineskip=1pt
    {\small\textbf{Wan2.1-T2V-14B}}\\[1pt]
    \makebox[\linewidth][c]{\includegraphics[width=\vsframe]{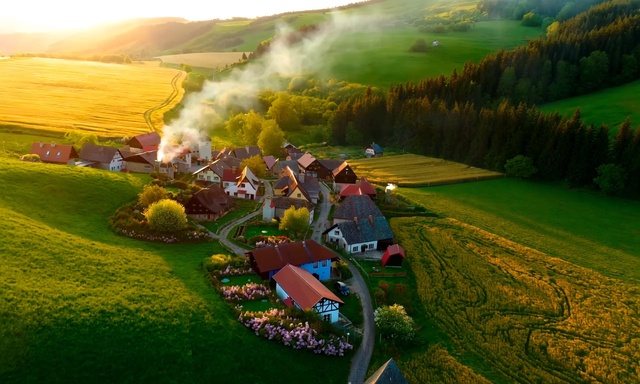}\hspace{\vsgap}\includegraphics[width=\vsframe]{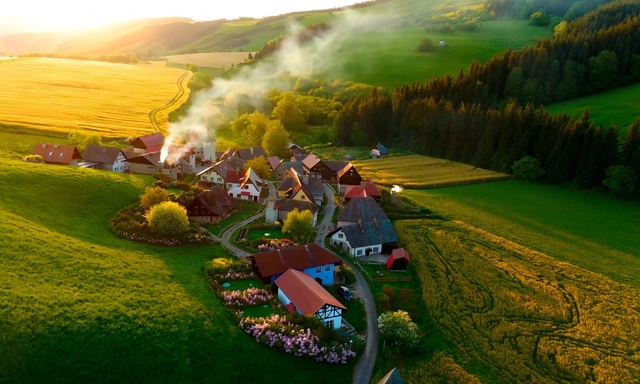}\hspace{\vsgap}\includegraphics[width=\vsframe]{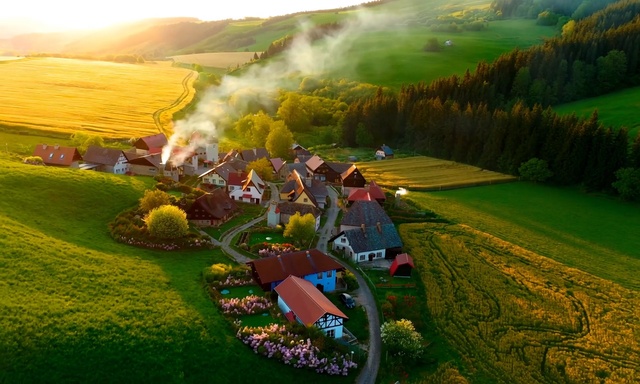}\hspace{\vsgap}\includegraphics[width=\vsframe]{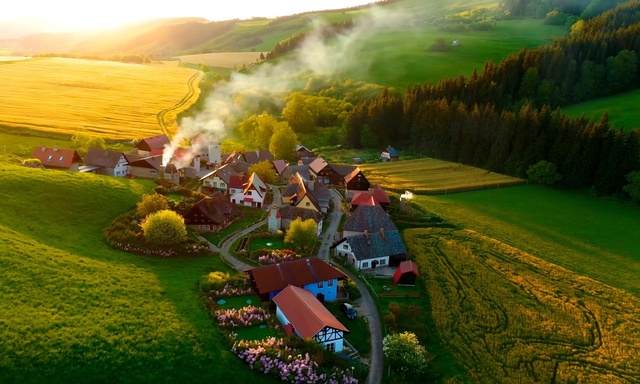}\hspace{\vsgap}\includegraphics[width=\vsframe]{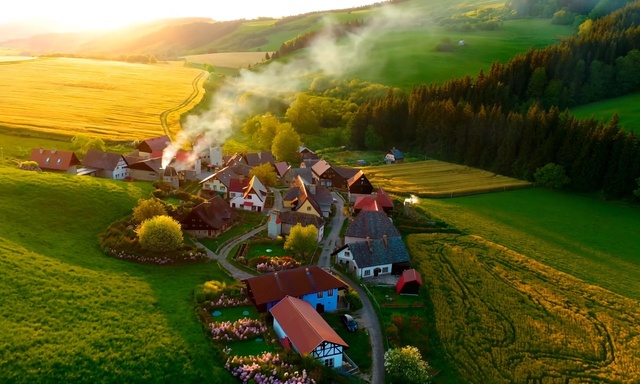}\hspace{\vsgap}\includegraphics[width=\vsframe]{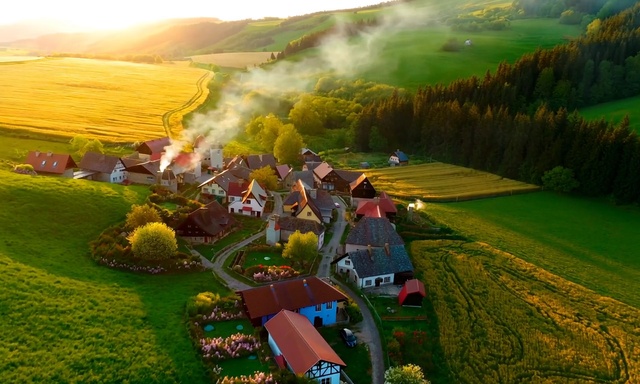}}\\
    [3pt]{\small\textbf{JetDiffusion (Ours)}}\\[1pt]
    \makebox[\linewidth][c]{\includegraphics[width=\vsframe]{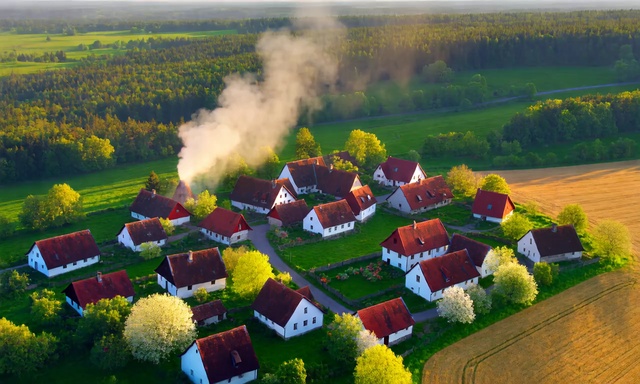}\hspace{\vsgap}\includegraphics[width=\vsframe]{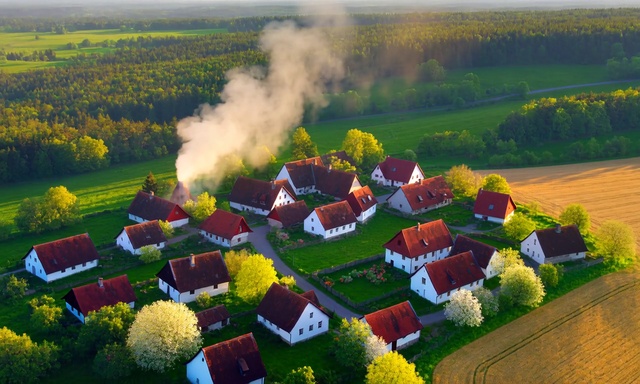}\hspace{\vsgap}\includegraphics[width=\vsframe]{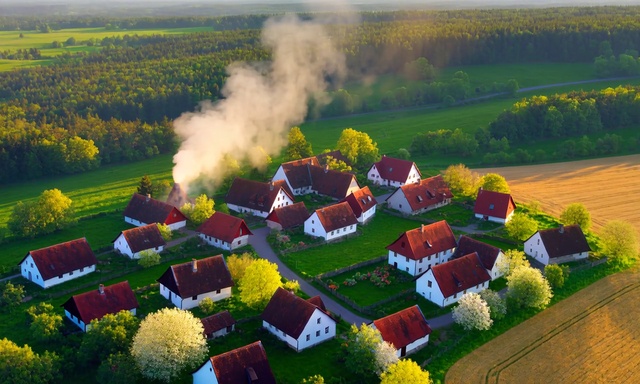}\hspace{\vsgap}\includegraphics[width=\vsframe]{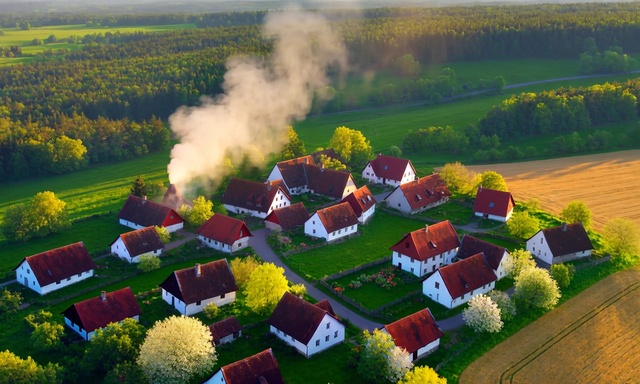}\hspace{\vsgap}\includegraphics[width=\vsframe]{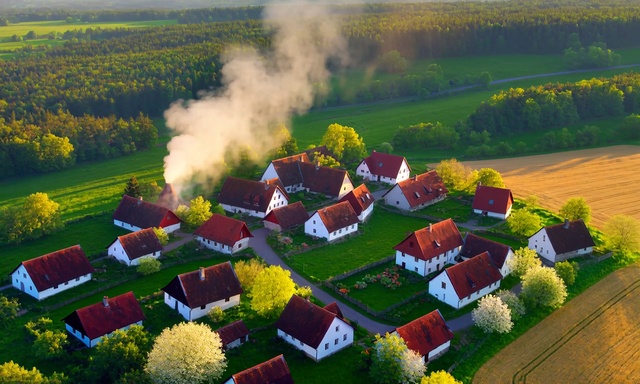}\hspace{\vsgap}\includegraphics[width=\vsframe]{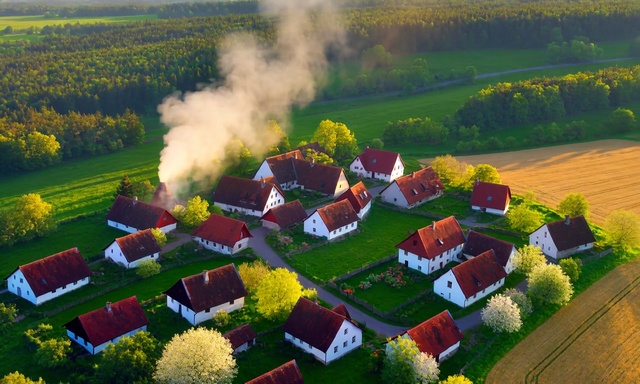}}\\
    \caption{Video generation with Wan2.1-T2V-14B. Top: the base model. Bottom: JetDiffusion.
    Text prompt: \textit{``Aerial view of a peaceful village nestled between lush, rolling green hills. The village consists of quaint, colorful houses with red roofs, winding cobblestone streets, and small gardens filled with blooming flowers. Smoke gently rises from chimneys as the sun casts a warm glow over the scene. The landscape is dotted with fields of golden wheat and patches of dense forests. The camera slowly descends to a medium-altitude shot, capturing the tranquility and beauty of the rural setting. Wide shot.''}}
    \label{fig:video_wan14b_s18}
\end{figure}

\begin{figure}[!htb]
    \centering\lineskip=1pt
    {\small\textbf{Wan2.1-T2V-14B}}\\[1pt]
    \makebox[\linewidth][c]{\includegraphics[width=\vsframe]{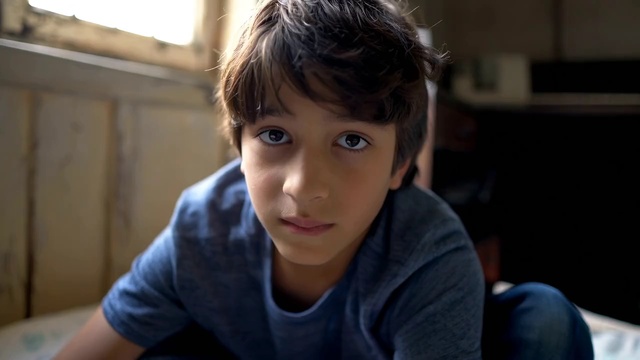}\hspace{\vsgap}\includegraphics[width=\vsframe]{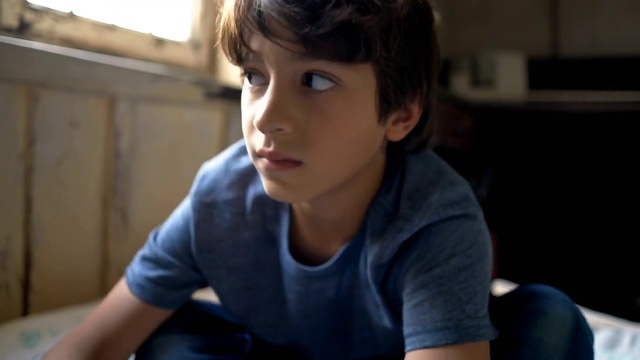}\hspace{\vsgap}\includegraphics[width=\vsframe]{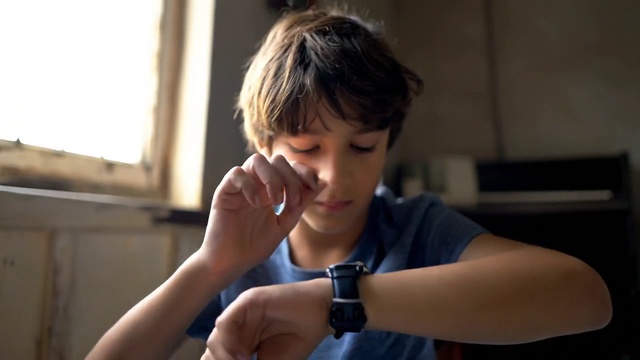}\hspace{\vsgap}\includegraphics[width=\vsframe]{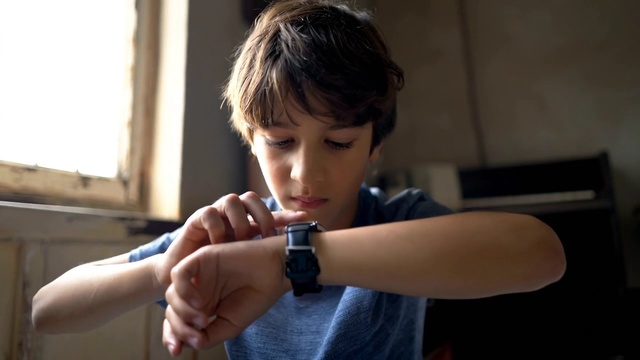}\hspace{\vsgap}\includegraphics[width=\vsframe]{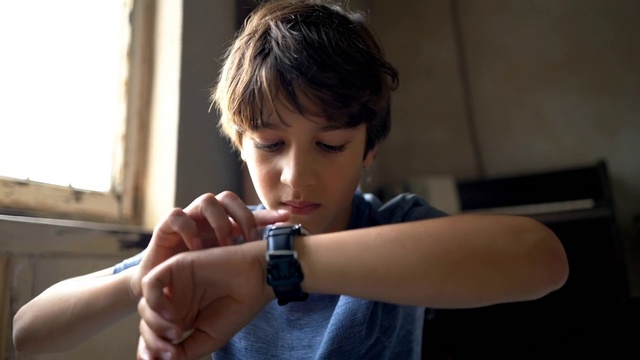}\hspace{\vsgap}\includegraphics[width=\vsframe]{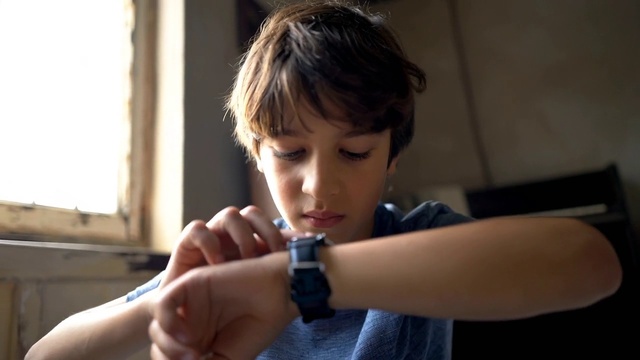}}\\
    [3pt]{\small\textbf{JetDiffusion (Ours)}}\\[1pt]
    \makebox[\linewidth][c]{\includegraphics[width=\vsframe]{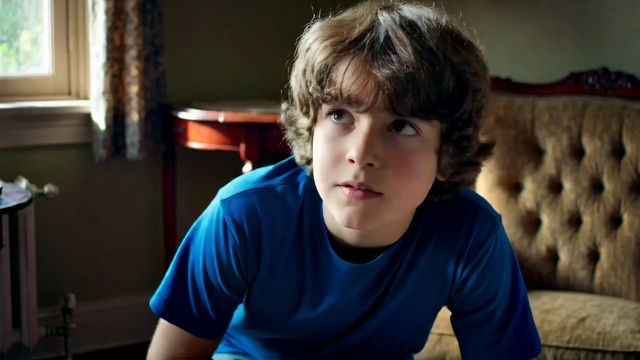}\hspace{\vsgap}\includegraphics[width=\vsframe]{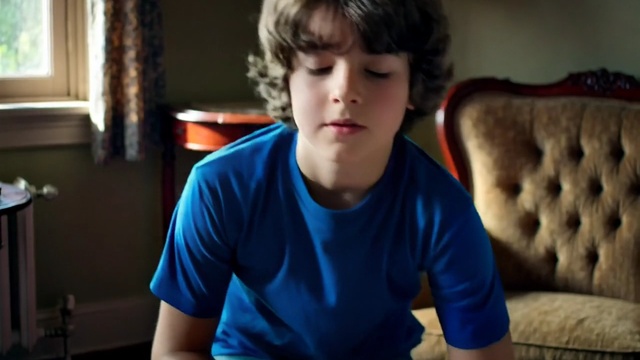}\hspace{\vsgap}\includegraphics[width=\vsframe]{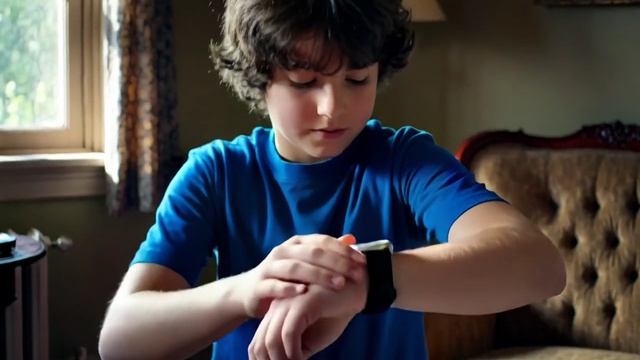}\hspace{\vsgap}\includegraphics[width=\vsframe]{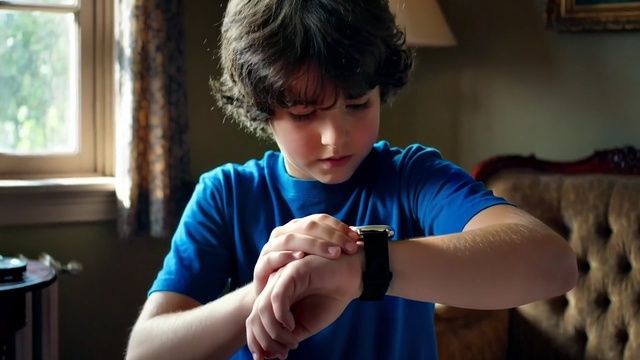}\hspace{\vsgap}\includegraphics[width=\vsframe]{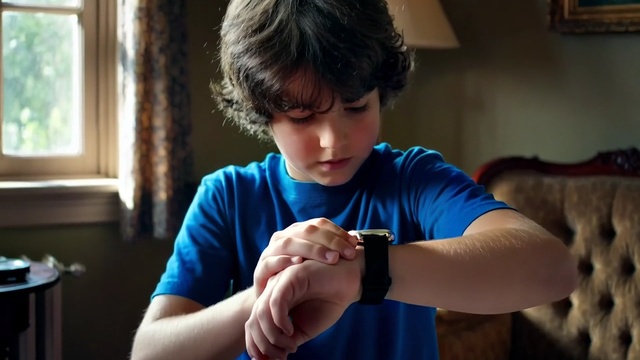}\hspace{\vsgap}\includegraphics[width=\vsframe]{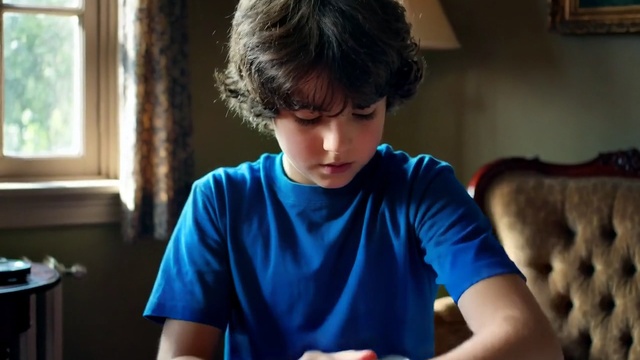}}\\
    \caption{Video generation with Wan2.1-T2V-14B. Top: the base model. Bottom: JetDiffusion.
    Text prompt: \textit{``A young boy blinks his eyes and realizes he is somewhere unfamiliar, reaching to check his wristwatch repeatedly. He has messy brown hair and wears a casual outfit consisting of a blue t-shirt and jeans. His expression shows a mix of confusion and urgency as he glances down at his watch each time. The background includes a dimly lit room with old furniture and a window letting in some sunlight. The scene captures the boy's anxious movements and the close-up detail of his watch. Close-up, medium shot focusing on the boy's face and hands.''}}
    \label{fig:video_wan14b_s19}
\end{figure}

\begin{figure}[!htb]
    \centering\lineskip=1pt
    {\small\textbf{Wan2.1-T2V-14B}}\\[1pt]
    \makebox[\linewidth][c]{\includegraphics[width=\vsframe]{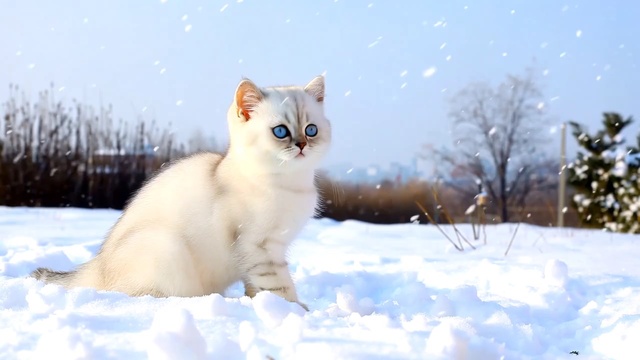}\hspace{\vsgap}\includegraphics[width=\vsframe]{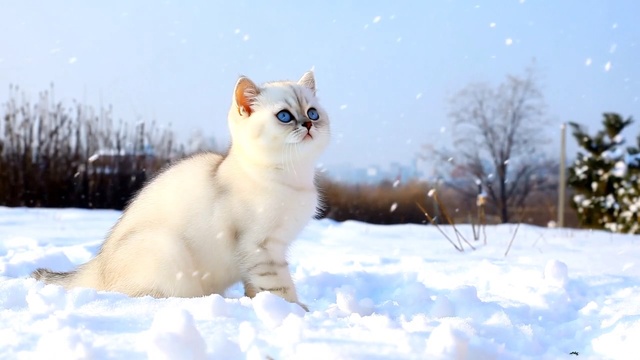}\hspace{\vsgap}\includegraphics[width=\vsframe]{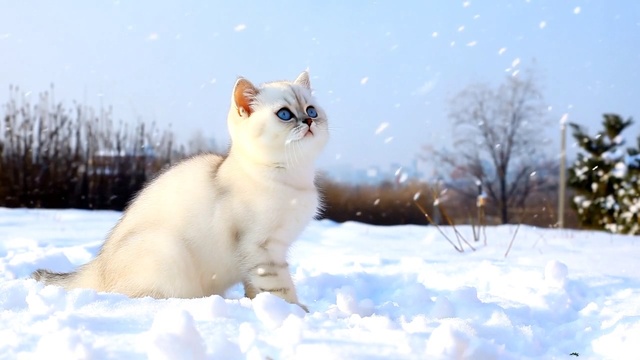}\hspace{\vsgap}\includegraphics[width=\vsframe]{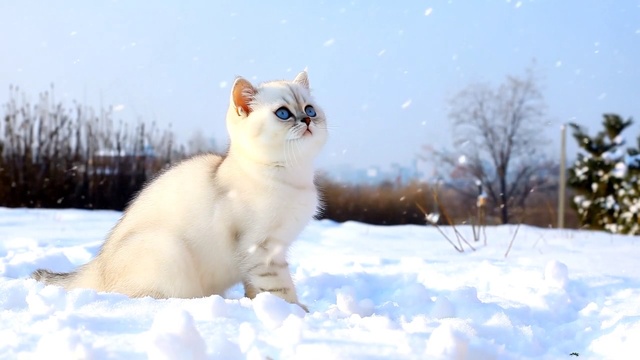}\hspace{\vsgap}\includegraphics[width=\vsframe]{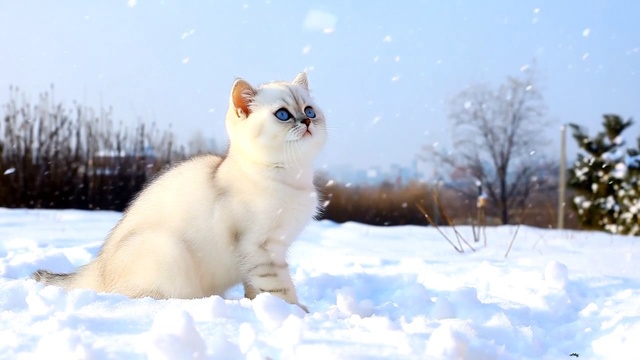}\hspace{\vsgap}\includegraphics[width=\vsframe]{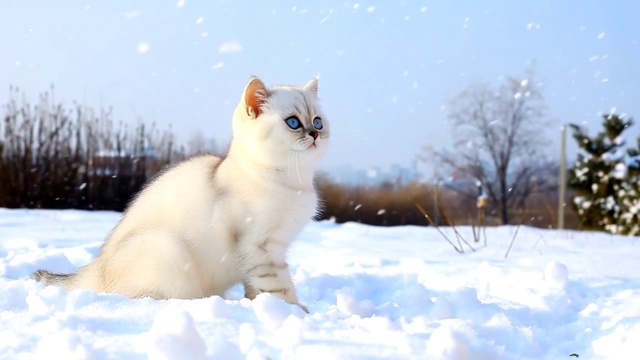}}\\
    [3pt]{\small\textbf{JetDiffusion (Ours)}}\\[1pt]
    \makebox[\linewidth][c]{\includegraphics[width=\vsframe]{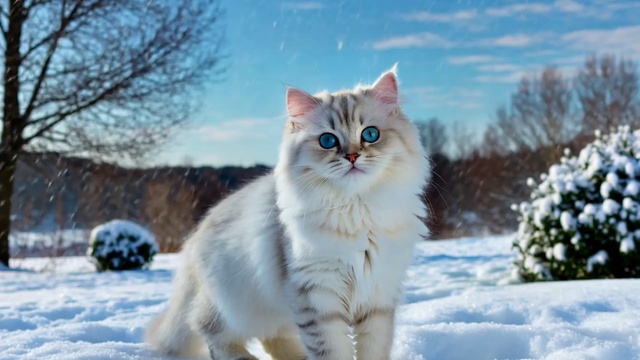}\hspace{\vsgap}\includegraphics[width=\vsframe]{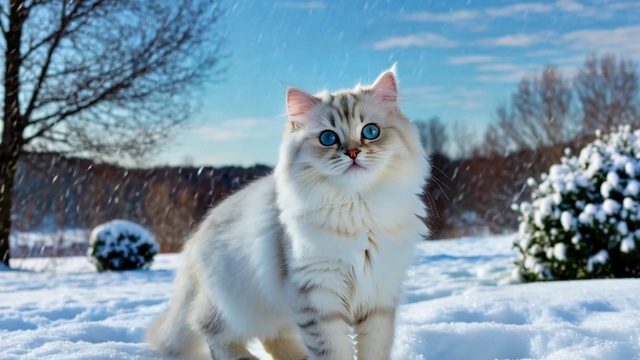}\hspace{\vsgap}\includegraphics[width=\vsframe]{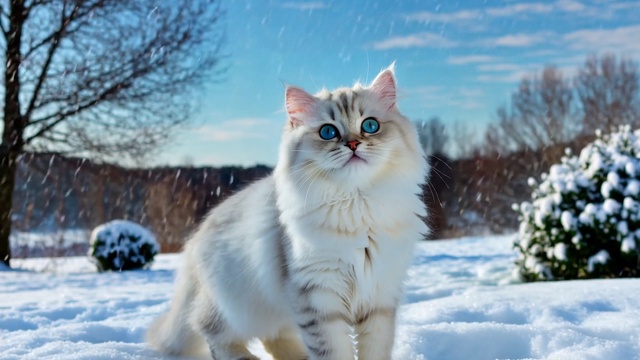}\hspace{\vsgap}\includegraphics[width=\vsframe]{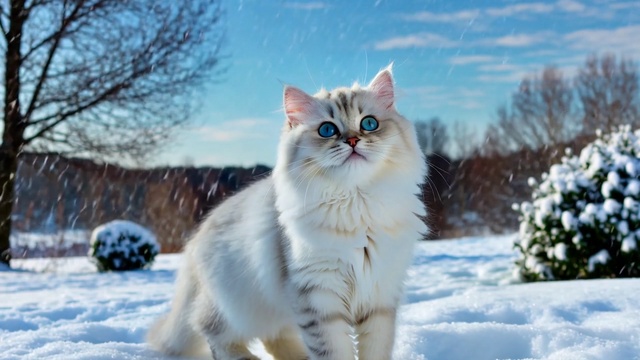}\hspace{\vsgap}\includegraphics[width=\vsframe]{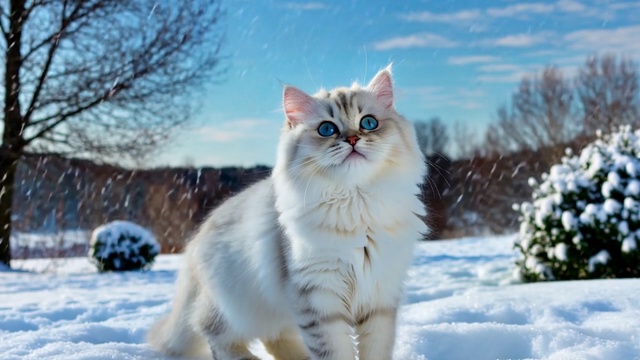}\hspace{\vsgap}\includegraphics[width=\vsframe]{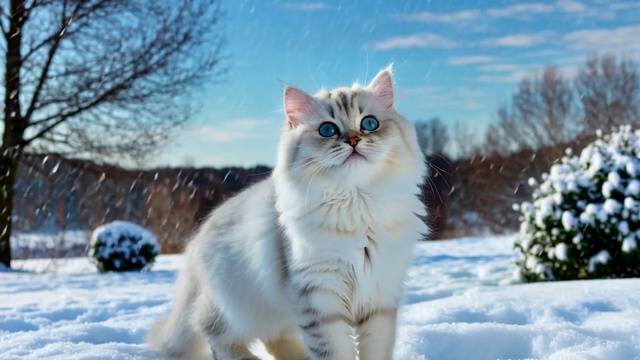}}\\
    \caption{Video generation with Wan2.1-T2V-14B. Top: the base model. Bottom: JetDiffusion.
    Text prompt: \textit{``A serene winter scene with gentle snowflakes falling softly from a clear, pale blue sky. In the foreground, a fluffy, white kitten with large, expressive blue eyes stands in a relaxed posture, gently pawing at the snowflakes as they land on its fur. The kitten has a soft, curious expression, tilting its head slightly upwards to watch the snowfall. The background features a snowy landscape with bare trees and a few scattered snow-covered bushes. The scene is captured in a medium close-up, focusing on the kitten and the snowfall around it, with a calm and peaceful atmosphere.''}}
    \label{fig:video_wan14b_s20}
\end{figure}

\FloatBarrier

\end{document}